%% file: 00-main.tex
\documentclass[manuscript,screen, nonacm]{acmart}
\renewcommand\footnotetextcopyrightpermission[1]{}
\AtBeginDocument{%
  \providecommand\BibTeX{{%
    \normalfont B\kern-0.5em{\scshape i\kern-0.25em b}\kern-0.8em\TeX}}}

\input{000_newcommands}

\setcopyright{acmlicensed}
\copyrightyear{2018}
\acmYear{2018}
\acmDOI{XXXXXXX.XXXXXXX}

\acmConference[Conference acronym 'XX]{Make sure to enter the correct
  conference title from your rights confirmation emai}{June 03--05,
  2018}{Woodstock, NY}
\acmBooktitle{Woodstock '18: ACM Symposium on Neural Gaze Detection,
 June 03--05, 2018, Woodstock, NY} 
\acmISBN{978-1-4503-XXXX-X/18/06}

\begin{document}

\title{Public Perceptions of Fairness Metrics Across Borders}

\author{Yuya Sasaki}
\affiliation{%
  \institution{The University of Osaka}
  \city{Suita}
  \country{Japan}}
\email{sasaki@ist.osaka-u.ac.jp}

\author{Sohei Tokuno}
\affiliation{%
  \institution{NAIST}
  \city{Nara}
  \country{Japan}
}

\author{Haruka Maeda}
\affiliation{%
  \institution{Kyoto University}
  \city{Kyoto}
  \country{Japan}
}

\author{Kazuki Nakajima}
\affiliation{%
  \institution{Tokyo Metropolitan University}
  \city{Tokyo}
  \country{Japan}
}

\author{Osamu Sakura}
\affiliation{%
  \institution{The University of Tokyo}
  \city{Tokyo}
  \country{Japan}
}

\author{George Fletcher}
\affiliation{%
  \institution{Eindhoven University of Technology}
  \city{Eindhoven}
  \country{The Netherland}
}

\author{Mykola Pechenizkiy}
\affiliation{%
  \institution{Eindhoven University of Technology}
  \city{Eindhoven}
  \country{The Netherlands}
}

\author{Panagiotis Karras}
\affiliation{%
  \institution{University of Copenhagen}
  \city{Copenhagen}
  \country{Denmark}
}

\author{Irina Shklovskii}
\affiliation{%
  \institution{University of Copenhagen}
  \city{Copenhagen}
  \country{Denmark}
}


\begin{abstract}
\input{01-abstract}
\end{abstract}




\received{20 February 2007}
\received[revised]{12 March 2009}
\received[accepted]{5 June 2009}

\maketitle

\input{1-intro}

\input{2-related}
\input{3-method}

\input{4-result}

\input{5-discussion}
\input{6-conclusion}

\newpage
\section*{Ethical concerns}
In this paper, we conducted questionnaires to evaluate fairness metrics in multiple countries.
In the questionnaires, we followed laws of each country and consult lawyers to ask legal issues about protection of personal information and anti-discrimination.
We did not collect personal information (e.g., names and addresses) to identify individuals. In addition, we only opened statistical results instead of each answer.

We analyzed the difference in decision-making between personal attributes, including gender, age, religion, and education.
Our results revealed the impact of personal attributes on fairness decisions.
They are potentially misused or abused in AI systems.
On the other hand, we compared our results with those in past studies to carefully use fairness metrics in decision-making scenarios, and thus it may work to avoid misusing the past results.
We believe our study helps to build fair AI systems from various perspectives.

\bibliographystyle{ACM-Reference-Format}
\bibliography{aifairnesssurvey}

\newpage
 \appendix

\input{99-appendix}

\end{document}

%% file: 000_newcommands.tex
\usepackage{times}
\usepackage{soul}
\usepackage{url}
\usepackage[utf8]{inputenc}
\usepackage{graphicx}
\usepackage{amsmath}
\usepackage{amsthm}
\usepackage{booktabs}

\usepackage[subrefformat=parens]{subcaption}

\usepackage{ascmac}
\usepackage{listings}
\usepackage{xcolor}

\usepackage{subcaption}

\usepackage{multirow}

\definecolor{mycolor}{RGB}{216, 216, 255}


%% file: 01-abstract.tex
Which fairness metrics are appropriately applicable in your contexts?
There may be instances of discordance regarding the perception of fairness, even when the outcomes comply with established fairness metrics.
Several questionnaire-based surveys have been conducted to evaluate fairness metrics with human perceptions of fairness. However, these surveys were limited in scope, including only a single country. 
In this study, we conduct an international survey to evaluate public perceptions of various fairness metrics in decision-making scenarios. 
We collected responses from 1,000 participants in each of China, France, Japan, and the United States, amassing a total of 4,000 participants, to analyze the preferences of fairness metrics. Our survey consists of three distinct scenarios paired with four fairness metrics. This investigation explores the relationship between personal attributes and the choice of fairness metrics, uncovering a significant influence of national context on these preferences.


%% file: 1-intro.tex
\section{Introduction}
Fairness in artificial intelligence and machine learning systems is an emerging problem. 
Notable examples include gender bias in job-related ads~\cite{datta2014automated}, racial bias in evaluating names on resume~\cite{caliskan2017semantics}, racial bias in predicting criminal recidivism~\cite{angwin2022machine}.
To address these concerns, fair machine learning methods have been studied to mitigate biased outputs against sensitive attributes (e.g., gender and race) of individuals~\cite{10.1145/3616865}. This approach is crucial for preventing the perpetuation of societal discrimination.

Group fairness aims to ensure equitable prediction outcomes across groups of people associated with sensitive attributes~\cite{mehrabi2021survey}.
The evaluation of group fairness is conducted through the use of fairness metrics, which provide a quantifiable measure of the extent to which the results of these methods are equitable in sensitive attributes. 
The fairness metrics have been extensively studied, aiming to meet diverse user needs within various contexts, such as demographic parity~\cite{chouldechova2017fair} and equal opportunity~\cite{hardt2016equality}. 
It is practically difficult to simultaneously satisfy all fairness metrics or even theoretically impossible~\cite{hardt2016equality}. 
This observation brings us to a crucial question: {\it Which fairness metrics should be selected in our contexts?}
Let us consider a hiring example where we would like to fairly hire male and female candidates.
What fairness metrics are suitable for our context to evaluate the balance of accepted male and female candidates, and what would the public perceive as fair in our decision?

To address the question, it is necessary to comprehend the perceptions of fairness among users, particularly those who are not experts in the field of AIs and fairness, across various scenarios. 
Several survey-based studies conducted questionnaires to understand opinions on fairness metrics and effects of personal attributes~\cite{srivastava2019mathematical,cheng2021soliciting,sengewald2023}.
Table~\ref{tab:questionnaire} shows the summary of empirical studies on surveys that evaluated human perceptions of fairness metrics.
These surveys have been limited in scale, generally involving a few hundred participants from a single country.
Thus, they are not suitable to understand (1) perception across countries and (2) detailed personal attribute analysis.
First, since the extent of societal inequality often varies from country to country\footnote{\url{https://www.oecd.org/social/inequality.htm/}}, we expect that perceptions of fairness also vary significantly across countries as well as personal attributes. 
Second, a few hundred participants may not be sufficient to robustly analyze statistical differences between fine personal attributes.

\begin{table}[t]
\centering
\caption{Empirical studies for evaluating fairness metrics.}
\label{tab:questionnaire}
\centering
\scalebox{1.0}{
\begin{tabular}{lcc} \toprule
\multicolumn{1}{c}{{\bf Paper}} & \multicolumn{1}{c}{{\bf \# participants}} & \multicolumn{1}{c}{{\bf Country}}\\ \midrule
\cite{srivastava2019mathematical}   & 100 & US \\
\cite{cheng2021soliciting}  & 12 & US\\
\cite{sengewald2023}  & 258 & Germany \\\midrule
\multirow{2}{*}{Ours} & 4,000 & China, France,  \\
&(1,532) & Japan, US \\
\bottomrule
\end{tabular}
}
\end{table}

In this study, we conduct a large-scale questionnaire-based survey across multiple countries to answer three research questions:
\begin{itemize}
    \item {\bf RQ1}: Should we use a single fairness metric across countries? If no, what are some implications of differences in the perceptions of fairness metrics across countries?
    \item {\bf RQ2}: What countries have similar public perceptions of fairness metrics in decision-making scenarios?
    \item {\bf RQ3}: Do personal attributes such as gender and ethnicity have similar effects on public perceptions of fairness metrics even in different countries?
\end{itemize}
This survey includes a total of 4,000 participants, with 1,000 each from China, France, Japan, and the United States. 
The large-scale survey enables us to find the statistical differences between personal attributes and the choice of fairness metrics.

We design a questionnaire to compile statistics on the correlations between personal attributes and the choice of fairness metrics. We aim to reveal the specific personal attributes that significantly influence the choice of fairness metrics.
The questionnaire consists of three sections: personal attributes of the participants, explanations of fairness metrics, and decision-making scenarios.
First, participants provide information on their age, gender, ethnicity\footnote{As France is not allowed to collect ethnicity legally, we did not collect them in France.}, religion, education, and experience~\footnote{We do not have any intentions of the orders of personal attributes and their elements.}.
Subsequently, we explain four fairness metrics: quantitative parity, demographic parity, equal opportunity, and equalized odds. Notably, quantitative parity, a simple metric that can be considered fair if the number of positive predictions across groups is equal, has not been explored in fairness literature.
In the decision-making scenarios, participants are asked to agree the reality of given scenarios and the applicability of the fairness metrics.
We use three scenarios; hiring, art project, and employee award.
Finally, they choose their best single metric for each scenario.

We investigate the difference in selected fairness metrics between countries, genders, religions, ethnicities, and other personal attributes in each scenario.
We used 1,532 participants who correctly understood fairness metrics among 4,000 participants.
The statistical test can be meaningfully derived thanks to a large number of participants.
We summarize our key findings as follows:
\begin{itemize}
    \item We statistically reveal a single fairness metric cannot be applied globally.
    \item The choice of fairness metrics is most significantly influenced by the country of the participants. Notably, participants from France demonstrate a preference for quantitative parity. 
    \item Gender differences in metric selection are not significant in most scenarios. A trend is slightly observed where males tend to favor demographic parity. 
    \item Religion does not have a large impact on the choice of fairness metrics, while ethnicity in different countries has different trends for the choice of fairness metrics.
\end{itemize}

Finally, we discuss the reasons for the influence of the preference of fairness metrics, the difference between results in our and past studies, and the limitations of our study.

%% file: 2-related.tex
\section{Related work}
\label{sec:related}

\input{table/relatedwork}

We review existing works related to fairness metrics and empirical studies.

\smallskip
\noindent
{\bf Cross-country/cultural analysis.}
Studies on differences in fairness and morals between countries and between cultures have been widely conducted in various fields such as psychology, economics, and politics~\cite{KIM200783,smith2002cultural, peterson1994event,smith1994event,smith1999social,alesina2005fairness,roth1991bargaining,alesina2001doesn,konow2003fairest,reeskens2012disentangling,awad2018moral,kelley2021exciting,ullstein2024attitudes}.
Existing studies indicate that fairness and justice are influenced not only by individual or organizational differences in welfare, distribution, and employee environments but also by beliefs about ``what is right'' across nations and cultures.
For example, moral machine experiments~\cite{awad2018moral} revealed that morals of autonomous machines are different across areas.
Our study aims to reveal whether the choice of fairness metrics is influenced by countries and personal attributes.

\smallskip
\noindent
{\bf Fairness definitions.}
Several fairness definitions have been studied, such as group fairness, individual fairness, and ranking fairness~\cite{mehrabi2021survey,carey2023statistical}.
Group fairness aims to be considered fair if individuals from different groups (e.g., genders and races) have the same probabilities, such as demographic parity~\cite{chouldechova2017fair}, predictive parity~\cite{hardt2016equality}, equal opportunity~\cite{hardt2016equality}, equalized odds~\cite{pleiss2017fairness}, causality fairness~\cite{kusner2017counterfactual}, and error parity~\cite{buolamwini2018gender}.
Individual fairness aims to be considered fair if individuals who have similar features have the same predictive outputs~\cite{fleisher2021s,kang2020inform}.
There are several definitions to measure similarities between two individuals.
Ranking fairness aims to fairly rank data points, such as image retrieval~\cite{yang2017measuring,celis2017ranking,singh2018fairness}.
This also focuses on sensitive attributes to appear in the top ranks balanced well (e.g., the same number of male and female images within top-10). 

Our survey focuses specifically on metrics related to group fairness.
Some metrics cannot be satisfied simultaneously such as demographic parity and equalized odds~\cite{hardt2016equality,chouldechova2017fair}.
It is intuitive to consider that two groups include the same number of individuals but positive labels are heavily skewed towards one group.
Demographic parity requires predicting an equal number of positives for both groups, while equalized odds requires predicting a skewed number of positives for the group including a large number of positive labels.
This underlies the necessity of selecting metrics that are most appropriate for specific scenarios.
The pressing question remains: {\it In terms of human perceptions, which fairness metrics should be used in our scenarios?}

\smallskip
\noindent
{\bf Empirical study.},
Empirical studies evaluate human perceptions of fairness in specific contexts. Strake et al.~\cite{starke2022fairness} show a systematic survey of fairness perceptions of algorithmic decision-making, but no analysis across countries.
Table~\ref{tab:related} shows a summary of empirical studies.

Woodruff et al.~\cite{woodruff2018qualitative} conducted a workshop and interviews to explore ethical and pragmatic aspects of public perception of algorithmic fairness.
Grgic-Hlaca et al.~\cite{grgic2018human} conducted surveys in COMPAS (Correctional Offender Management Profiling for Alternative Sanctions) systems~\cite{angwin2022machine} to reveal what features (e.g., personality and criminal history) should be avoided for fair decision-making.
Plane et al.~\cite{plane2017exploring} conducted surveys about online advertising to evaluate what personal attributes (e.g., race) have a large impact on responsibility and ethical concerns.

Harison et al.~\cite{harrison2020empirical} conducted surveys to investigate what properties of results (e.g., accuracy and false positive rate) are preferable in COMPAS scenarios. 
Saha et al.~\cite{saha2020measuring} conducted two surveys with 147 and 349 participants to assess understanding of fairness metrics.
The first survey evaluated the comprehension of demographic parity in three scenarios; hiring, art project, and employee award. The second survey examined three fairness metrics, demographic parity, equal opportunity, and equalized odds, in the hiring scenario.
They revealed that equalized odds is hard to understand for non-experts.

A few works have evaluated fairness metrics to investigate appropriate fairness metrics in specific contexts. 
Srivastava et al.~\cite{srivastava2019mathematical} and Sengewald et al.~\cite{sengewald2023} conducted surveys to evaluate fairness metrics in crime risk prediction and skin cancer risk prediction scenarios and in a recruitment context, respectively.
Cheng et al.~\cite{cheng2021soliciting} developed the interactive interface to examine the human notion of fairness metrics. They conducted interviews with social workers or people with children. 
Yurrita et al.~\cite{yurrita2023disentangling} statistically evaluate the effects of explanations, human oversight, and contestability on informational and procedural fairness perceptions in a loan approval scenario, but they did not focus on personal attributes.
In each study, candidates of fairness metrics are different, and preferred metrics are different in each study and context.

These prior studies enriched our understanding of public perception of fairness metrics.
However, their focus was on single-country studies, especially in the United States, and involved small-scale surveys (i.e., at most a few hundred participants).
It led to a difficult statistical comparison of the correlations between personal attributes and choice of fairness metrics.





%% file: table/relatedwork.tex
\begin{table*}[ttt]
\centering
\caption{Summary of empirical study. $*$ indicates that the studies did not mention the nationalities of participants, but they compare the statistics of participants with specific counties.}
\label{tab:related}
\vspace{-2mm}
\centering
\scalebox{0.80}{
\begin{tabular}{llccl}\hline
\multicolumn{1}{c}{{\bf Paper}} & \multicolumn{1}{c}{{\bf Style}}  & \multicolumn{1}{c}{{\bf \# participants}} & \multicolumn{1}{c}{{\bf Country}}  & \multicolumn{1}{c}{{\bf Purpose}} \\\hline
\cite{woodruff2018qualitative}  & Workshop and interview & 44 & US & Exploring ethical and pragmatic aspects of algorithmic fairness.\\
\cite{grgic2018human}  & Online survey & 576 & US & Discovering unfair features for decision-making scenarios.\\
\cite{plane2017exploring}  & Online survey & 891 & US & Surveys for fair online advertising.\\
\cite{harrison2020empirical}  & Online survey & 502 & US$*$ & Evaluating preferable results in COMPAS scenarios.\\
\cite{van2021effect} & Online survey & 80 & US & Evaluating the effect of information presentation on  fairness perceptions. \\
\cite{yurrita2023disentangling} & Online survey & 267 & Unknown & Evaluating the impact of transparency on fairness perceptions.\\
\cite{srivastava2019mathematical}  & Online survey & 100 & US$*$ & Evaluating appropriate fairness metrics.\\
\cite{saha2020measuring}  & Online survey & 147/349 & US$*$ & Evaluating comprehension of fairness metrics.\\
\cite{cheng2021soliciting}  & Interview on video chat & 12 & US & Evaluating appropriate fairness metrics.\\
\cite{sengewald2023}  & Online survey & 258 & Germany$*$ & Evaluating appropriate fairness metrics.\\

\hline
\end{tabular}
}
\end{table*}

%% file: 3-method.tex
\section{Method}
\label{sec:method}

To empirically understand public perception of fairness metrics in multiple countries, we conduct an online questionnaire study. In this study, participants answer their preferences for fairness metrics in decision-making scenarios.

\subsection{Survey Design}
The primary objective of this study is to analyze the influence of personal attributes on the choice of fairness metrics in multiple countries.

\smallskip
\noindent
{\bf Assumption and Purpose.}
People (i.e., non-experts) may not agree with the fairness of services, even when policymakers and service developers design machine learning models that satisfy specific fairness metrics.
We assume that each individual has their own experiences in making fair decisions, meaning they hold unique perceptions of fairness metrics.
In particular, these experiences might be strongly influenced by the countries in which they live~\cite{KIM200783}.
Therefore, we aim to investigate non-experts’ perceptions of fairness metrics across multiple countries to ensure services are fair for people.

\smallskip
\noindent
{\bf Country Selection.}
We select four countries in our survey: China, France, Japan, and the U.S.
The reason why we selected these countries is to reveal the cultural differences between Asia and Western areas \footnote{We discuss the limitation of the selected countries.}. We assume that regions with strong correlations to specific ethnicities and religions exhibit similar trends in fairness metric preferences.
Each country has different situations (e.g., economy, politics, culture, and history).
For example, France and US have higher (i.e., better) gender gap indices than China and Japan~\footnote{\url{https://www.weforum.org/publications/global-gender-gap-report-2024/}}.
It is known that France highly influences southern areas~\cite{awad2018moral}. 
We expect to have different perceptions of fairness.

\smallskip
\noindent
{\bf Fairness Metrics Selection.}
We here evaluate the fairness metrics from two aspects: ``unproportional quota vs proportional quota'' and ``statistical parity vs predictive parity''.
\begin{itemize}
    \item Unproportional quota vs proportional quota: Unproportional quota sets a uniform numerical target regardless of the candidates in groups, while proportional quota varies numerical target depending on the composition of the groups. Each country has different preferences. For example, the Japanese government often uses unproportional quotas~\cite{yokoyama2024can}, and in the US, different quotas are used in different organizations in different situations~\cite{anderson2004pursuit}.
    \item Statistical parity vs Predictive parity: Statistical parity does not focus on the qualification of candidates, while predictive parity does. It is closely tied to whether or not we consider fairness by taking into account the presence or absence of ability.
\end{itemize}

To evaluate these aspects, we use four metrics: quantitative parity, demographic parity, equal opportunity, and equalized odds (see Section~\ref{ssec:metrics}). 
Demographic parity, equal opportunity, and equalized odds are common and often used in empirical studies~\cite{srivastava2019mathematical,cheng2021soliciting,sengewald2023}.
We define {\it quantitative parity} as a simple metric that can be considered fair if the numbers of positive predictions across groups are equal, which is highly related to unproportional quota. 
We note that the quantitative parity is not included in previous studies.


\smallskip
\noindent
{\bf Scenario Selection.}
Our survey includes three scenarios that are similar to those in the previous survey~\cite{saha2020measuring}: Hiring, Art project, and Employee award.
We have other candidates, such as crime risk and skin cancer risk predictions~\cite{srivastava2019mathematical}, but we avoid scenarios related to race and ethnicity due to two reasons: (1) the color-blind policy and (2) different major and minor races and ethnicities across countries, which may not be suitable for the international survey.
We also check how realistic the scenarios are in our survey.

\smallskip
\noindent
{\bf Questionnaire Overview.}
We provide an overview of our questionnaires (see the full description in the appendix).
Our questionnaire consists of three sections: personal attributes, explanations of fairness metrics, and decision-making scenarios.

Initially, participants are provided a consent form, and if they agree on the content, the participants answer their attributes; gender, age, ethnicity, religion, education, and experience. 
Following this, we explain the notions of quantitative parity, demographic parity, equal opportunity, and equalized odds, and the participants take two {\it quizzes} for each metric (i.e., a total of eight quizzes) to check their understanding.  
The final section of the questionnaire involves three decision-making scenarios; hiring, art project, and employee award  (see Section~\ref{ssec:scenarios}). 
Each participant scores their {\it agreement levels} to the reality of scenarios and fairness metrics from one (``very strongly agree'') to seven (``not at all'') in decision-making scenarios.
Subsequently, they choose the {\it single} fairness metric that seems the most appropriate for each scenario.

This survey will allow us to compile statistics on the relationships between personal attributes and the choice of fairness metrics. We aim to reveal the specific personal attributes that significantly influence the choice of fairness metrics in each country.


\subsection{Fairness Definitions}
\label{ssec:metrics}

We explain the fairness metrics used in our survey.
Our scenarios are based on binary classification tasks with binary-sensitive attributes.

\smallskip
\noindent
{\bf Notation.}
Each data sample includes label $y$, sensitive attribute $S$, and other attributes.
$y=1$ and $y=0$ represent positive and negative labels (e.g., the qualified and unqualified in recruitment), respectively.
$S$ is either $s_0$ or $s_1$ as sensitive attributes, for example, male or female.
We use the notation $\hat{y}$ to represent predicted labels, and $\hat{y}=1$ and $\hat{y}=0$ indicate positive and negative predicates (e.g., selected and not selected in recruitment), respectively.

\smallskip
\noindent
{\bf Fairness Metrics.}
The focus of our fairness metrics is to evaluate equitable outcomes between the groups $s_0$ and $s_1$.
We use four fairness metrics: quantitative parity, demographic parity, equal opportunity, and equalized odds.

Quantitative parity and demographic parity do not consider predictive accuracy.
These metrics can be practically useful in preventing the recurrence of historical biases because (even true) predictive labels are often influenced by unconscious biases.


{\bf Quantitative parity} is defined by the following equation:
\begin{equation}
|\{\hat{y}=1 \land S=s_0\}| = |\{\hat{y}=1 \land S=s_1\}|
\end{equation}
This metric aims for an equal number of positive predicates (i.e., $\hat{y}=1$) across the sensitive attribute groups ($S=s_0$ and $S=s_1$),  irrespective of the groups' population sizes.
While simple, this metric is not commonly examined in existing fairness studies.

{\bf Demographic parity} is defined by the following equation:
\begin{equation}\label{eq:DP}
P(\hat{y}=1|S=s_0) = P(\hat{y}=1|S=s_1)
\end{equation}
This aims to ensure equal probability ratios of positive predicates across different sensitive attribute groups.
Thus, in demographic parity, the number of positive predicates on $s_1$ (resp. $s_2$) increases as the number of candidates in $s_1$ (resp. $s_2$). 
If the number of candidates in $s_1$ and $s_2$ is the same, demographic parity is equivalent to quantitative parity.


Equal opportunity and equalized odds focus on the same accuracy between the two groups, which aim to accurately predict positive labels from both groups equally.
These metrics are useful if we focus only on data samples with positive labels in the groups rather than the size of data samples of groups.

{\bf Equal opportunity} aims to output the same true positive rate for each sensitive attribute group. 
\begin{equation}\label{eq:TPR}
P(\hat{y}=1|y=1, S=s_0) = P(\hat{y}=1|y=1,S=s_1)
\end{equation}
Since equal opportunity does not consider data samples with negative labels, if negative labels are correctly/wrongly predicated, it is not affected to equal opportunity.

{\bf Equalized odds} aims to satisfy both the true positive rate equality of equal opportunity (Equation~\ref{eq:TPR}) and the following condition:
\begin{equation}
P(\hat{y}=1|y=0, S=s_0) = P(\hat{y}=1|y=0,S=s_1)
\end{equation}
Equalized odds satisfies the same true positive and false positive rates.
When a machine learning model satisfies equalized odds, it inherently complies with the criteria for equal opportunity as well.


\subsection{Decision-making Scenarios}
\label{ssec:scenarios}

We briefly explain our three scenarios.

\noindent
{\bf Hiring} scenario assumes how we fairly hire male and female candidates.
The hiring manager's goal is to make an offer to the applicant with the highest net sales after one year of employment.
$s_0$ and $s_1$ represent male and female, respectively.
$y=1$ and $y=0$ represent high and low sales after one year of employment, respectively.

\noindent
{\bf Art Project} scenario assumes how we fairly give awards in an art project to students whose parents are artists and non-artists.
$s_0$ and $s_1$ represent artist and non-artist parents, respectively.
$y=1$ represents a student who has created an excellent work of art solely through his/her efforts, and $y=0$ represents a student who created a less-than-excellent work of art or a student whose parent helped in any way.

\noindent
{\bf Employee award} scenario assumes how we fairly give awards to male and female employees. $s_0$ and $s_1$ represent male and female, respectively.
$y=1$ and $y=0$ represent high and not-high sales at the end of the fiscal year, respectively.

\subsection{Participants}

We conducted the distribution of our questionnaires and the collection of responses through a professional marketing company. 
Participants who were registered users of the company's service completed the online questionnaires with wages that followed the legal standards of their respective countries. 
If participants do not agree to our consent form, the questionnaire is terminated immediately.

As our study did not specify the selection of individual participants, we have no access to their personal information (e.g., names and addresses). 
The marketing company closed their information and wages even for us.\footnote{We note that it is difficult to reproduce our results due to no personal information, which is similar to other survey studies.} 
All procedures were carried out in compliance with the ethical guidelines of our organization's ethics committee and legal advice from lawyers.

The responses were collected from China, France, Japan, and the United States from 1,000 participants in each country (totaling 4,000 participants), representative of the population ratios of gender and age in these countries.
Notably, the questionnaire was available in multiple languages; Chinese, French, Japanese, and English, thereby minimizing potential language barriers and ensuring comprehension among participants from different countries.



%% file: 4-result.tex
\input{table/participant}

\section{Result}
\label{sec:result}

This section presents the statistical results derived from our questionnaires.
We first provide an overview of the participant demographics, and then the distribution of responses for each question. 
Finally, we show the influence of various personal attributes on the choice of fairness metrics. 
We mainly show the results of the hiring scenario, and the results of the other two scenarios are in the appendix.

\subsection{Participants Demographics}
Table~\ref{tab:demographics_age_gender_eth_rel} shows the demographics of participants.
We collected participants to follow demographics of age/gender in each country, while other personal properties may not follow the actual demographics due to sampling biases.
For example, in China, the number of participants who have a bachelor's and above education is quite high compared to national statistics\footnote{\url{https://data.oecd.org/eduatt/adult-education-level.htm}}. We will discuss such sampling biases as the limitation.

This table also shows the number of participants who correctly answered at least five quizzes to check metric understanding. China and Japan have around 500 participants while France and the US only have 240 and 284 participants, respectively.
Since the number of correct answers may have significant differences, we analyze the difference between them.



\subsection{Statistical test}
Participants answer the preferences of fairness metrics in two ways: agreement levels and single choice.
Table \ref{tab:pvalue} shows p-values that indicate the significant difference. 
We conducted a one-way analysis of variance (ANOVA) test~\cite{blanca2017non} to check the significant difference in agreement levels between four countries, and a chi-square test for choices of fairness metrics to check the significant difference in countries and personal attributes.
We first describe the correctness (the bottom of Table~\ref{tab:pvalue}), which is the significant difference between the numbers of corrected answers for quizzes to check their understanding.
Since this shows significant differences, choices are statistically different across the numbers of correct answers, that is the responses from participants who do not understand correctly may not be reliable. 
Thus, we only used participants who correctly answered at least five quizzes among eight (totally, 1532 participants). 

In terms of agreement levels, small p-values indicate that the agreement levels of the reality of scenarios and fairness metrics are statistically different between countries.
We can see that agreement levels between countries are statistically different in all fairness metrics.

In terms of single choices, we can see each country has different confirmatory results across scenarios.
In personal attributes, Japan and the US do not have large significant differences between genders compared to other profiles; on the other hand, China and France do not have large significant differences between ages.
Each scenario also has different trends in each country: For example, in the ethnicity of Asia, the art project has a small p-value while the employee award has a large p-value.
These also indicate that countries and areas affect the preference of fairness metrics depending on scenarios.

In further results, we mark *, **, and *** if p-values are less than 0.5, 0.1, and 0.01, respectively.

\input{table/pvalues}

\subsection{Agreement levels}

Figure~\ref{fig:hiring_boxplot} shows the distribution of scores for questions assessing participants' agreement or disagreement with the reality of the scenario and the applicability of various fairness metrics in the hiring scenario. 
Recall that lower scores reflect higher levels of agreement.

From this figure, we can see that the hiring scenario is averagely agreed in all countries.
Regarding the fairness metrics, most definitions were accepted as appropriate for use in the scenario. However, it is noteworthy that quantitative parity received slightly more disagreement in China, Japan, and the US compared to other metrics. In contrast, in France, there was not a significant variance in agreement levels across the different fairness metrics.
 

\input{image/tex/scores_hiring_high}

\subsection{Correlations between personal attributes and the choice of fairness metrics}

Each participant chose one of the fairness metrics that they considered the most appropriate in each scenario. 
We provide the ratio of selected fairness metrics in different personal attributes; country, gender, religion, ethnicity, correctness, age, education, and experience.

 \begin{figure*}[!t]
 \centering
    \begin{minipage}[t]{0.32\linewidth}
        \centering
        \includegraphics[width=1.0\linewidth]{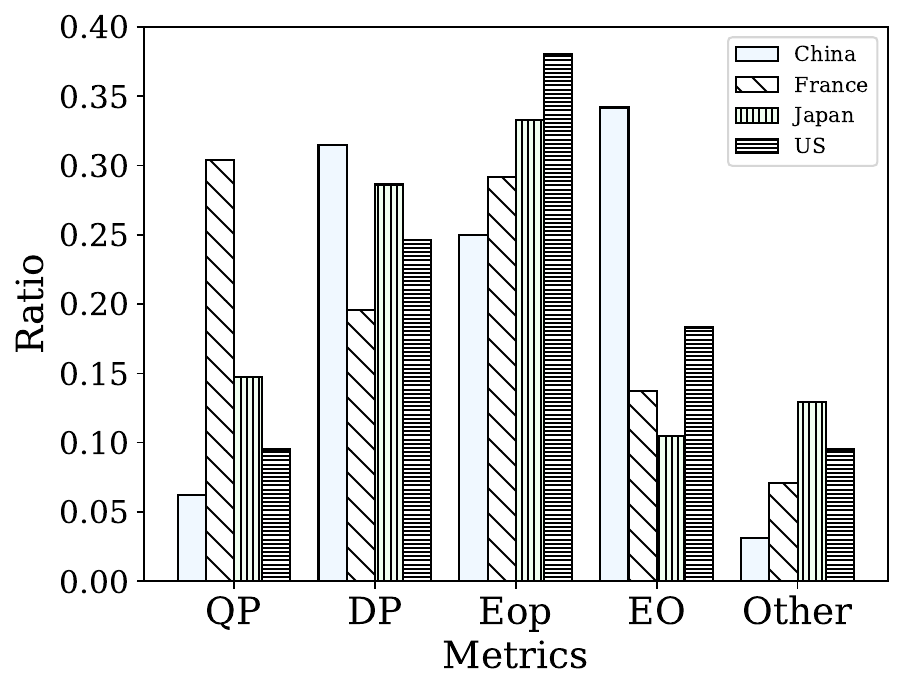}
       \subcaption{Hiring***}
       \label{fig:hiring_country}
    \end{minipage}
    \begin{minipage}[t]{0.32\linewidth}
        \centering
        \includegraphics[width=1.0\linewidth]{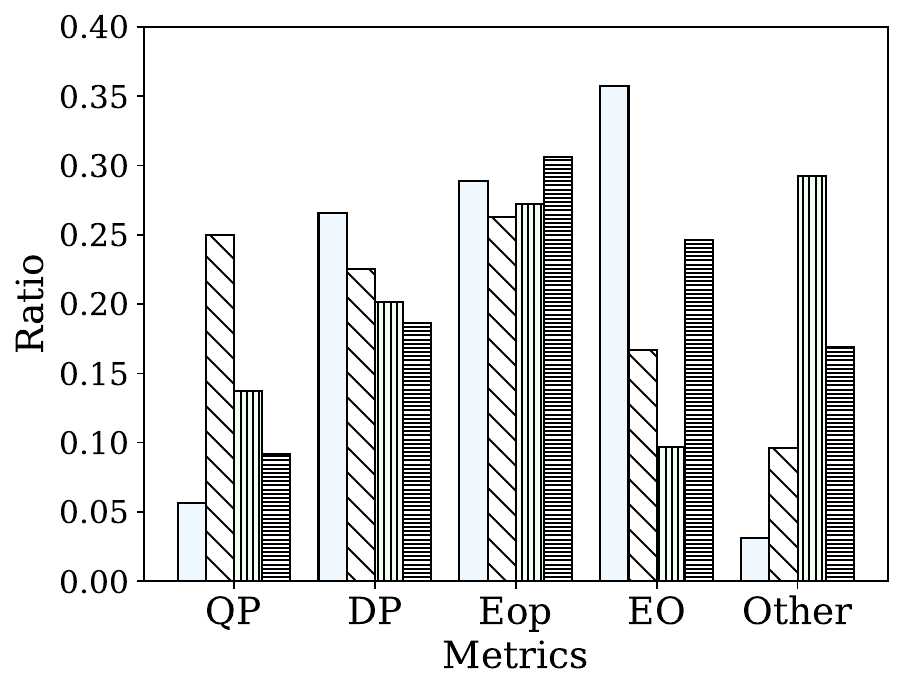}
        \subcaption{Art project***}
        \label{fig:art_country}
    \end{minipage}
    \begin{minipage}[t]{0.32\linewidth}
        \centering
        \includegraphics[width=1.0\linewidth]{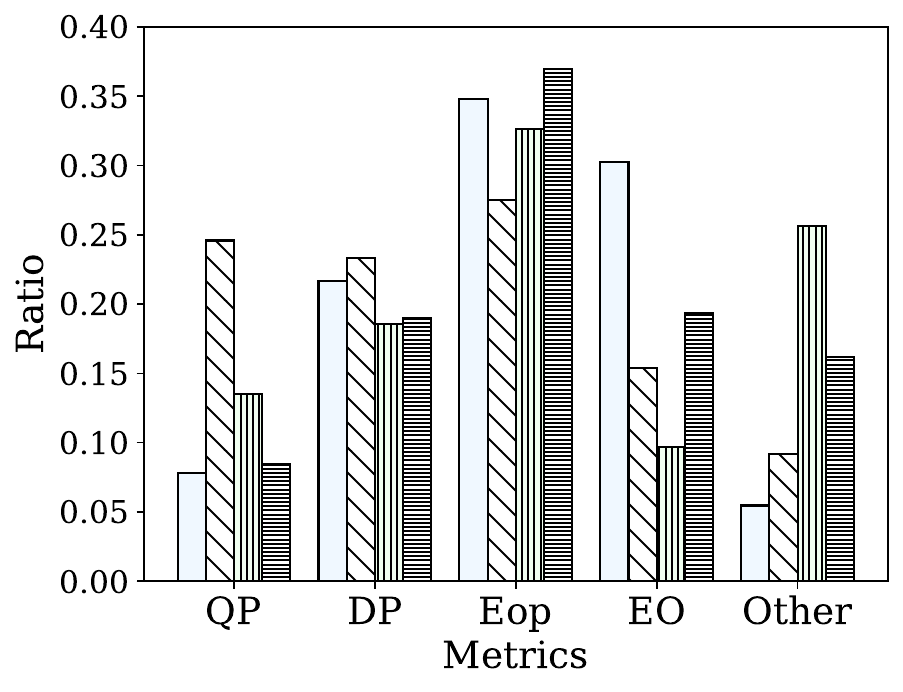}
        \subcaption{Employee award***}
        \label{fig:employee_country}
    \end{minipage}
\vspace{-2 mm}
\caption{Difference between countries.}
\vspace{-2 mm}
\label{fig:country}
\end{figure*}


\smallskip
\noindent
{\bf Difference between countries.}
Figure~\ref{fig:country} shows the difference between countries.
The ratios are quite different across countries.
We observe that equal opportunity is often selected in all countries. 
France prefers quantitative parity compared to other countries.
Since past studies did not use quantitative parity, it may be preferred in other countries and scenarios.
Interestingly, Japan and US have similar trends even though they have different cultures (i.e., different areas, ethnicities, and religions), while similar areas (i.e., China and Japan, France and US) have different trends. This indicates that it may not be suitable using the same metrics even if similar countries use them. 

In China, the selected metrics are largely different across the scenarios, while Japan and US are almost constant.
This indicates whether we may need to change the fairness metrics if scenarios change depending on countries.

In Japan, ``Other'' is often selected in the art project scenario.
The main difference between art project and other scenarios is that the sensitive attributes are related to either parents or genders. 
Candidates mentioned ``the award for students should not be considered their parents'' in comments of reasons why they selected ``Other''. Thus, they suppose that fairness metrics should not be used in the scenario.



In Japan and the US, equal opportunity is mostly selected, and in China, the selected one depends on the scenarios.
They may think that group fairness should consider predictive accuracy.

 \begin{figure*}[!t]
 \centering
    \begin{minipage}[t]{0.24\linewidth}
        \centering
        \includegraphics[width=1.0\linewidth]{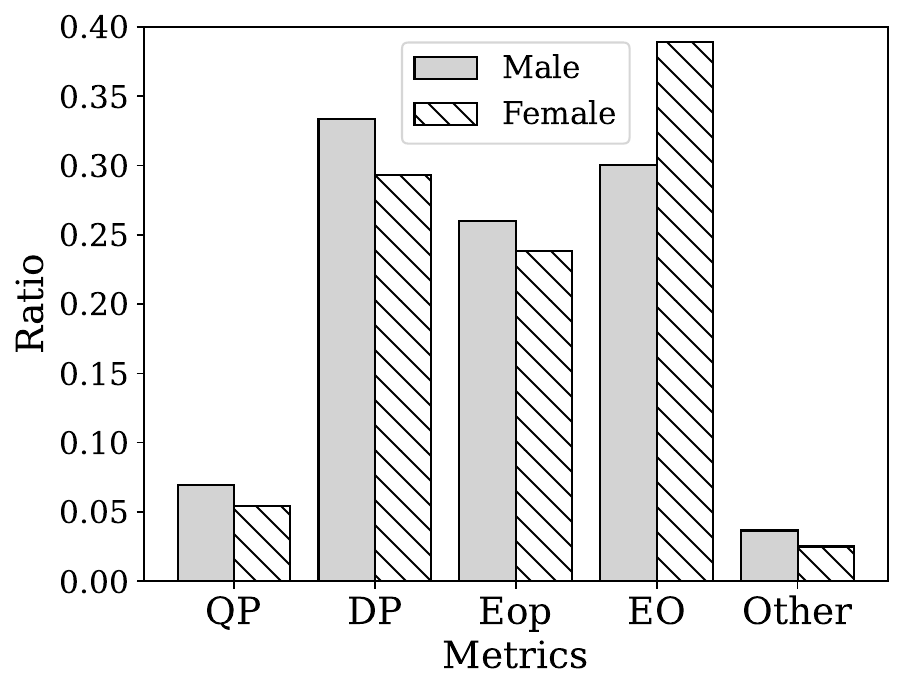}
       \subcaption{China**}
       \label{fig:hiring_boxplot_china}
    \end{minipage}
    \begin{minipage}[t]{0.24\linewidth}
        \centering
        \includegraphics[width=1.0\linewidth]{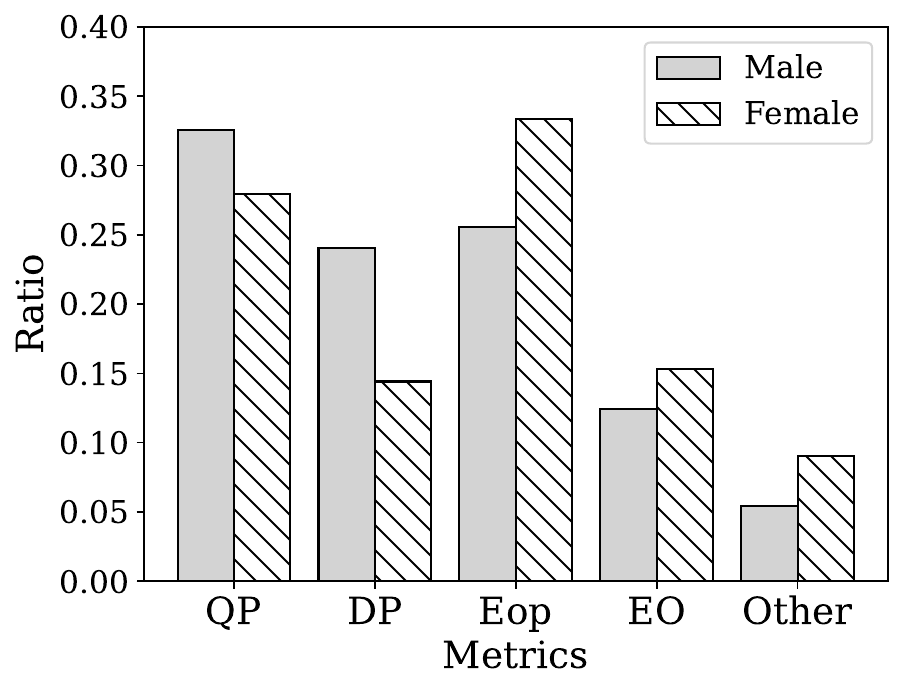}
        \subcaption{France**}
        \label{fig:hiring_boxplot_japan}
    \end{minipage}
    \begin{minipage}[t]{0.24\linewidth}
        \centering
        \includegraphics[width=1.0\linewidth]{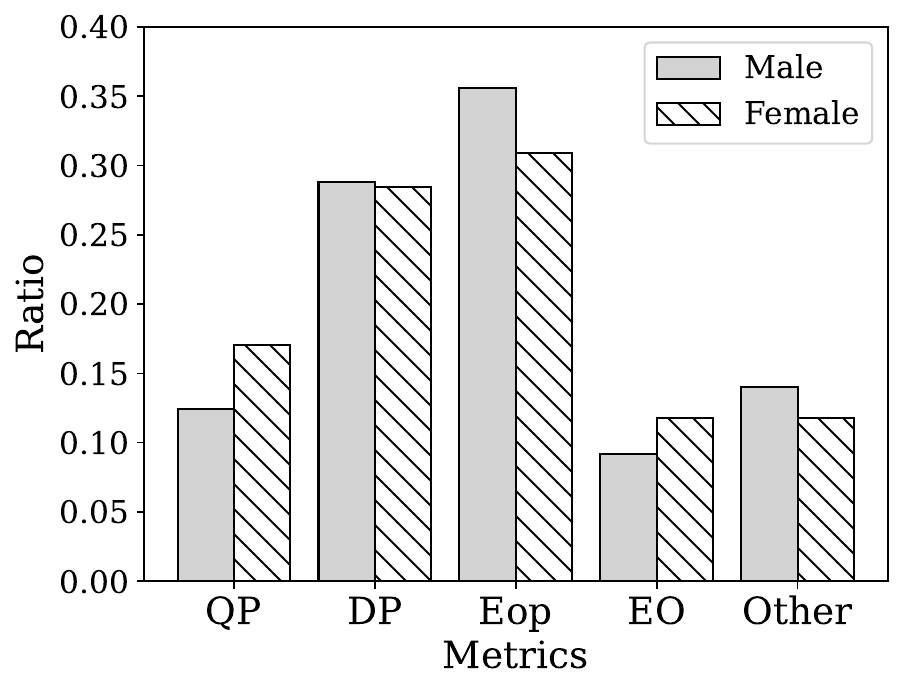}
        \subcaption{Japan*}
        \label{fig:hiring_boxplot_france}
    \end{minipage}
    \begin{minipage}[t]{0.24\linewidth}
    \centering
    \includegraphics[width=1.0\linewidth]{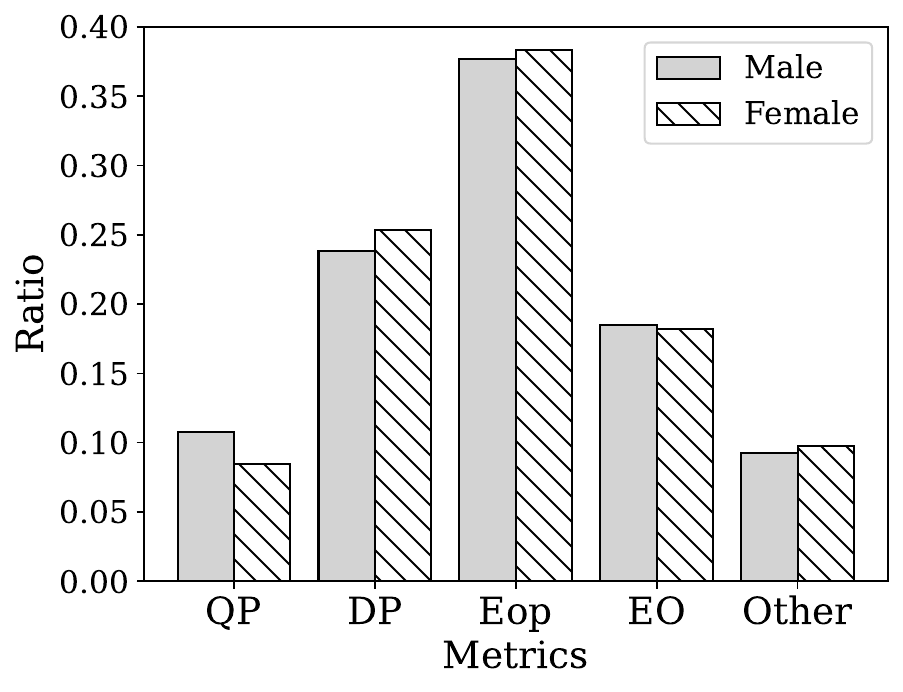}
    \subcaption{US}
    \label{fig:hiring_boxplot_us}
    \end{minipage}
\vspace{-3 mm}
\caption{Difference between genders in countries in hiring scenario}
\vspace{-3 mm}
\label{fig:hiring_gender_county}
\end{figure*}

\smallskip
\noindent
{\bf Difference between genders.}
\label{sssec:genders}
We analyze the difference between choices of fairness metrics in genders in each country.
Figure~\ref{fig:hiring_gender_county} shows the result of the hiring scenario.
The differences are not substantial, in particular, the US.
However, there are certain trends: males often prefer demographic parity more than females, and females prefer equalized odds more than males.
Interestingly, France has the largest difference, while it has the best gender gap index among four countries.
This analysis shows that there is no big gap between genders, but they have some characteristics between genders in different countries and the similarity of the choice fairness metrics between genders may not follow the gender gap index.

 \begin{figure*}[!t]
 \centering
    \begin{minipage}[t]{0.32\linewidth}
        \centering
        \includegraphics[width=1.0\linewidth]{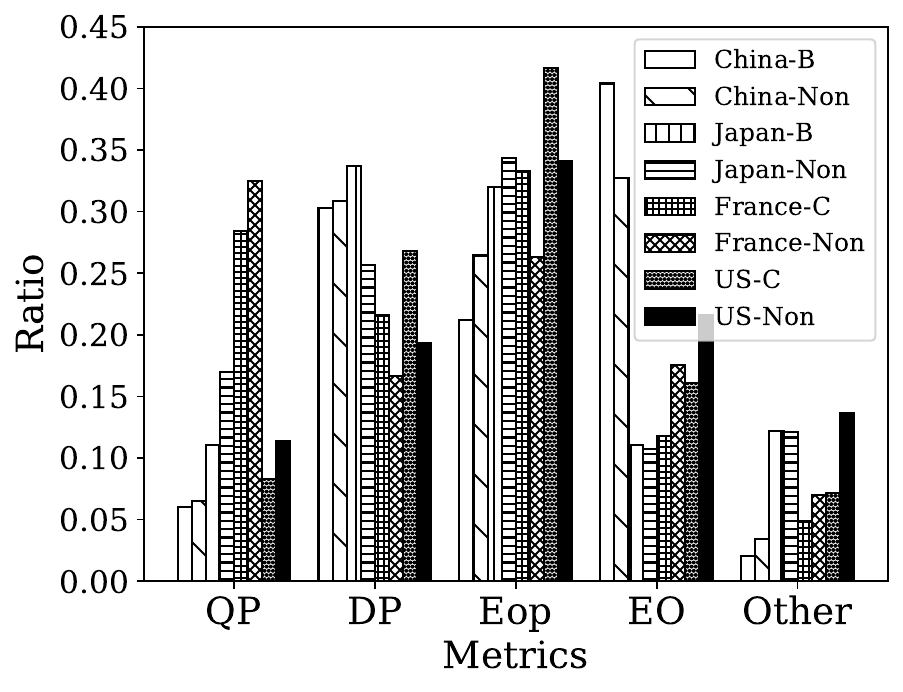}
       \subcaption{Religions: France*, Japan*, and US*}
    \end{minipage}
\begin{minipage}[t]{0.32\linewidth}
        \centering
        \includegraphics[width=1.0\linewidth]{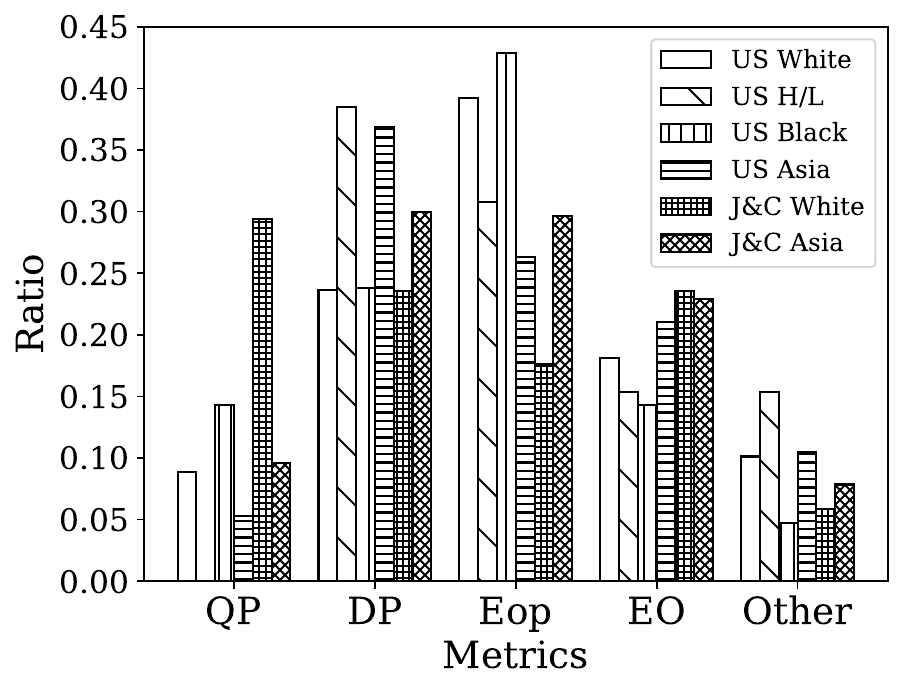}
       \subcaption{Ethnicity: Asia**}
    \end{minipage}
    \begin{minipage}[t]{0.32\linewidth}
        \centering
        \includegraphics[width=1.0\linewidth]{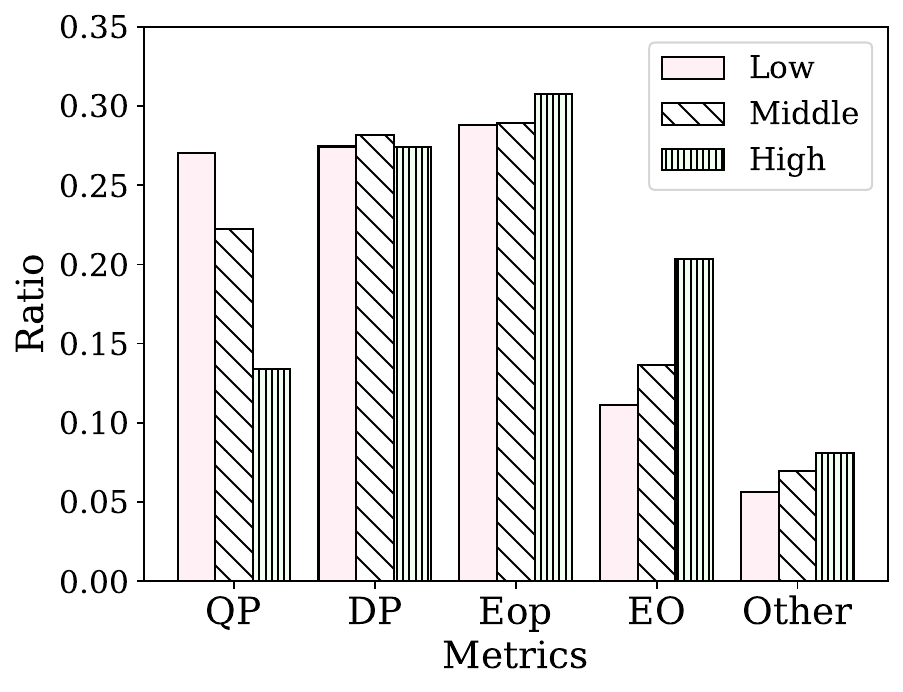}
       \subcaption{Correctness***}
    \end{minipage}
   \vspace{-2 mm}
\caption{Difference between religion/ethiniticy; B, C, and Non indicate Buddhism, Christianity, and no religion, respectively.}
\vspace{-3 mm}
\label{fig:religion}
\end{figure*}

\smallskip
\noindent
{\bf Difference between religions.}
Figure~\ref{fig:religion} (a) shows the difference between religions. We compared people affiliated with the major and no religions in each country.
The results reveal only minor discrepancies between major and no religions in all countries. 
This observation may imply that cultural influences from major religions might impact even those who do not actively practice the religion. 
For instance, in Japan, despite a high proportion of no religions, Buddhist practices significantly influence cultures, such as seasonal events and funerals.

Furthermore, when comparing the preferences of participants from different countries who share the same major religions, we observed notable dissimilarities. These differences also show the stronger influence of national context than religious affiliation. This shows the impact of country-specific factors is larger than religious influences in the choice of fairness metrics.

\smallskip
\noindent
{\bf Difference between ethnicity.}
Figure~\ref{fig:religion} (b) shows the difference between ethnicity. We compared each ethnicity in the US and Asia. Recall that we did not collect ethnicity in France due to legal issues.
We removed Black and H/L in Asia and NA/AN from analysis due to their small numbers.

From these results, we can see that even if ethnicity is the same, their choices are different between countries.
In the US, there are no large differences between White, H/L, and Black, while Asian has different trends.
In Japan and China, White often selects quantitative parity and Asian often selects demographic parity and equal opportunity.
Therefore, we can consider that the current (and historical) situations of each ethnicity influence the choice of fairness metrics.

\smallskip
\noindent
{\bf Difference between correctness.}
Figure~\ref{fig:religion} (c) shows the difference between correctness. 
We divide the participants based on the number of correct answers; low, middle, and high indicate that the number of correct answers are between 0 and 3, 4 and 5, and 6 and 8, respectively. 
Lower correctness selects more quantitative parities, which might be caused by the random selection.
Therefore, we only focus on high correctness candidates.

 \begin{figure*}[!t]
 \centering
    \begin{minipage}[t]{0.24\linewidth}
        \centering
        \includegraphics[width=1.0\linewidth]{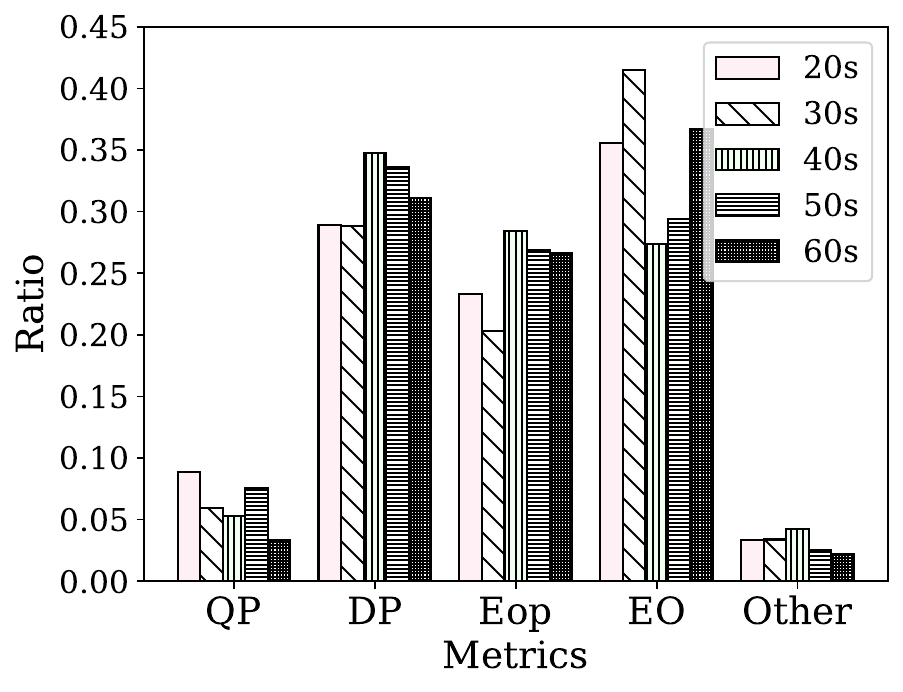}
       \subcaption{China}
       \label{fig:hiring_boxplot_china}
    \end{minipage}
    \begin{minipage}[t]{0.24\linewidth}
        \centering
        \includegraphics[width=1.0\linewidth]{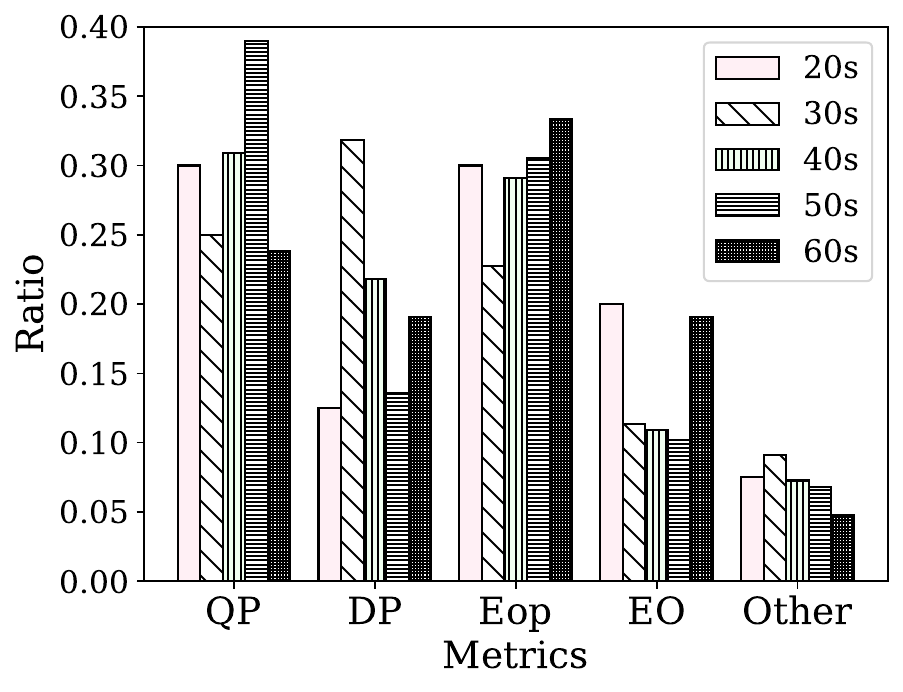}
        \subcaption{France}
        \label{fig:hiring_boxplot_japan}
    \end{minipage}
    \begin{minipage}[t]{0.24\linewidth}
        \centering
        \includegraphics[width=1.0\linewidth]{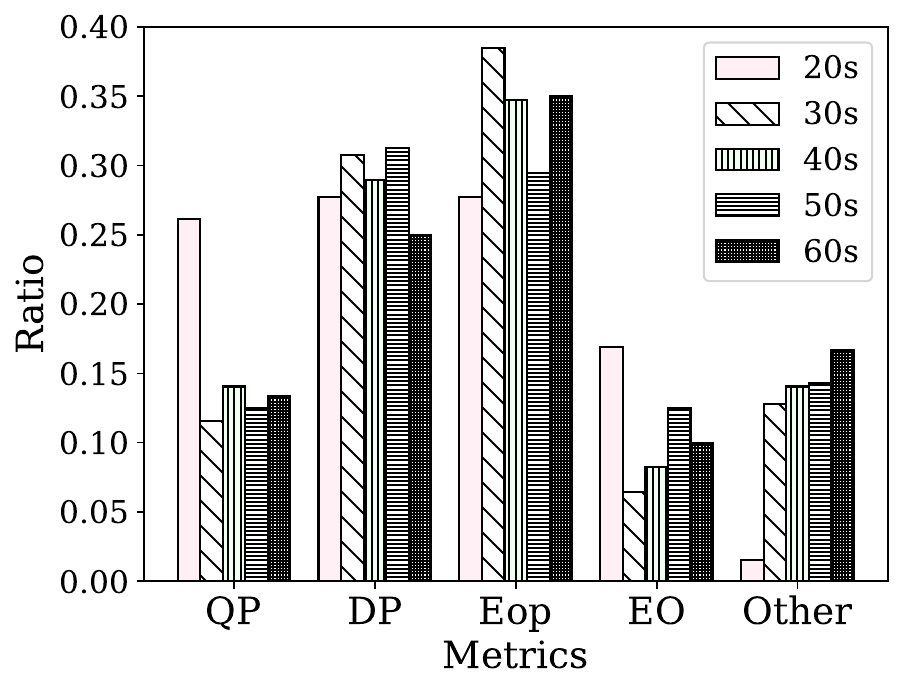}
        \subcaption{Japan*}
        \label{fig:hiring_boxplot_france}
    \end{minipage}
    \begin{minipage}[t]{0.24\linewidth}
    \centering
    \includegraphics[width=1.0\linewidth]{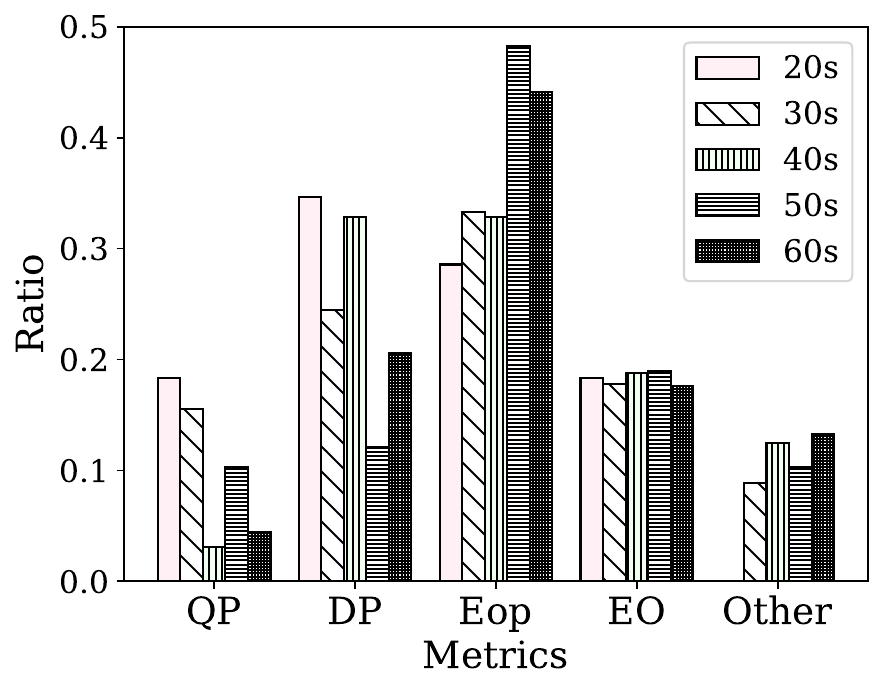}
    \subcaption{US***}
    \label{fig:hiring_boxplot_us}
    \end{minipage}
\vspace{-3 mm}
\caption{Difference between age in countries in hiring scenario}
\vspace{-3 mm}
\label{fig:hiring_age_county}
\end{figure*}



\smallskip
\noindent
{\bf Difference between ages.}
Figure~\ref{fig:hiring_age_county} shows the difference between ages. 
Each country has different trends.
There are three characteristics; 20s often select quantitative parity than others, 30s and 40s select demographic parity, and 50s and 60s select predicted parity (i.e., equal opportunity or equalized odds).
First, in terms of the 20s, it might increase the awareness for the unproportional quota of affirmative action in the hiring.
Second, in terms of 30s and 40s, they are often large numbers among ages (see Table~\ref{tab:demographics_age_gender_eth_rel}), so they may select the advantageous fairness metrics for them.
Finally, 50s and 60s might consider that the performance of candidates might be taken into account for fair hiring.

\begin{figure*}[!t]
 \centering
    \begin{minipage}[t]{0.24\linewidth}
        \centering
        \includegraphics[width=1.0\linewidth]{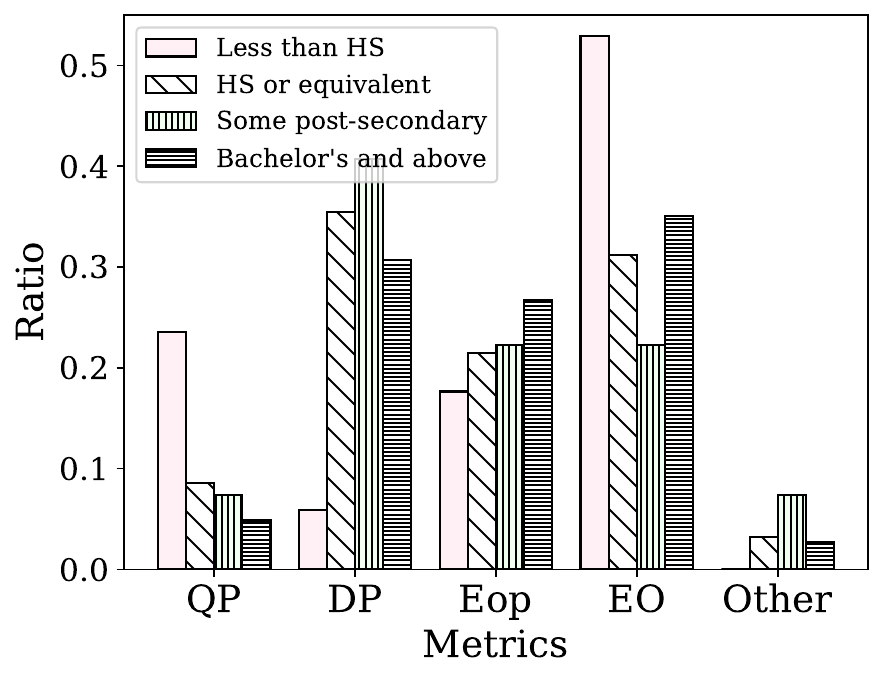}
       \subcaption{China***}
    \end{minipage}
    \begin{minipage}[t]{0.24\linewidth}
        \centering
        \includegraphics[width=1.0\linewidth]{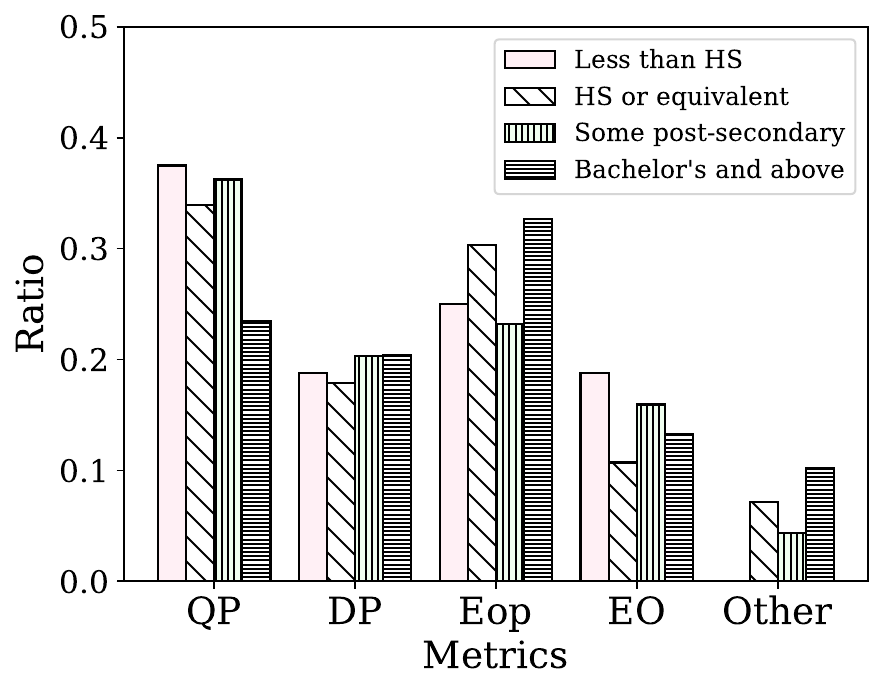}
        \subcaption{France}
    \end{minipage}
    \begin{minipage}[t]{0.24\linewidth}
        \centering
        \includegraphics[width=1.0\linewidth]{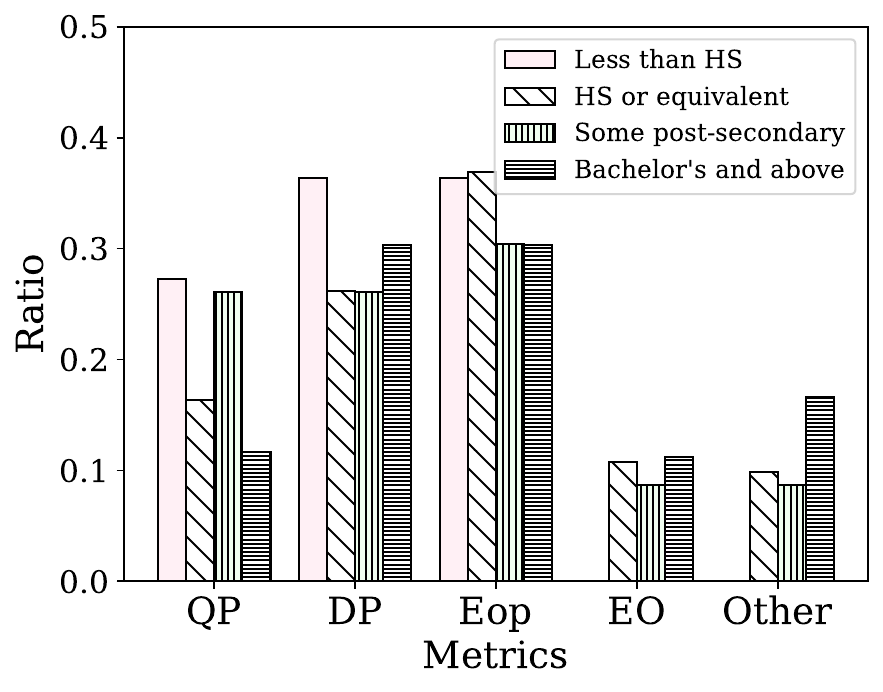}
        \subcaption{Japan*}
    \end{minipage}
    \begin{minipage}[t]{0.24\linewidth}
    \centering
    \includegraphics[width=1.0\linewidth]{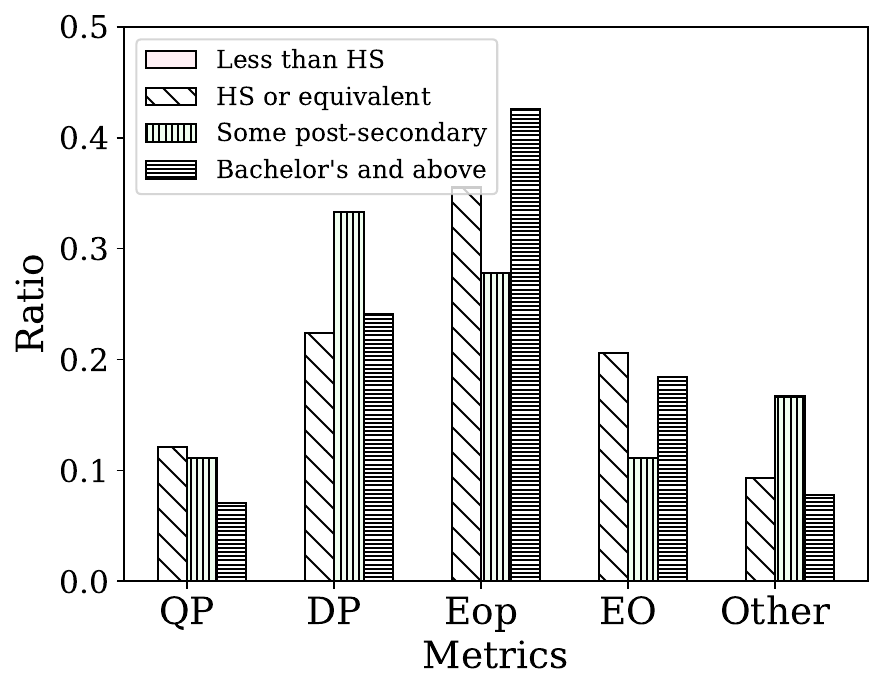}
    \subcaption{US*}
    \end{minipage}
  \vspace{-2 mm}
\caption{Difference between education in countries in hiring scenario. Note that no ``Less than HS'' in US who correctly answered at least five quizzes.}
\vspace{-2 mm}
\label{fig:hiring_education_county}
\end{figure*}

\smallskip
\noindent
{\bf Difference between educations.}
Figure~\ref{fig:hiring_education_county} shows the difference between education. 
Notably, ``less than HS'' and ``some post-secondary'' often select quantitative parity, while ``Bachelor's and above'' often selects equal opportunity.
A bachelor's degree often requires many cases such as hiring.
It may affect the choice of fairness metrics in their educations.

\begin{figure*}[!t]
 \centering
    \begin{minipage}[t]{0.24\linewidth}
        \centering
        \includegraphics[width=1.0\linewidth]{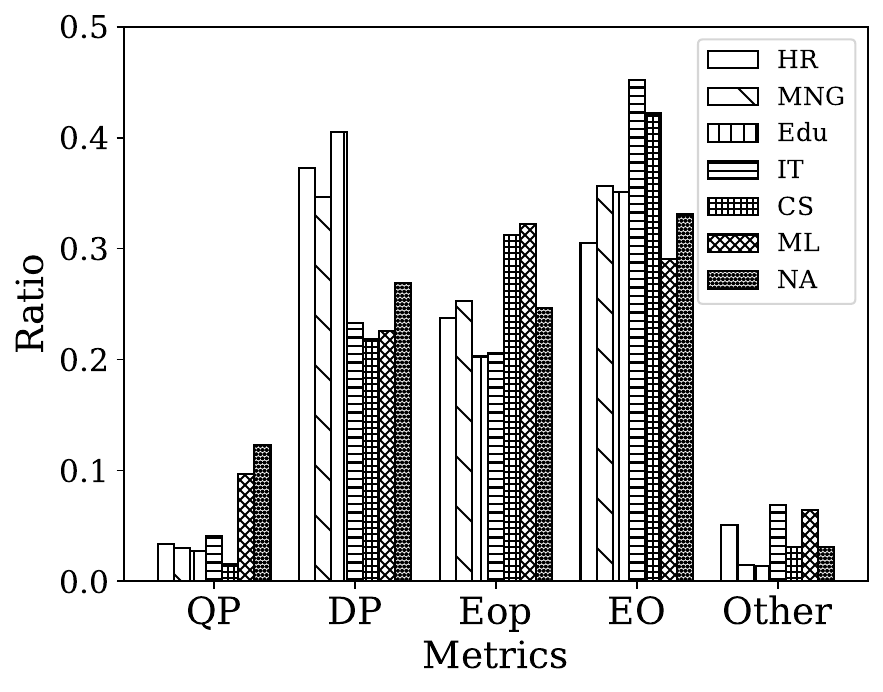}
       \subcaption{China***}
    \end{minipage}
    \begin{minipage}[t]{0.24\linewidth}
        \centering
        \includegraphics[width=1.0\linewidth]{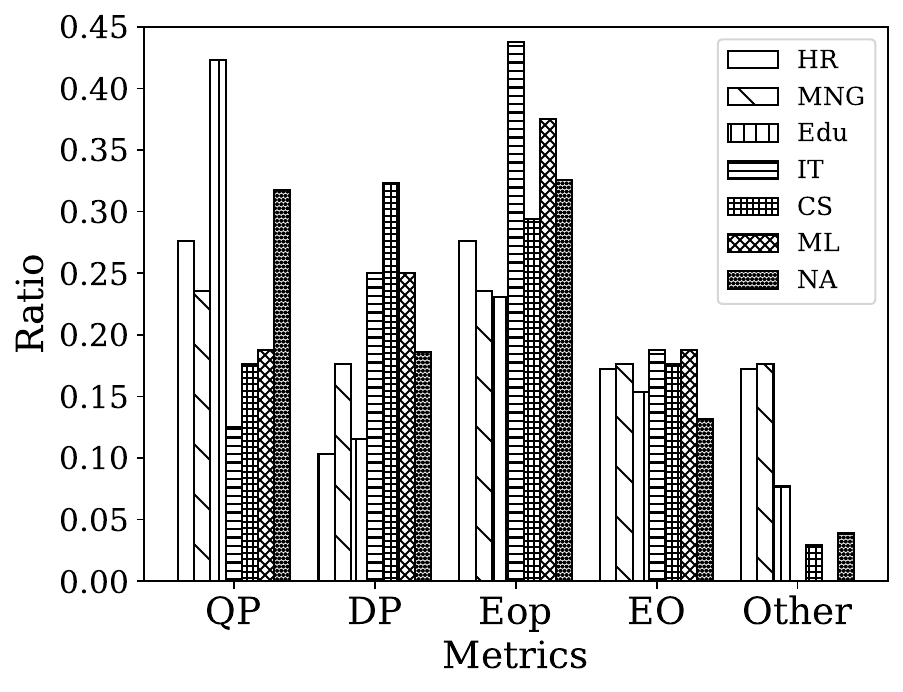}
        \subcaption{France*}
    \end{minipage}
    \begin{minipage}[t]{0.24\linewidth}
        \centering
        \includegraphics[width=1.0\linewidth]{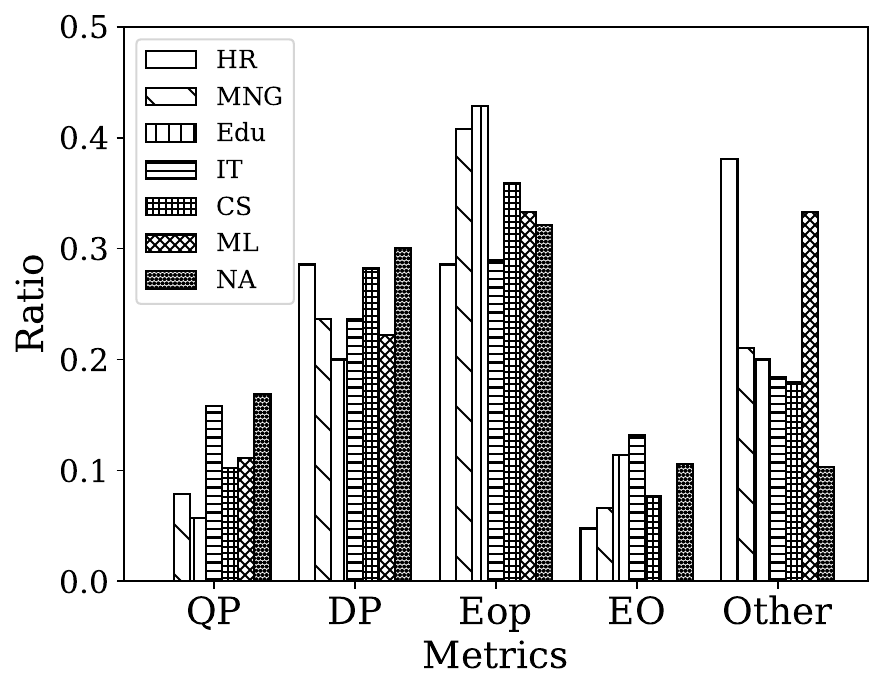}
        \subcaption{Japan***}
    \end{minipage}
    \begin{minipage}[t]{0.24\linewidth}
    \centering
    \includegraphics[width=1.0\linewidth]{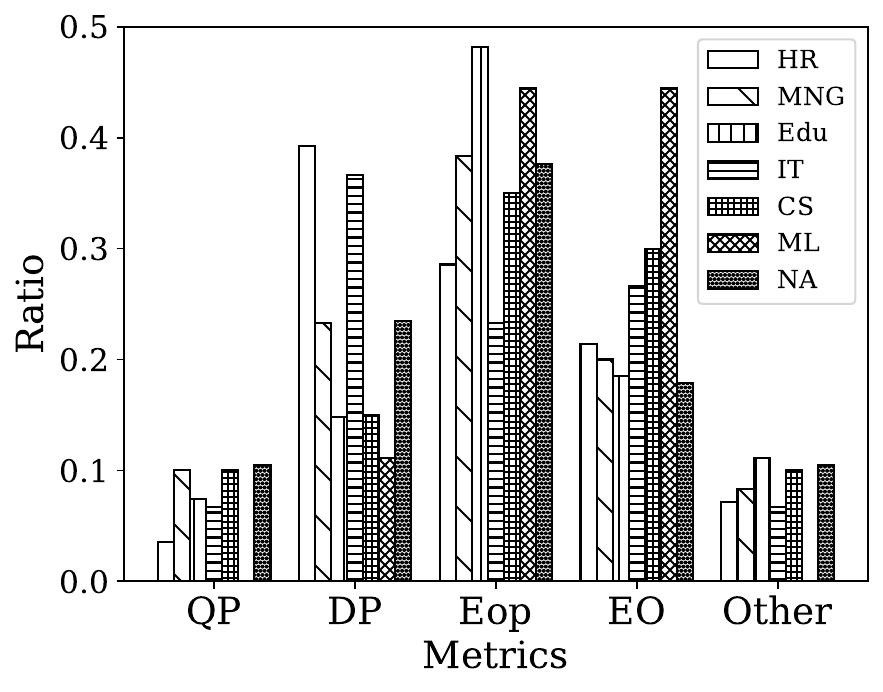}
    \subcaption{US}
    \end{minipage}
   \vspace{-2 mm}
\caption{Difference between experience in countries in hiring scenario}
\vspace{-3 mm}
\label{fig:hiring_experience_county}
\end{figure*}

\smallskip
\noindent
{\bf Difference between experiences.}
Figure~\ref{fig:hiring_experience_county} shows the difference between experiences. 
There are no clear trends among experiences.
Candidates with non-applicable slightly prefer quantitative parity compared to others.
Since many candidates belong to non-applicable, it may be better to divide this group into fine groups in order to understand their trends.

%% file: table/participant.tex
\begin{table}[ttt]

\centering
\caption{The number of participants in each country. The numbers in parentheses indicate how many participants correctly answered at least five quizzes to check metrics understanding.
AA, NA, AN, NH, and PI indicate African American, Native American, Alaska native, Native Hawaiian, and Pacific Islander, respectively.
}
\label{tab:demographics_age_gender_eth_rel}
\begin{minipage}{0.45\textwidth}
\centering
\scalebox{0.75}{
\begin{tabular}{lrrrr}\hline 
 & \multicolumn{4}{c}{{\bf Number of samples}} \\
 & \multicolumn{1}{c}{{\bf China}}& \multicolumn{1}{c}{{\bf France}}  & \multicolumn{1}{c}{{\bf Japan}}   & \multicolumn{1}{c}{{\bf US}} \\\hline
\multicolumn{1}{c}{{\bf Total}} & 1,000 (512) & 1,000 (240) & 1,000 (496)& 1,000 (284) \\
\multicolumn{1}{c}{{\bf Age/Gender}} & \multicolumn{4}{c}{} \\
20s/Male & 93 (51)& 99 (20) & 81 (28) & 107 (23)\\
20s/Female & 82 (65)& 94 (26) & 78 (39) & 102 (18)\\
30s/Male & 120 (48)& 99 (28)& 92 (62) & 110 (24)\\
30s/Female & 113 (59)& 96 (32) & 92 (60) & 107 (31)\\
40s/Male & 102 (50)& 101 (23)& 121 (61) & 96 (34)\\
40s/Female & 99 (39)& 98 (20)& 117 (37) & 97 (26)\\
50s/Male & 120 (53)& 107 (18)& 107 (39) & 94 (27)\\
50s/Female & 118 (47)& 109 (27)& 108 (59) & 97 (40)\\
60s/Male & 76 (60)& 94 (27) & 99 (52) & 91 (27)\\
60s/Female & 77 (40)& 103 (19) & 105 (59) & 99 (34) \\\hline
\multicolumn{1}{c}{{\bf Ethnicity}} & \multicolumn{4}{c}{}\\
White & 17 (5) &--- & 59 (12) & 718 (237) \\
Hispanic or Latinx & 0&--- & 1 (0) & 86 (13)\\
Black or AA &  1 (0)&--- & 1 (0)& 171 (21)\\
NA or AN  & 16 (0)&--- & 3 (0)& 21 (5) \\
Asian, NH or PI  & 940 (500)&--- & 850 (455) & 54 (19) \\
Other  & 41 (7)&--- & 66 (24) & 14 (4) \\\hline
\multicolumn{1}{c}{{\bf Religion}} & \multicolumn{4}{c}{}\\
Christianity & 37 (15) & 411 (102) & 19 (2) & 630 (168) \\
Islam & 7 (3)& 92 (15)& 1 (0) & 18 (3)\\
Hinduism &  8 (1)& 3 (0)& 1 (0) & 7 (2)\\
Buddhism  & 208 (99)& 9 (2)& 329 (172) & 11 (2)\\
No religion  & 700 (382)& 409 (114)& 548 (288) & 231 (88)\\
Other  & 30 (10) & 33 (2) & 57 (23) & 76 (19) \\\hline
\end{tabular}
}
\end{minipage}
\hspace{0.04\columnwidth}
\begin{minipage}{0.45\textwidth}
\centering
\scalebox{0.75}{
\begin{tabular}{lrrrr}\hline
 & \multicolumn{4}{c}{{\bf Number of samples}} \\
 & \multicolumn{1}{c}{{\bf China}}& \multicolumn{1}{c}{{\bf France}}  & \multicolumn{1}{c}{{\bf Japan}}   & \multicolumn{1}{c}{{\bf US}} \\\hline
\multicolumn{1}{c}{{\bf Education}} & \multicolumn{4}{c}{} \\
Less high & 35 (17) & 36 (16) & 146 (11) & 36 (0) \\
High and equivalent &208 (93) & 471 (56) & 292 (214)& 504 (107)\\
Some post-secondary &  60 (27)& 44 (69)& 280 (23) & 155 (36)\\
Bachelor's and above  & 687 (371)& 415 (98)& 269 (241) & 304 (141)\\
Other  & 10 (4) & 34 (1) & 13 (7) & 1 (0)\\\hline
\multicolumn{1}{c}{{\bf Experience}} & \multicolumn{4}{c}{} \\
Human resources & 131 (59)& 119 (29) & 58 (21)  & 91 (28)\\
Management & 418 (202)& 133 (34)& 92 (76) & 168 (60)\\
Education &  136 (74)& 112 (26)& 65 (35) & 82 (27)\\
IT infra  & 166 (73)& 68 (16)& 76 (38) & 88 (30)\\
Computer science  & 143 (64)& 129 (34)& 58 (39) & 60 (20)\\
Machine learning & 73 (31)& 49 (16) & 17 (9) & 50 (9) \\
Not applicable & 243 (130)& 544 (129) & 715 (349)  & 620 (162) \\\hline
\end{tabular}
}
\end{minipage}
\end{table}

%% file: table/pvalues.tex
\begin{table}[ttt]
\centering
\caption{Summary of p-values in  ANOVA for agreement levels and chi-square test for others. A smaller p-value indicates stronger significant differences. Bold fonts indicate that p-values are less than 0.1. 
We note that China and Japan have larger sample sizes than France and US, so their p-values tend to be smaller than those of France and US.}
\label{tab:pvalue}
\centering
\begin{tabular}{ll|ccc}\hline
 & &\multicolumn{1}{c}{{\bf Hiring}}& \multicolumn{1}{c}{{\bf Art project}}  & \multicolumn{1}{c}{{\bf Employee award}}  \\\hline
\multirow{5}{*}{{\bf Agreement levels}}&Scenario& {\bf 6.8e-51} & {\bf 3.9e-60} & {\bf 1.9e-51} \\
&QP& {\bf 5e-25} & {\bf 2.5e-25} & {\bf 1.3e-21} \\
&DP& {\bf 9.6e-21} & {\bf 4.1e-35} & {\bf 3.5e-24} \\
&Eop& {\bf 5.7e-25} & {\bf 1.3e-36} & {\bf 1.7e-37} \\
&EO& {\bf 1.7e-36} & {\bf 1.1e-52} & {\bf 1.9e-45} \\\hline
\multicolumn{2}{c|}{{\bf Country}} & {\bf 3.4e-37}& {\bf 2.8e-49}& {\bf 1.9e-32}\\\hline
\multirow{5}{*}{{\bf Gender}} &All& 0.44& {\bf 0.074}& {\bf 0.01}\\
&China& 0.30& {\bf 0.071}& {\bf 0.085}\\
&France& 0.21& 0.51& {\bf 0.068}\\
&Japan& 0.42& 0.24& 0.37\\
&US& 0.97& 0.80& 0.64\\\hline
\multirow{4}{*}{{\bf Religion}}&China& 0.61& 0.89& 0.56\\
&France& 0.48& 0.76& 0.99\\
&Japan& 0.27& 0.92& {\bf 0.067}\\
&US& 0.18& 0.27& 0.75\\\hline
\multirow{2}{*}{{\bf Ethnicity}} &Asia& 0.10& {\bf 0.027}& 0.94\\
&US& 0.88& 0.67& 0.18\\\hline
\multirow{5}{*}{{\bf Age}}&All& {\bf 0.0055}& 0.28& {\bf 0.057}\\
&China& 0.84& 0.73& {\bf 0.073}\\
&France& 0.70& 0.63& 0.69\\
&Japan& 0.12& {\bf 0.052}& 0.97\\
&US& {\bf 0.025}& 0.4& {\bf 0.081}\\\hline
\multirow{5}{*}{{\bf Education}}&All& {\bf 8.1e-07}& {\bf 3.5e-05}& {\bf 6.5e-05}\\
&China& {\bf 0.039}& 0.72& {\bf 0.031}\\
&France& 0.75& 0.12& 0.60\\
&Japan& 0.25& {\bf 0.043}& {\bf 0.074}\\
&US& 0.37& 0.96& 0.33\\\hline
\multirow{5}{*}{{\bf Experience}}&All& {\bf 2.7e-05}& {\bf 0.039}& 0.12\\
&China& {\bf 0.016}& 0.65& 0.62\\
&France& 0.17& 0.86& 0.95\\
&Japan& {\bf 0.063}& {\bf 0.00082}& {\bf 0.08}\\
&US& 0.77& 0.41& 0.92\\\hline
\multirow{5}{*}{{\bf Correctness}}&All& {\bf 3e-22}& {\bf 4.3e-26}& {\bf 3.9e-18}\\
&China& {\bf 3.8e-10}& {\bf 3.5e-11}& {\bf 0.0011}\\
&France& {\bf 0.00077}& {\bf 0.033}& 0.25\\
&Japan& {\bf 0.004}& {\bf 6.4e-05}& {\bf 0.00012}\\
&US& {\bf 0.0003}& {\bf 1.5e-05}& {\bf 0.0017}\\

\hline
\end{tabular}
\vspace{-3mm}
\end{table}

%% file: image/tex/scores_hiring_high.tex
 \begin{figure*}[!t]
 \centering
    \begin{minipage}[t]{0.24\linewidth}
        \centering
        \includegraphics[width=1.0\linewidth]{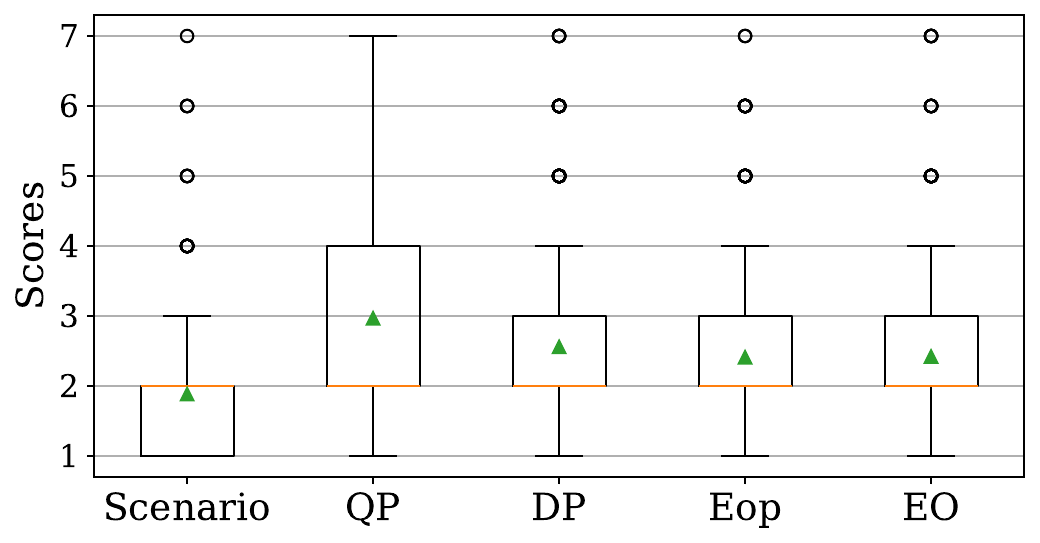}
       \subcaption{China}
       \label{fig:hiring_boxplot_china}
    \end{minipage}
    \begin{minipage}[t]{0.24\linewidth}
        \centering
        \includegraphics[width=1.0\linewidth]{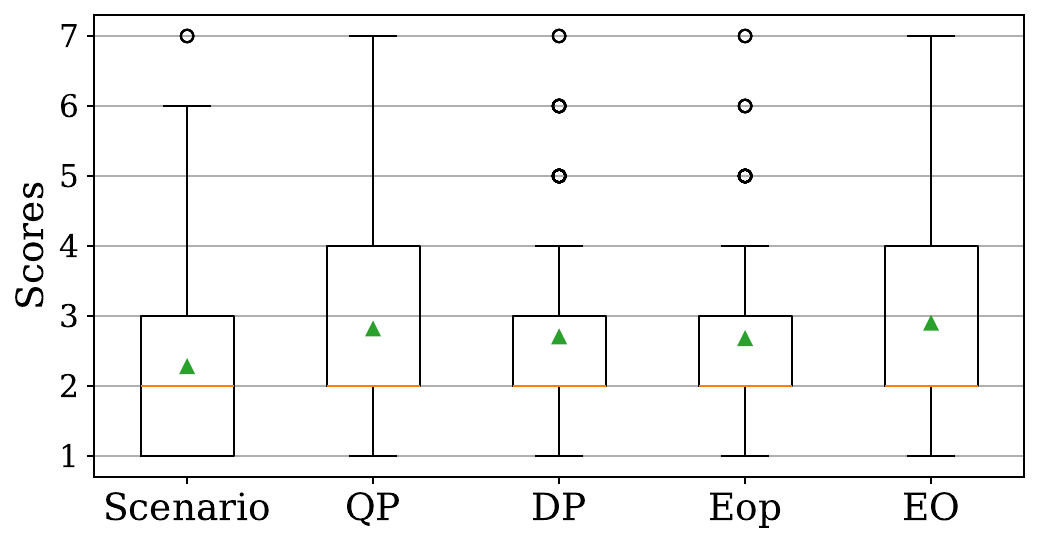}
        \subcaption{France}
        \label{fig:hiring_boxplot_japan}
    \end{minipage}
    \begin{minipage}[t]{0.24\linewidth}
        \centering
        \includegraphics[width=1.0\linewidth]{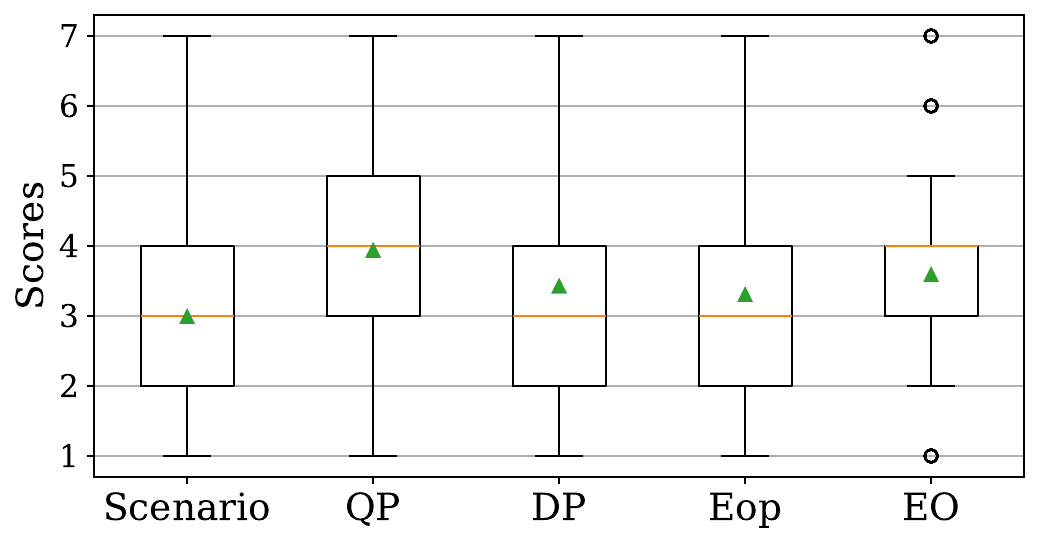}
        \subcaption{Japan}
        \label{fig:hiring_boxplot_france}
    \end{minipage}
    \begin{minipage}[t]{0.24\linewidth}
    \centering
    \includegraphics[width=1.0\linewidth]{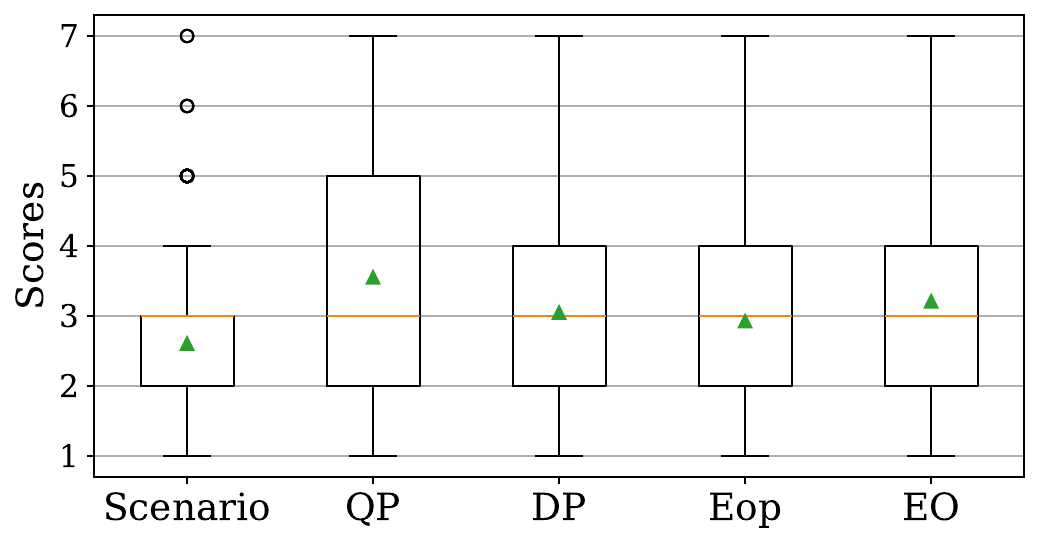}
    \subcaption{US}
    \label{fig:hiring_boxplot_us}
    \end{minipage}
\vspace{-1.6 mm}
\caption{Scores of answers for hiring scenario}
 \vspace{-3 mm}
\label{fig:hiring_boxplot}
\end{figure*}

%% file: 5-discussion.tex
\section{Discussion}
\label{sec:discussion}


\subsection{Cross-country valuations}
In our results, countries have the most impact on the choices of fairness metrics. 
This trend may be caused by varying political climates and societal situations in different countries.
For example, in France, public perception might be affected by universalism, which broadly refers to the concept of all members being equal under a state formed since the Declaration of Human Rights. Universalism closely aligns with the idea of quantitative parity~\cite{bereni2007french}.
In contrast, in most scenarios, equal opportunity in China, Japan, and the US are higher than other metrics. 
These metrics are predictive parties that are related to predictive accuracy.  
They may consider that selected and non-selected people should be based on their performance in each attribute. 
Religion does not have a large impact compared to countries, while ethnicity in different countries has an impact on the choice of fairness metrics. The political climate and societal situations regarding the ethnicity of each country may also cause this.

We assumed regions with strong correlations to specific ethnicities and religions might exhibit similar trends in fairness metric preferences.
Thus, our initial assumption was that ``China and Japan'' and ``France and the US'' would demonstrate similar trends.
Contrary to our assumptions, the results revealed notable differences between France and the US, while Japan and the US showed more similarities in their responses. 
This outcome suggests that the influence of regional factors on perceptions of fairness metrics is complex.
While our study provides insights into the difference in fairness metric preferences between countries, it does not sufficiently capture the differences between larger geographical areas or cultural regions.

\subsection{Comparison with other empirical studies}
Prior studies have established that participants prefer distinct fairness metrics depending on scenarios. For instance, in crime and health scenarios, participants tend to favor demographic parity~\cite{srivastava2019mathematical}. In the context of social care scenarios, the preference shifts towards equalized odds~\cite{cheng2021soliciting}. Similarly, in recruitment scenarios, equalized odds is preferable~\cite{sengewald2023}.

In our study, participants often selected equal opportunity in the US, which contradicts the findings reported in the aforementioned studies. The issue of non-experts struggling to comprehend equalized odds~\cite{saha2020measuring} may be one of the factors. Participants in our questionnaire may not adequately differentiate between equal opportunity and equalized odds.
The evaluation of quantitative parity in France has not been explored in previous studies, making it a subject for further studies.

\subsection{Difference between scenarios}
We used three distinct scenarios. We assume different preferences for fairness metrics in each scenario and correlations between them. 
Table~\ref{tab:fairmetrics} shows the number of participants who selected the same fairness metrics in different scenarios.
First, about 30--40\% people chose the same fairness metrics in scenarios.
Second, we assumed that the hiring and employee awards scenarios have large correlations because they are both related to jobs and genders, but there are no large correlations.
Therefore, since we cannot assess their correlations, we need to investigate what scenarios (e.g., sensitive attributes) have correlations.

\input{table/samemetrics}

\subsection{Limitation}

Our study has two main limitations.

\smallskip
\noindent
{\bf Demographic of participants.}
The participant demographics align with national age and gender ratios.
However, we do not control other attributes, potentially leading sampling bias.
This is due to the participants are from those already registered with the monitoring services.
For example, in China, the proportion of participants holding bachelor's degrees significantly exceeds the national statistics\footnote{\url{https://data.oecd.org/eduatt/adult-education-level.htm}}.
It is hard to collect responses from participants that exactly follow the national demographic, in particular, in online questionnaires.
Consequently, while our analysis yields valuable insights into the preferences for fairness metrics, the results should be interpreted with caution. 

\smallskip
\noindent
{\bf Other areas.}
Our questionnaires were distributed into Asia and Western hemisphere.
We do not evaluate other regions such as Africa and South America, as well as areas where Islam and Hinduism are the major religions, would provide a broader perspective.
Therefore, our survey finds that public perceptions are different across countries, yet we cannot explain similarities between countries and areas.

%% file: table/samemetrics.tex
\begin{table}[ttt]
\centering
\caption{The ratios of the same fairness metrics [\%]. ``All same'' and ``All different'' indicate a participant selects the same fairness metrics and different ones in all scenarios, respectively. }
\label{tab:fairmetrics}
\vspace{-2mm}
\centering
\begin{tabular}{lrrrr}\hline
 & \multicolumn{1}{c}{{\bf China}}& \multicolumn{1}{c}{{\bf France}}  & \multicolumn{1}{c}{{\bf Japan}}   & \multicolumn{1}{c}{{\bf US}} \\\hline
All same& 39.5 & 34.6 & 40.1 & 30.3 \\
Hiring = Art & 15.8 & 19.2 & 12.3 & 15.5 \\
Hiring = Employee &  15.6 & 13.3 & 14.5 & 14.1 \\
Art = Employee  & 14.1 & 17.5 & 18.1 & 20.8 \\
All different  & 15.0 & 15.4 & 14.9 & 19.3 \\\hline
\end{tabular}
\end{table}

%% file: 6-conclusion.tex
\section{Conclusion}
\label{sec:conc}
We conducted online questionnaires across four countries and analyzed the correlations between personal attributes and the choice of fairness metrics in decision-making scenarios.
The results of our questionnaires showed that countries have the largest impact on selected fairness metrics. These results suggest that we need to consider what fairness metrics used in our scenarios depending on which countries are targeted in the scenarios.

In the future, we extend our questionnaires to evaluate other countries and scenarios, investigate the differences between areas in the same country, and conduct different ways of surveys such as in-person workshops.

%% file: 99-appendix.tex
\section{Detailed candidate information}

\begin{figure*}[!t]
 \centering
    \begin{minipage}[t]{0.48\linewidth}
        \centering
        \includegraphics[width=1.0\linewidth]{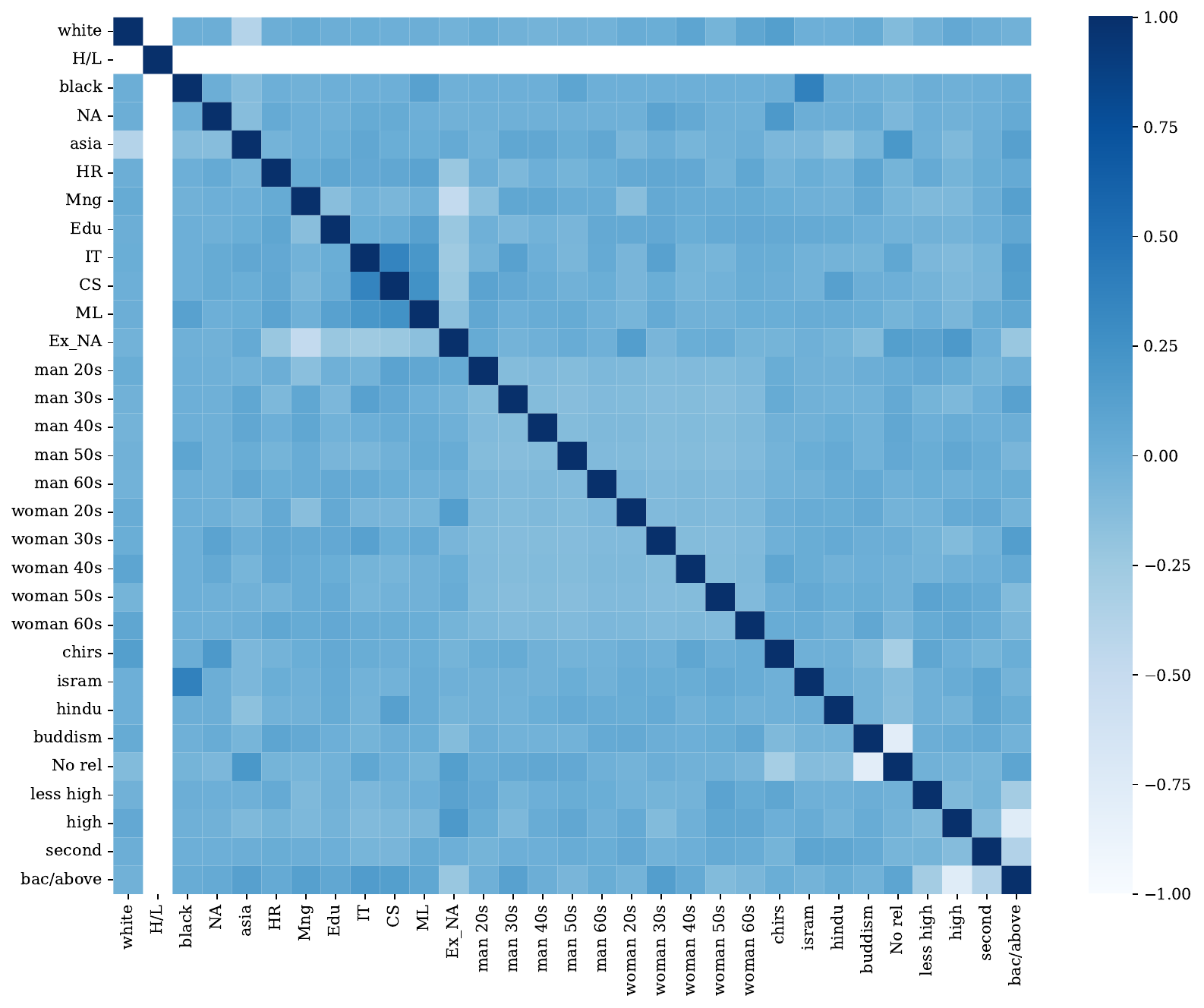}
       \subcaption{China}
       \label{fig:hiring_age}
    \end{minipage}
    \begin{minipage}[t]{0.48\linewidth}
        \centering
        \includegraphics[width=1.0\linewidth]{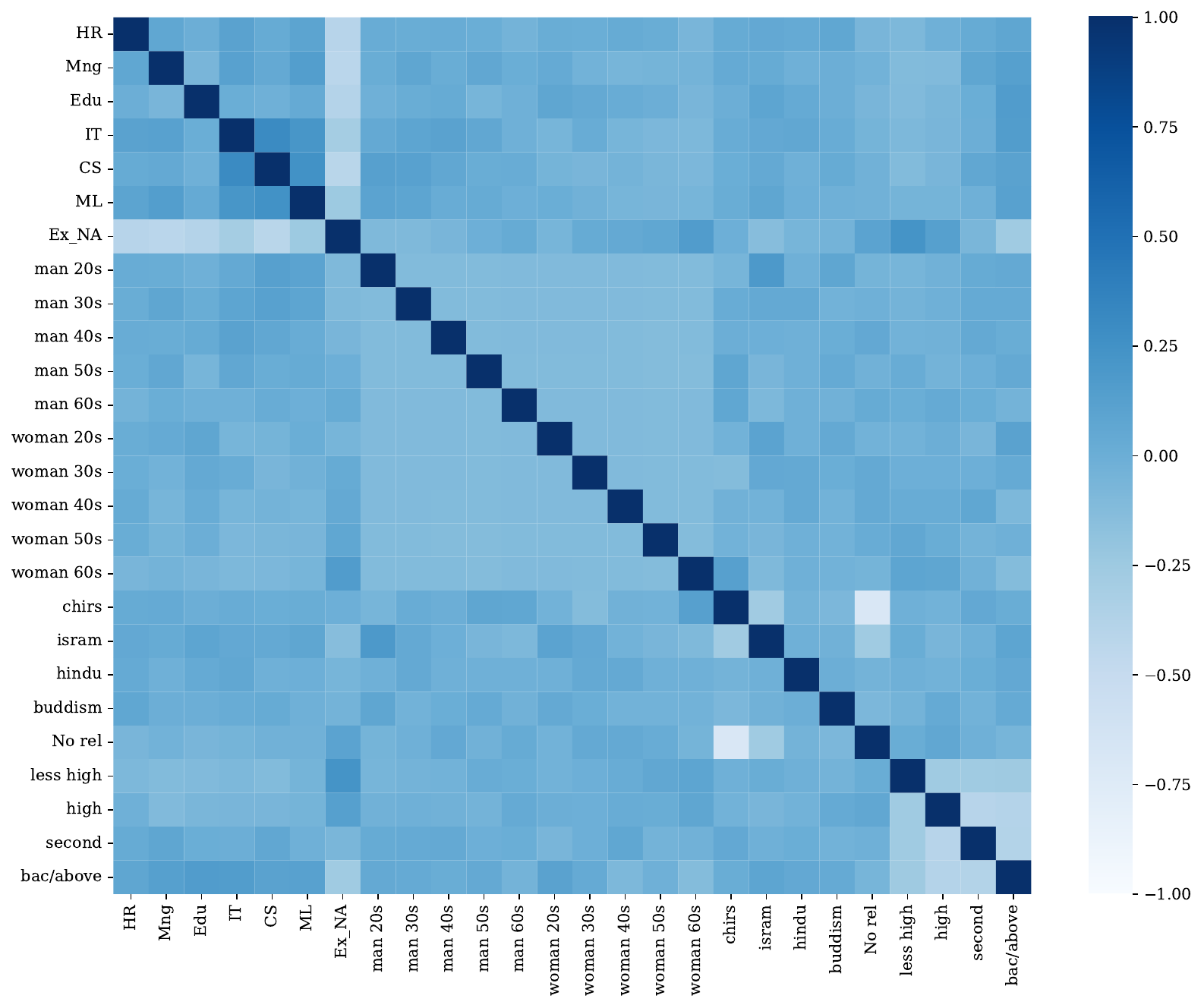}
        \subcaption{France}
        \label{fig:art_age}
    \end{minipage}
    \begin{minipage}[t]{0.48\linewidth}
        \centering
        \includegraphics[width=1.0\linewidth]{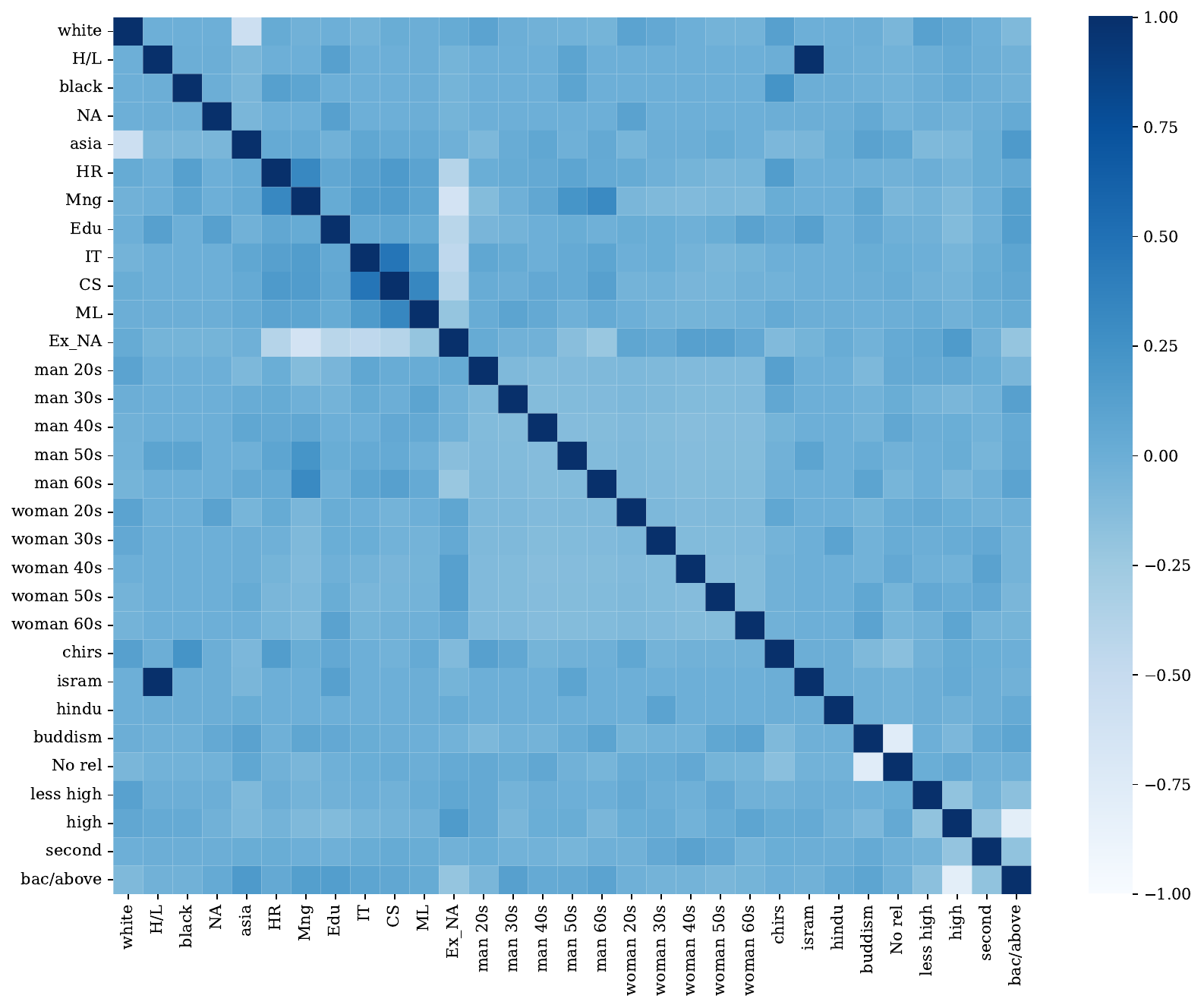}
        \subcaption{Japan}
        \label{fig:employee_age}
    \end{minipage}
        \begin{minipage}[t]{0.48\linewidth}
        \centering
        \includegraphics[width=1.0\linewidth]{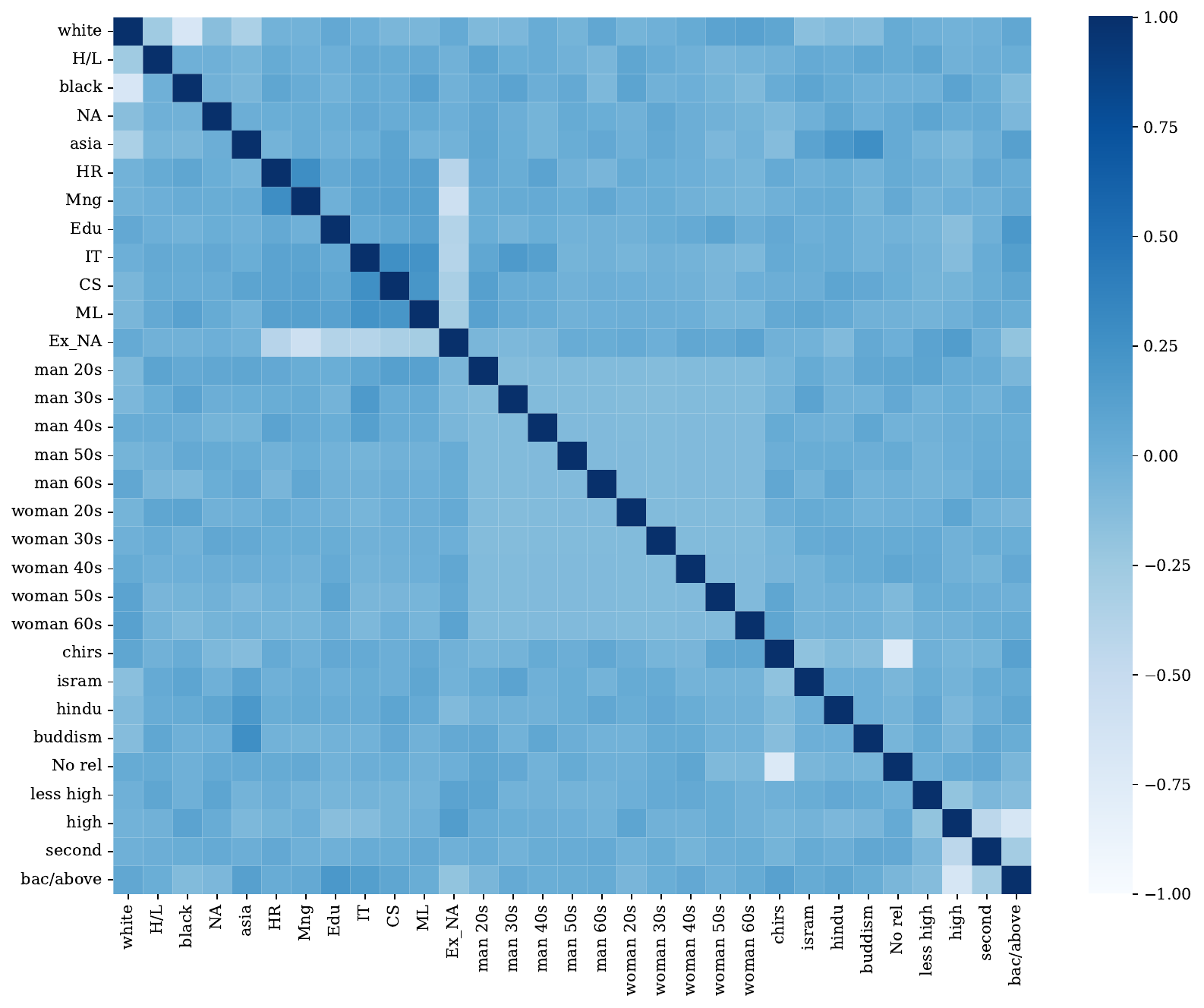}
        \subcaption{US}
        \label{fig:employee_age}
    \end{minipage}
\caption{Heatmaps for Correlation between attributes.}
\label{fig:heatmap}
\end{figure*}

Figure~\ref{fig:heatmap} shows heat maps to examine the correlations between personal attributes in each country. 
These heat maps are computed by seaborn.heatmap function in Python.
Dark and light colors indicate strong and weak correlations between personal attributes at rows and columns, respectively.
From these results, there are specifically large correlations between them except for the correlation between Hispanic/Latinx and Islam in Japan.

Figure~\ref{fig:correct_answer} shows the number of participants who answered correctly. 
China and Japan are similar and highly accurate and France and US are similar and low accurate.
We here note that if we use only participants who accurately understand the fairness metrics, the results did not significantly change.

\input{image/tex/result_correct_answer}

\section{Full results}

\subsection{Agreement levels on art project and employee award scenarios}

Figures~\ref{fig:art_boxplot} and~\ref{fig:employee_boxplot} show the distribution of agreement levels for questions assessing participants' agreement or disagreement with the reality of the hiring scenario and the applicability of various fairness metrics in art project and employee award scenarios, respectively. 
Recall that lower scores reflect higher levels of agreement.
Each scenario and fairness metric agreed with the participants.

\input{image/tex/scores_art_high}

\input{image/tex/score_employee_high}

\subsection{Full results of correlations between personal attributes and the choice of fairness metrics}

Each participant chose one of the fairness metrics that they considered the most appropriate in each scenario.
We provide the ratio of selected fairness metrics in different personal attributes; country, gender, age, ethnicity, religion, education, and experience. 

\subsubsection{Difference between genders}
Figures~\ref{fig:gender}, \ref{fig:art_gender_county}, \ref{fig:employee_gender_county} show the difference between the choices of fairness metrics in genders.
From Figure~\ref{fig:gender}, we can see that males tend to select demographic parity and equal opportunity, while female tend to select equalized odds.
Each county has slightly different trends compared with analysis of all countries (i.e., Fig.~\ref{fig:gender}). 
This indicates that we need to take a consideration the difference between countries, and it may not be the best even if we use the majority of fairness metrics that most people agree.

\subsubsection{Difference between ethnicities}
Figure~\ref{fig:ethnicity} shows the difference between the choices of fairness metrics in ethnicities.
These results indicate that the selected fairness metrics are largely different across scenarios. For example, in Asia (i.e., J\&C), White and Asian are largely different in art project, but in emplyee award, the difference is small.

\subsubsection{Difference between religions}
Figure~\ref{fig:religion} shows the difference between the choices of fairness metrics in religions.
Different from ethinicties, there are no large difference between art project and employee award scenarios.

\input{image/tex/result_gender}
\input{image/tex/result_ethnicity}
\input{image/tex/result_religions}

\subsubsection{Difference between ages}
Figures~\ref{fig:age}, \ref{fig:art_age_county}, \ref{fig:employee_age_county} show the difference between the choices of fairness metrics in ages.
From Figure~\ref{fig:age}, we can see that all ages have similar trends in all scenarios. This indicate that the fairness metrics selections are almost constant even if the scenarios are changed, for all ages.
However, if we see each result in each country, the selection change across secenarios. For example, in China, 20th often select equalized odds in art project, but they select equal opportunity in employee award.
Therefore, we need to examine the preference of fairness metrics in each country instead of global.



\subsubsection{Difference between educations}
Figures~\ref{fig:education}, \ref{fig:art_education_county}, \ref{fig:employee_education_county} show the difference between the choices of fairness metrics in educations.
From Figure~\ref{fig:education}, ``less than HS'' is prefer the quantitative parity in hiring. In art project, the participants in China are not prefer the quantitative parity (see Fig. \ref{fig:art_education_county}).


\subsubsection{Difference between experiences}
Figures~\ref{fig:experience}, \ref{fig:art_experience_county}, \ref{fig:employee_experience_county} show the difference between the choices of fairness metrics in experiences.
In experiences, there are no clear trends.
In Japan, HR often seems to select Other, but they have no clear reasons.


\input{image/tex/result_age}
\input{image/tex/result_education}
\input{image/tex/result_experience}
\input{image/tex/result_correctness}

\section{Questionnaire}

We show the actual questionnaires that we used in our study.
Note that to avoid technical terms, we used different fairness metrics name for non-experts; equal number, equal ratio, equal probability for qualified candidates, and equal probability for qualified/unqualified candidates as quantitative parity, demographic parity, equal opportunity, and equalized odds, respectively.
Although we also used four languages, Chinese, English, French, and Japanese, we here show questionnaires in English.

\subsection{Consent form}
The purpose of this research project is to investigate what kind of fairness is important to the world.
The project director is ---. 

Fairness in this study is the concept of treating certain groups without discrimination and without bias. 
Fairness is important from an ethical standpoint to prevent unfair treatment or bias based on certain attributes (e.g., race, gender, socio-economic status, etc.).
For example, in employment, the balance of how many male and female to hire is a matter of fairness.

All research participants will be asked to complete a Fairness Questionnaire consisting of the following.

\begin{enumerate}
    \item Questions regarding the demographics of the study participants
    \item Explanation and examples of fairness metrics
    \item Survey in decision-making scenarios
\end{enumerate}

Research participants will be first asked questions regarding their demographics. No personally identifiable information is collected, but an identifier for each participant is allocated. 
Next, the participants will be asked to read and understand an explanation of the fairness metrics. 
After that, the participants will be asked to read a description of the decision-making scenarios and be asked some questions regarding those scenarios.  Finally, the participants will be asked if they had any difficulties in answering the above questions.

It is extremely important that you read each question carefully and consider your answers. 
The success of our research depends on whether you are willing to take the task seriously. The survey will take approximately 30 minutes to complete.

You may find yourself getting tired completing this Questionnaire, but other than that there are minimal risks to participating.
The data collected will be stored for 10 years after the study and then deleted. 
Only the project director and co-researchers will have access to the data, which will be stored securely on a password-protected computer. The data may be provided to third parties for research use only.

If for any reason you feel that you wish to discontinue participation, you may do so at any time. Participation in this study is completely voluntary. You may choose not to participate.
If you decide to participate in this study, you may discontinue participation at any time. 
If and when you decide not to participate in this study, you will not be penalized or lose any benefits to which you are otherwise entitled.

Your identity will be protected to the greatest extent possible if we prepare a report, publication, or article about this research project. 
If you decide to discontinue participation in this study, have questions, concerns, or complaints, or need to report a research-related injury, please contact us using the Contact Us form on the survey site.

Q1. Do you agree to participate in this study?

\begin{itemize}
\item Yes
\item No
\end{itemize}

\subsection{Questions for personal attribute}

A1. Please specify the gender with which you most closely identify:

\begin{itemize}
\item Male
\item Female
\item Other
\item Prefer not to answer
\end{itemize}

A2. Please give your age:

A3. Please specify your ethnicity (you may select more than one):

\begin{itemize}
\item White
\item Hispanic or Latinx
\item Black or African American
\item Native American or Alaska Native
\item Asian, Native Hawaiian, or Pacific Islander
\item Other
\item Prefer not to answer
\end{itemize}

A4. Please specify your religion:
\begin{itemize}
\item Christianity
\item Islam
\item Hinduism
\item Buddhism
\item No Relegion
\item Other
\item Prefer not to answer
\end{itemize}

A5. Please specify the highest degree or level of school you have completed:
\begin{itemize}
\item Some high school credit, no diploma or equivalent
\item High school graduate, diploma or the equivalent (for example: GED)
\item Some college credit, no degree
\item Trade/technical/vocational training
\item Associates degree
\item Bachelors degree
\item Masters degree
\item Professional or doctoral degree (JD, MD, PhD)
\end{itemize}

A6. Please indicate which of the following areas you have worked or studied in for at least 2 years (multiple choice):
\begin{itemize}
\item Human resources (making hiring decisions)
\item Management (of employees)
\item Education (teaching)
\item IT infrastructure/systems administration
\item Computer science/programming
\item Machine learning/data science
\end{itemize}

\subsection{Explanation of fairness metrics}

It is extremely important that you read each question carefully and consider your answers. The success of our research depends on whether you are willing to take the task seriously. 

You will now consider how to select candidates from a pool of candidates divided into two attributes. Each candidate is either qualified or unqualified, and you want to select the most qualified person possible. Qualified and unqualified will not be known at the time of selection.

\begin{itemize}
\item Attributes e.g. gender, place of birth, race, etc.
\item Examples of qualified persons: those who achieve high sales after joining the company, those who have produced good works through their own efforts, etc.
\end{itemize}

In the example below, there are 5 qualified persons in Attribute 1, 3 unqualified persons in Attribute 1, 10 qualified persons in Attribute 2, and 12 unqualified persons in Attribute 2.

 \begin{figure}[!h]
 \centering
\includegraphics[width=0.6\linewidth]{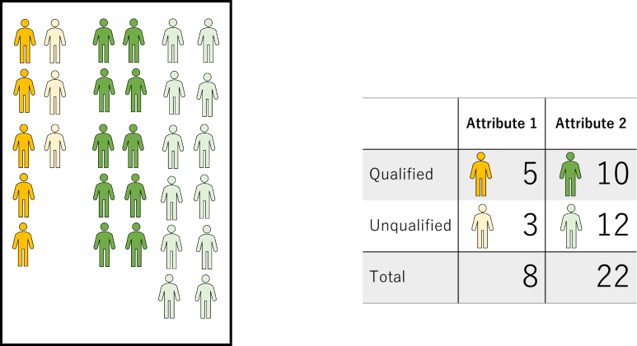}
\label{fig:quesionnares1}
\end{figure}

Fairness Metrics: Equal Number

Equal Number makes the number of selected people equal in each attribute. For example, it is fair if 100 male candidates and 100 female candidates are selected. Equal Number does not use the overall number of candidates or the number of qualified or unqualified persons in the calculation.

 \begin{figure}[!h]
 \centering
\includegraphics[width=0.9\linewidth]{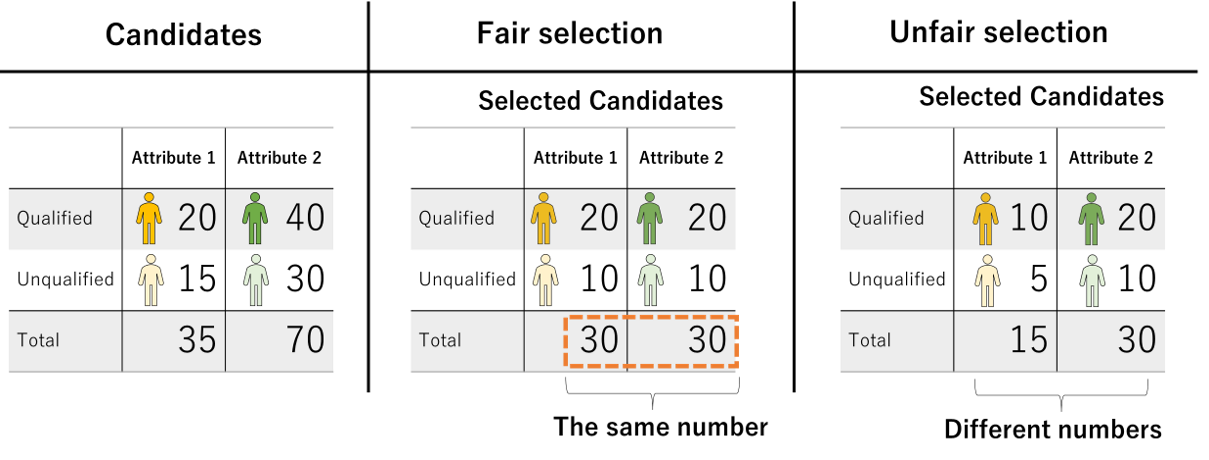}
\label{fig:quesionnares1}
\end{figure}

In the above examples, an example of selecting an equal number of 30 people each from Attributes 1 and 2 is fair.

Comprehension Checks

B1. Which of the following choices is fair?

\begin{enumerate}
    \item Select 20 persons for Attribute 1 and 20 persons for Attribute 2
    \item Select 50 persons for Attribute 1 and 20 persons for Attribute 2
    \item Select 20 persons for Attribute 1 and 10 persons for Attribute 2
    \item Select 10 persons for Attribute 1 and 20 persons for Attribute 2
\end{enumerate}

B2. There are 100 candidates for Attribute 1 and 10 of them have been selected. Which of the following choices is fair?

\begin{enumerate}
    \item There are 10 candidates for Attribute 2, and 10 of them are selected
    \item There are 100 candidates for Attribute 2, and 20 of them are selected
    \item There are 200 candidates for Attribute 2, and 20 of them are selected
    \item There are 200 candidates for Attribute 2, and 40 of them are selected
\end{enumerate}

Fairness Metrics: Equal Ratio

Equal Ratio makes the number of selected people equal in each attribute. For example, it is fair if 100 male candidates and 100 female candidates are selected. 
Equal Ratio does not use the overall number of candidates or the number of qualified or unqualified persons in the calculation.

 \begin{figure}[!h]
 \centering
\includegraphics[width=0.9\linewidth]{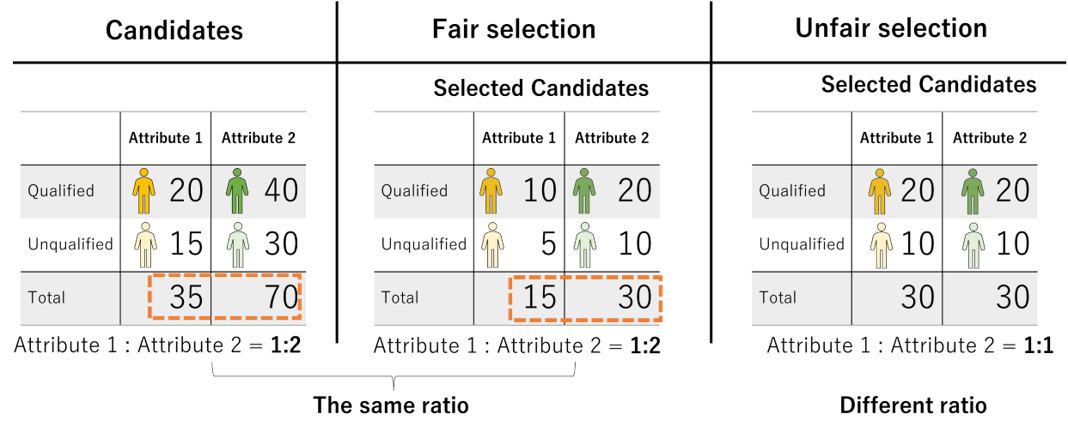}
\label{fig:quesionnares1}
\end{figure}

In the above example, the left example is fair, selecting 15 candidates from 35 candidates with Attribute 1 and 30 candidates from 70 candidates with Attribute 2 in the same proportion.

Comprehension Checks

B3. If there are 100 candidates for Attribute 1 and 50 candidates for Attribute 2, which of the following choices is fair?

\begin{enumerate}
    \item Select 20 persons for Attribute 1 and 20 persons for Attribute 2
    \item Select 50 persons for Attribute 1 and 20 persons for Attribute 2
    \item Select 20 persons for Attribute 1 and 10 persons for Attribute 2
    \item Select 10 persons for Attribute 1 and 20 persons for Attribute 2
\end{enumerate}

B4. There are 100 candidates for Attribute 1 and 10 of them have been selected. Which of the following selections is fair?

\begin{enumerate}
    \item There are 10 candidates for Attribute 2, and 10 of them are selected
    \item There are 100 candidates for Attribute 2, and 20 of them are selected
    \item There are 200 candidates for Attribute 2, and 20 of them are selected
    \item There are 200 candidates for Attribute 2, and 40 of them are selected
\end{enumerate}

Fairness Metric: Equal Probability for Qualified Candidates

Equal Probability for Qualified Candidates is a metric that selects the same proportion of qualified persons in each attribute. For example, it is fair if 5 persons are selected from a pool of 10 qualified male persons and 10 persons are selected from a pool of 20 qualified female persons.
This means that the number of qualified persons has an equal probability of being selected. Equal Probability for Qualified/Unqualified Candidates is calculated from qualified persons only, without regard to unqualified persons.

 \begin{figure}[!h]
 \centering
\includegraphics[width=0.9\linewidth]{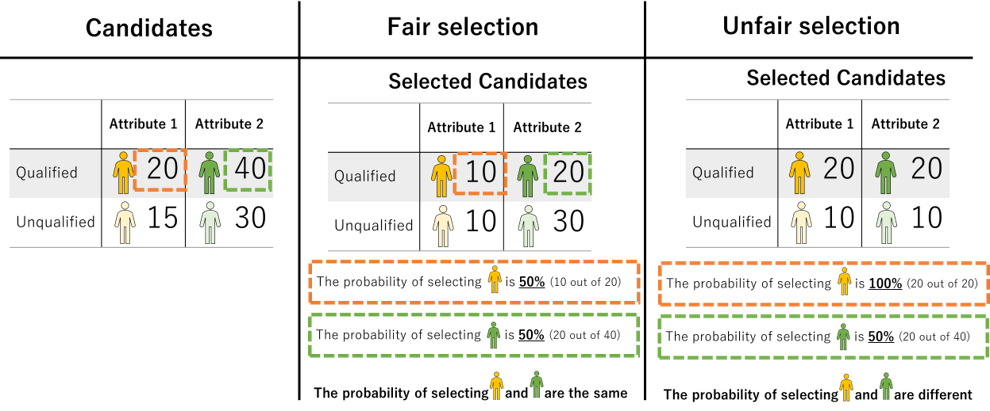}
\label{fig:quesionnares1}
\end{figure}

In the above example, the left example is fair, with 10 out of 20 qualified persons in Attribute 1 and 20 out of 40 qualified persons in Attribute 2, both selecting qualified persons at the same 50\% rate.

Comprehension Checks

B5. If 10 persons are selected from 40 qualified persons in Attribute 1, which of the following choices is fair?

\begin{enumerate}
    \item Select 10 out of 10 qualified persons with attribute 2.
    \item Select 5 out of 20 qualified persons with attribute 2.
    \item Select 20 out of 20 qualified persons with attribute 2.
    \item Select 10 out of 20 qualified persons with attribute 2.
\end{enumerate}

B6. 10 qualified persons were selected from Attribute 1 and 20 qualified persons were selected from Attribute 2. Which of the following situations is fair?

\begin{enumerate}
    \item There are 20 qualified persons in Attribute 1 and 40 qualified persons in Attribute 2 within the candidate
    \item There are 10 qualified persons in Attribute 1 and 40 qualified persons in Attribute 2 within the candidate
    \item There are 20 qualified persons in Attribute 1 and 20 qualified persons in Attribute 2 within the candidate
    \item There are 20 qualified persons in Attribute 1 and 30 qualified persons in Attribute 2 within the candidate
\end{enumerate}

Fairness Metric: Equal Probability for Qualified/Unqualified Candidates

Equal Probability for Qualified/Unqualified Candidates is a metric that selects the same percentage of qualified persons and the same percentage of unqualified persons in each attribute.
For example, if there are 10 qualified and 20 unqualified male candidates and 20 unqualified and 40 unqualified female candidates, it is fair to select 5 qualified and 10 unqualified persons from males and 10 qualified and 20 unqualified persons from females.

Compared to Equal Probability for Qualified Candidates, Equal Probability for Qualified/Unqualified Candidates is a more difficult metric to achieve because it also considers unqualified persons. If Equal Probability for Qualified/Unqualified Candidates is fair, then Equal Probability for Qualified Candidates is also fair.

 \begin{figure}[!h]
 \centering
\includegraphics[width=0.9\linewidth]{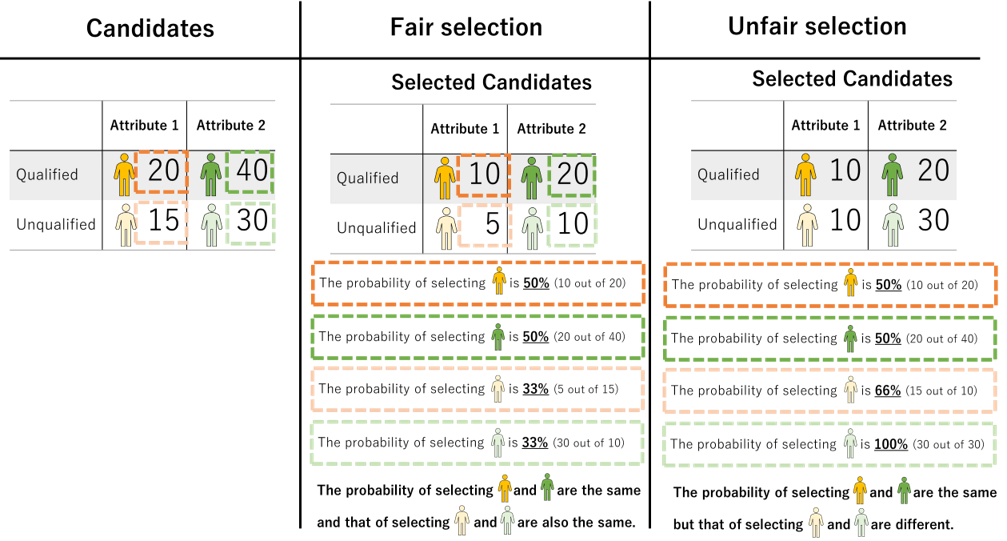}
\label{fig:quesionnares1}
\end{figure}

In the above example, the left example is fair, selecting 10 qualified persons from 20 qualified persons in Attribute 1, the same 50\% of the time as 20 qualified persons from 40 qualified persons in Attribute 2, and selecting 5 unqualified persons from 15 unqualified persons in Attribute 1, the same 33\% of the time as 10 unqualified persons from 30 qualified persons in Attribute 2.

Comprehension Checks

B7. Attribute 1 has 40 qualified persons and 60 unqualified persons. Attribute 2 has 80 qualified persons and 120 unqualified persons. If 10 qualified persons and 10 unqualified persons are selected from Attribute 1, which of the following choices is fair?

\begin{enumerate}
    \item Select 10 qualified persons and 10 unqualified persons from Attribute 2.
    \item Select 20 qualified persons and 20 unqualified persons from Attribute 2.
    \item Select 30 qualified persons and 30 unqualified persons from Attribute 2.
    \item Select 40 qualified persons and 40 unqualified persons from Attribute 2.
\end{enumerate}

B8. 10 unqualified persons were selected from 20 qualified persons in Attribute 1 and 5 unqualified persons were selected from 15 qualified persons. Which of the following cases is fair?

\begin{enumerate}
    \item Selected 20 out of 50 qualified persons and 10 out of 40 unqualified persons for Attribute 2.
    \item Selected 20 out of 40 qualified persons and 10 out of 20 unqualified persons for Attribute 2.
    \item Selected 20 out of 40 qualified persons and 10 out of 30 unqualified persons for Attribute 2.
    \item Selected 10 out of 30 qualified persons and 5 out of 15 unqualified persons for Attribute 2.
\end{enumerate}

\subsection{Scenario}

It is extremely important that you read each question carefully and consider your answers. The success of our research depends on whether you are willing to take the task seriously. 

In the following, please read the three scenarios and answer the questions about fairness metrics.

Scenario: Hiring	
	
A recruiter for a new sales company is reviewing job applications. Each applicant submits a resume and is interviewed. The hiring manager's goal is to make an offer to the applicant with the highest net sales after one year of employment.	
	
The recruiter uses the following scores to determine which applicants to make job offers to	
\begin{itemize}
\item Interviews	
\item Recommendation letters	
\item Years of experience in this field	
\end{itemize}
 
In the past, after one year of employment, male applicants have on average achieved higher net sales than female applicants. The number of male applicants is extremely high compared to female applicants. We hire according to the quality of the applicant (measured by interview scores, recommendation letters, and years of experience in the field). 	
There is a correlation between the quality of the applicant and sales after one year of employment, but on the other hand, high quality does not necessarily lead to high sales after one year of employment, and low quality may also lead to high sales.	
	
The recruiter wants to make sure that this hiring process is fair regardless of the gender of the applicant.	

\begin{itemize}  
\item Attribute 1: Male Applicants	
\item Attribute 2: Female applicants	
\item Qualified persons: Applicants with high net sales after one year of employment	
\item Unqualified persons: Applicants with low net sales after one year of employment	
\end{itemize}
	
For example, the following cases may be considered:	

\begin{itemize}  
\item Equal Number: Select an equal number of male and female applicants	
\item Equal Ratio : Select male and female applicants in equal proportions according to the proportion of male and female applicants	
\item Equal Probability for Qualified Candidates: Select male and female applicants in equal proportions according to the proportion of male and female applicants who are expected to have high net sales after one year of employment.	
\item Equal Probability for Qualified/Unqualified Candidates: Select male and female applicants in equal proportions according to the proportion of male and female applicants who are expected to have high net sales after one year of employment, while selecting male and female applicants in equal proportions who may not have high net sales after one year of employment.	
\end{itemize}

C1. Please answer the following

C1-1. Could the above scenario happen in reality?

C1-2. Do you agree with using Equal Number as a criterion of fairness?

C1-3. Do you agree with using Equal Ratio as a criterion of fairness?

C1-4. Do you agree with using Equal Probability for Qualified Candidates as a criterion of fairness?

C1-5. Do you agree with using Equal Probability for Qualified/Unqualified Candidates as a criterion for fairness?

\begin{enumerate}
    \item Very strongly agree
    \item Strongly agree
    \item Somewhat agree
    \item Neither agree nor disagree
    \item Not so much
    \item Disagree
    \item Not at all
\end{enumerate}

C1-6. If you have to choose one, which rule would you adopt?
\begin{enumerate}
    \item Equal Number 
    \item Equal Ratio 
    \item Equal Probability for Qualified Candidates
    \item Equal Probability for Qualified/Unqualified Candidates
    \item Adopt a different rule
\end{enumerate}

C1-7. Please explain the reasons for the choice you made in C1-6 in terms of fairness (if you chose "adopt a different rule," please also indicate that rule).

Scenario: Art Project

A teacher is judging a student's art project and would like to award a student who has put a lot of effort into the project. The teacher knows that some students have parents who are artists and might be helping their own children with their art projects.
The teacher's goal is to award students who have created outstanding artwork solely through the student's own efforts.

Students whose parents are not artists can be legitimately evaluated on the basis of their projects alone. Students whose parents are artists may have their work valued more highly due to parental assistance, but this is difficult to detect in advance.

In the past, students with parents who were artists tended to put more effort into their projects and thus received higher scores regardless of parental assistance. A teacher wants to make sure that students are awarded in an equitable way, regardless of whether their parents are artists or not.

\begin{itemize} 
\item Attribute 1: Students whose parents are not artists
\item Attribute 2: Students whose parents are artists
\item Qualified person: A student who has created an excellent work of art solely through his/her own efforts.
\item Unqualified person: A student who created a less than excellent work of art, or a student whose parent helped in any way.
\end{itemize}

For example, the following cases may be considered:

\begin{itemize}
\item Equal Number: Select an equal number of students according to the ratio of students whose parents are artists to those whose parents are not.
\item Equal Ratio: Select students whose parents are artists and students whose parents are not artists in equal proportions according to the proportion of students whose parents are artists and the proportion of students whose parents are not artists.
\item Equal Probability for Qualified Candidates: Select students whose parents are artists and students whose parents are not artists according to the proportion of students creating award-worthy work whose parents are artists and of those whose parents are not artists.
\item Equal Probability for Qualified/Unqualified Candidates: Select students who create award-worthy work from the group that has parents as artists and parents that aren’t artists. Then also select students who did not create award-worthy work from the group that has parents as artists and parents that are not artists.
\end{itemize}

C2. Please answer the following

C2-1. Could the above scenario happen in reality?

C2-2. Do you agree with using Equal Number as a criterion of fairness?

C2-3. Do you agree with using Equal Ratio as a criterion of fairness?

C2-4. Do you agree with using Equal Probability for Qualified Candidates as a criterion of fairness?

C2-5. Do you agree with using Equal Probability for Qualified/Unqualified Candidates as a criterion for fairness?

\begin{enumerate}
    \item Very strongly agree
    \item Strongly agree
    \item Somewhat agree
    \item Neither agree nor disagree
    \item Not so much
    \item Disagree
    \item Not at all
\end{enumerate}

C2-6. If you have to choose one, which rule would you adopt?

\begin{enumerate}
    \item Equal Number 
    \item Equal Ratio 
    \item Equal Probability for Qualified Candidates
    \item Equal Probability for Qualified/Unqualified Candidates
    \item Adopt a different rule
\end{enumerate}

C2-7. Please explain the reasons for the choice you made in C2-6 in terms of fairness (if you chose "adopt a different rule," please also indicate that rule).

Scenario: Employee Awards

A manager of a sales company has been asked to determine mid-year award winners from among his employees. The manager's goal is to award the employees with the high sales at the end of the fiscal year; however, the number of employees with high sales at the end of the fiscal year is likely to be less than the number of mid-year award recipients.
The manager will determine award winners based on mid-year sales.

In the past, women achieved higher year-end sales than men. The number of employees is higher for men. There is a strong correlation between mid-year sales and year-end sales.
The manager wants to ensure that this award process is fair to employees regardless of gender.

\begin{itemize}
\item Attribute 1: Male employees
\item Attribute 2: Female employees
\item Qualified person: An employees with high sales at the end of the fiscal year
\item Unqualified person: An employee who does not have high sales at the end of the fiscal year
\end{itemize}

For example, the following cases may be considered:

\begin{itemize}
\item Equal Number: Select an equal number of male and female employees.
\item Equal Ratio: Select an equal ratio of male and female employees according to the proportion of male and female employees.
\item Equal Probability for Qualified Candidates: Select an equal percentage of male and female employees based on the proportion of male and female employees who are expected to have higher sales at the end of the fiscal year.
\item Equal Probability for Qualified/Unqualified Candidates: Select an equal proportion of male and female employees who are expected to have high sales at the end of the fiscal year and an equal proportion of male and female employees who are likely to have low sales at the end of the fiscal year
\end{itemize}

C3. Please answer the following

C3-1. Could the above scenario happen in reality?

C3-2.Do you agree with using Equal Number as a criterion of fairness?

C3-3.Do you agree with using Equal Ratio as a criterion of fairness?

C3-4.Do you agree with using Equal Probability for Qualified Candidates as a criterion of fairness?

C3-5.Do you agree with using Equal Probability for Qualified/Unqualified Candidates as a criterion for fairness?

\begin{enumerate}
    \item Very strongly agree
    \item Strongly agree
    \item Somewhat agree
    \item Neither agree nor disagree
    \item Not so much
    \item Disagree
    \item Not at all
\end{enumerate}

C3-6. If you have to choose one, which rule would you adopt?

\begin{enumerate}
    \item Equal Number 
    \item Equal Ratio 
    \item Equal Probability for Qualified Candidates
    \item Equal Probability for Qualified/Unqualified Candidates
    \item Adopt a different rule
\end{enumerate}

C3-7. Please explain the reasons for the choice you made in C3-6 in terms of fairness (if you chose "adopt a different rule," please also indicate that rule).

\subsection{Overall Question}

What have you had difficulty in understanding or answering about the overall question so far?

%% file: image/tex/result_correct_answer.tex
 \begin{figure*}[!t]
 \centering
    \begin{minipage}[t]{0.24\linewidth}
        \centering
        \includegraphics[width=1.0\linewidth]{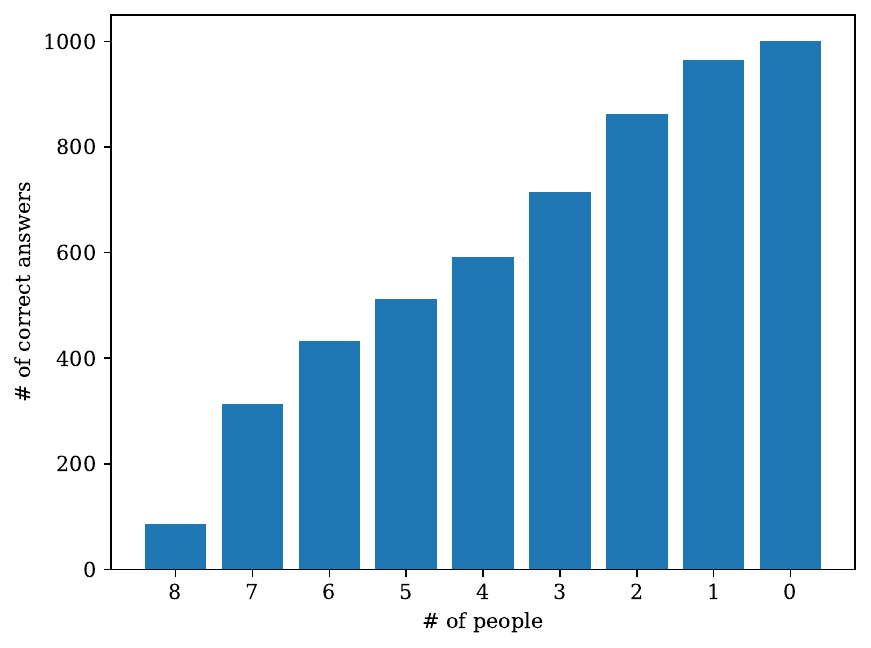}
       \subcaption{China}
       \label{fig:art_boxplot_china}
    \end{minipage}
    \begin{minipage}[t]{0.24\linewidth}
        \centering
        \includegraphics[width=1.0\linewidth]{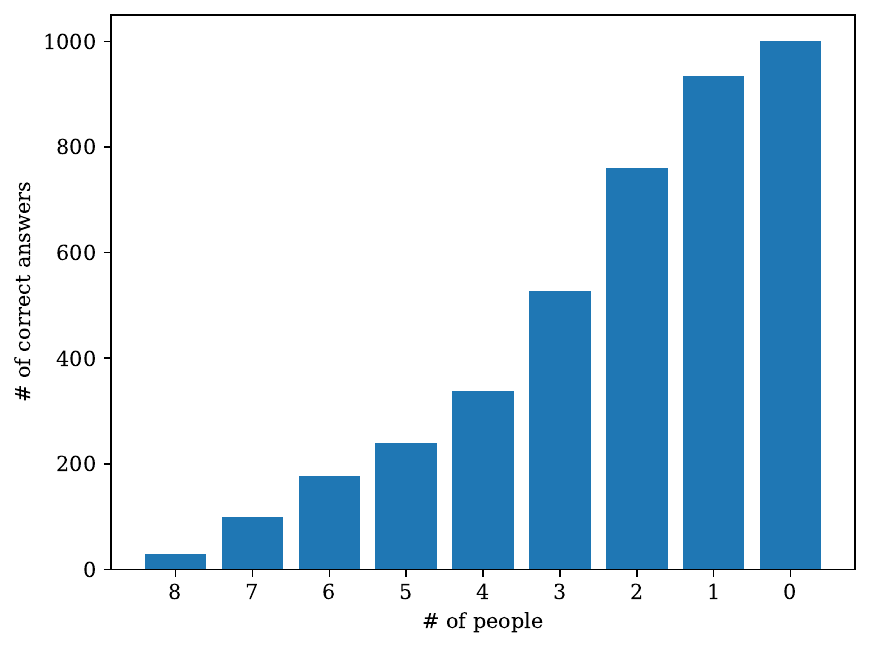}
        \subcaption{France}
        \label{fig:art_boxplot_japan}
    \end{minipage}
    \begin{minipage}[t]{0.24\linewidth}
        \centering
        \includegraphics[width=1.0\linewidth]{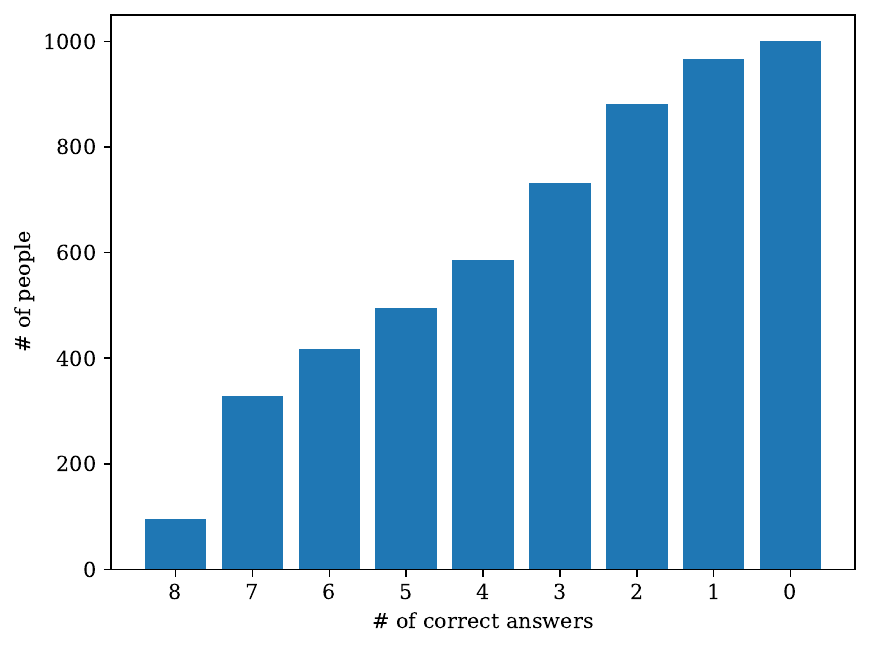}
        \subcaption{Japan}
        \label{fig:art_boxplot_france}
    \end{minipage}
    \begin{minipage}[t]{0.24\linewidth}
    \centering
    \includegraphics[width=1.0\linewidth]{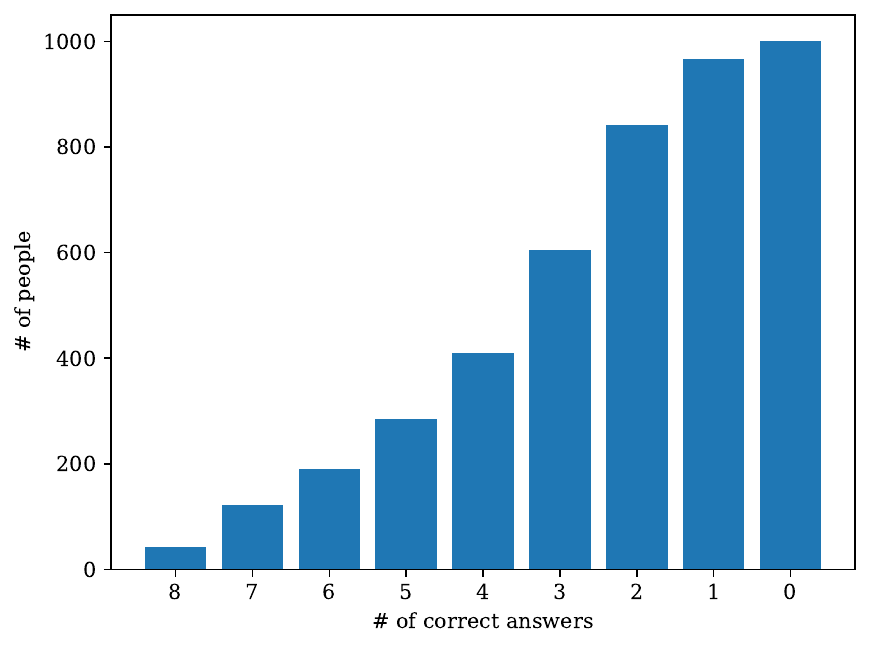}
    \subcaption{US}
    \label{fig:art_boxplot_us}
    \end{minipage}
\caption{The number of participants who correctly answer fairness metrics questions in each country}
\label{fig:correct_answer}
\end{figure*}

%% file: image/tex/scores_art_high.tex
 \begin{figure*}[!t]
 \centering
    \begin{minipage}[t]{0.24\linewidth}
        \centering
        \includegraphics[width=1.0\linewidth]{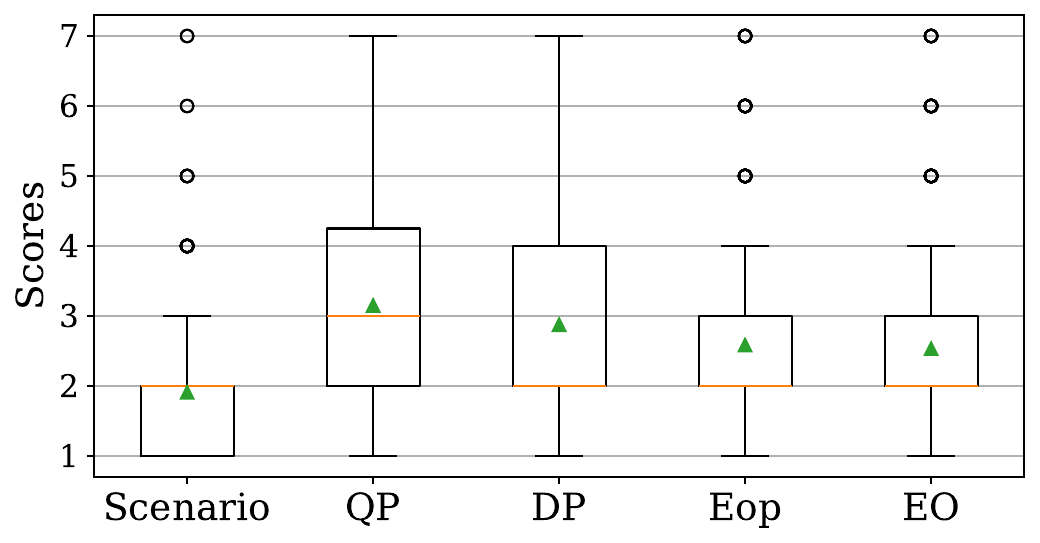}
       \subcaption{China}
       \label{fig:art_boxplot_china}
    \end{minipage}
    \begin{minipage}[t]{0.24\linewidth}
        \centering
        \includegraphics[width=1.0\linewidth]{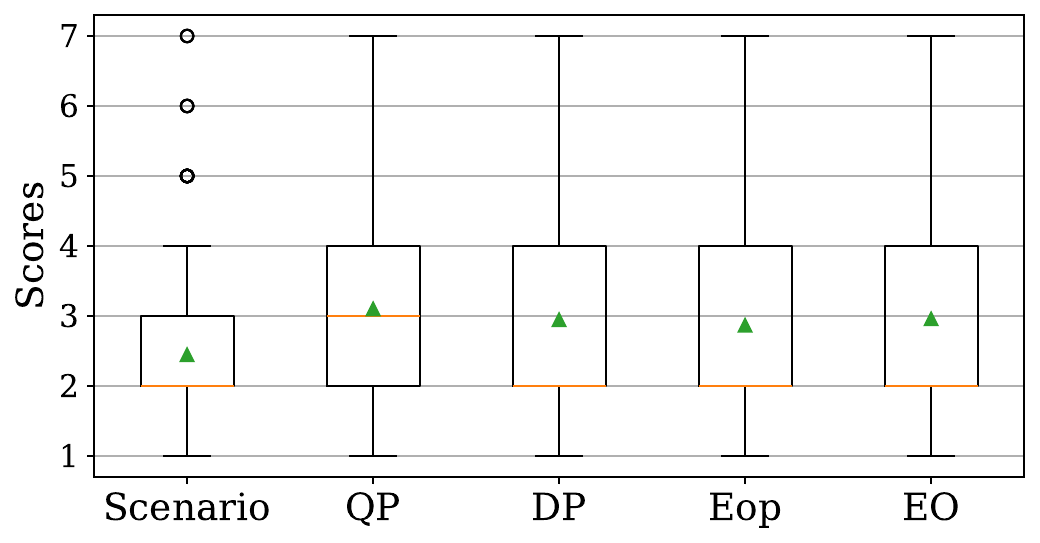}
        \subcaption{France}
        \label{fig:art_boxplot_japan}
    \end{minipage}
    \begin{minipage}[t]{0.24\linewidth}
        \centering
        \includegraphics[width=1.0\linewidth]{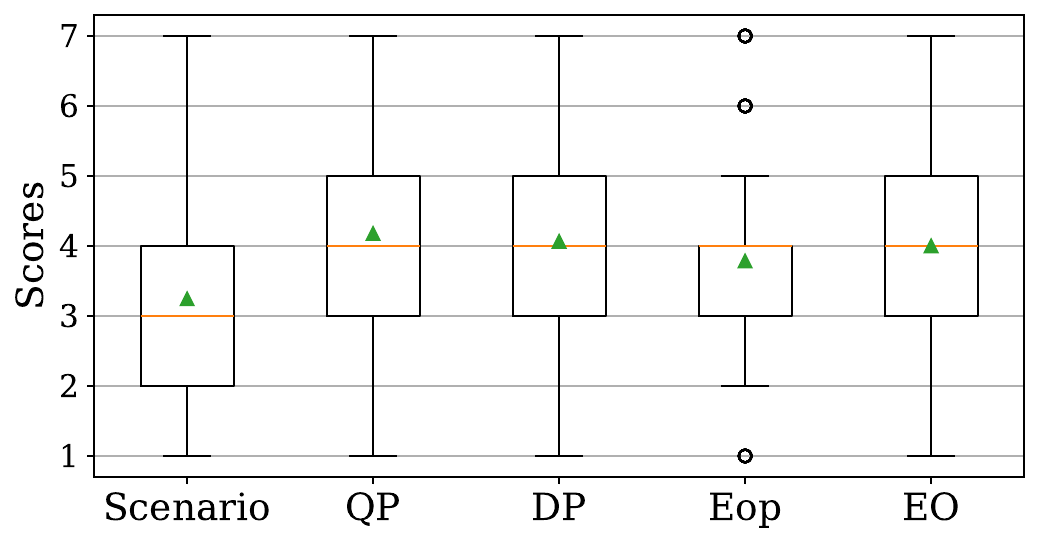}
        \subcaption{Japan}
        \label{fig:art_boxplot_france}
    \end{minipage}
    \begin{minipage}[t]{0.24\linewidth}
    \centering
    \includegraphics[width=1.0\linewidth]{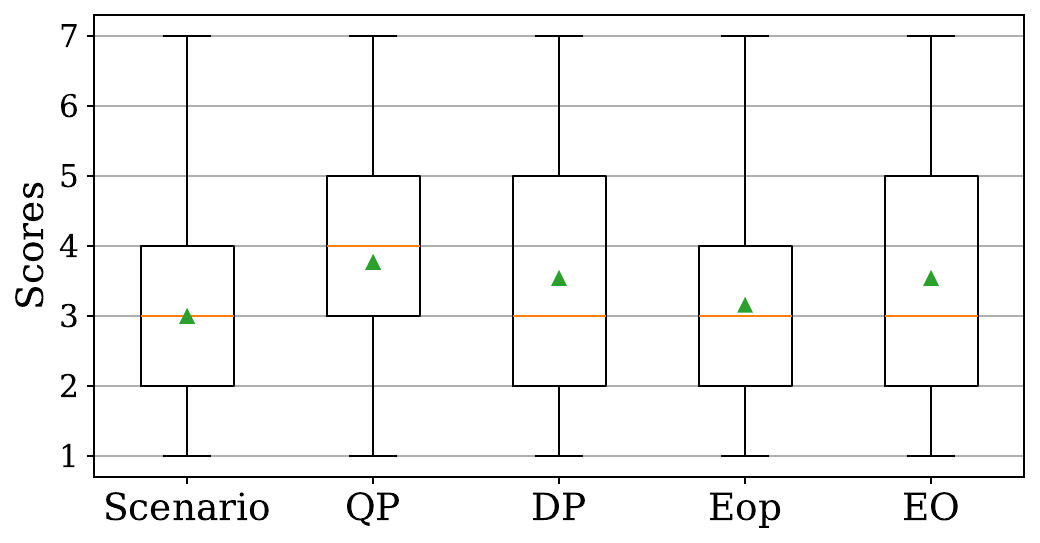}
    \subcaption{US}
    \label{fig:art_boxplot_us}
    \end{minipage}
\caption{Scores of answers for art scenario}
\label{fig:art_boxplot}
\end{figure*}

%% file: image/tex/score_employee_high.tex
 \begin{figure*}[!t]
 \centering
    \begin{minipage}[t]{0.24\linewidth}
        \centering
        \includegraphics[width=1.0\linewidth]{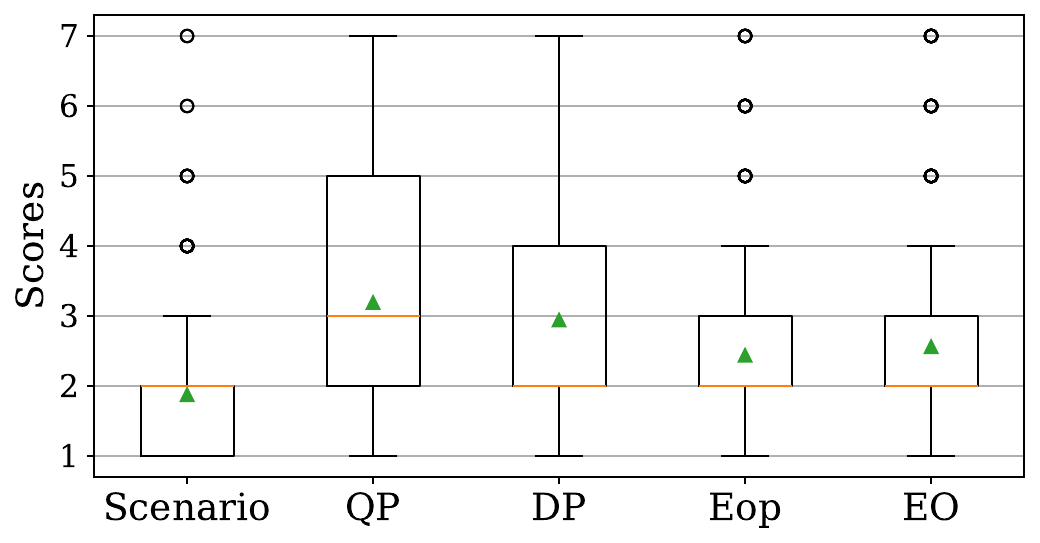}
       \subcaption{China}
       \label{fig:employee_boxplot_china}
    \end{minipage}
    \begin{minipage}[t]{0.24\linewidth}
        \centering
        \includegraphics[width=1.0\linewidth]{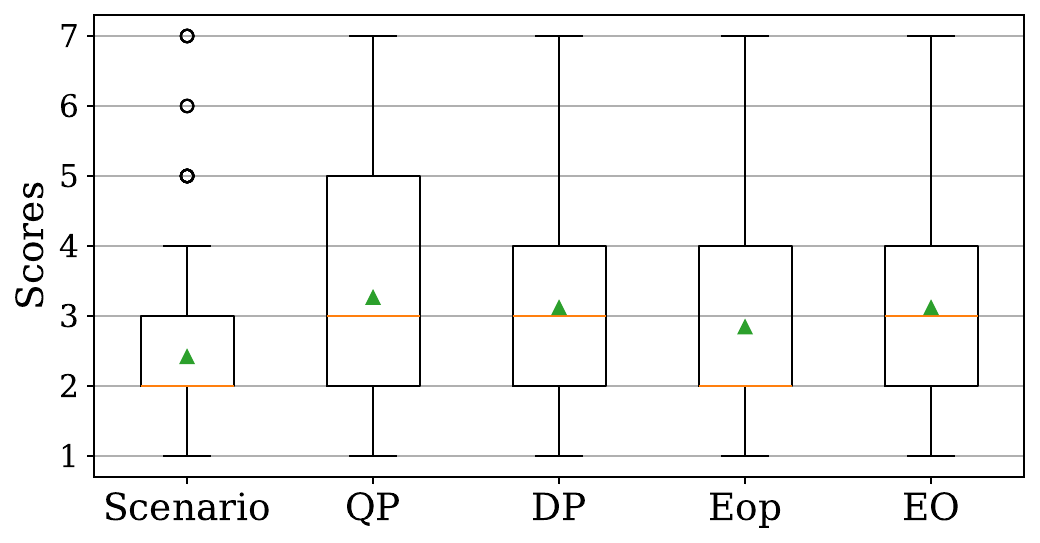}
        \subcaption{France}
        \label{fig:employee_boxplot_japan}
    \end{minipage}
    \begin{minipage}[t]{0.24\linewidth}
        \centering
        \includegraphics[width=1.0\linewidth]{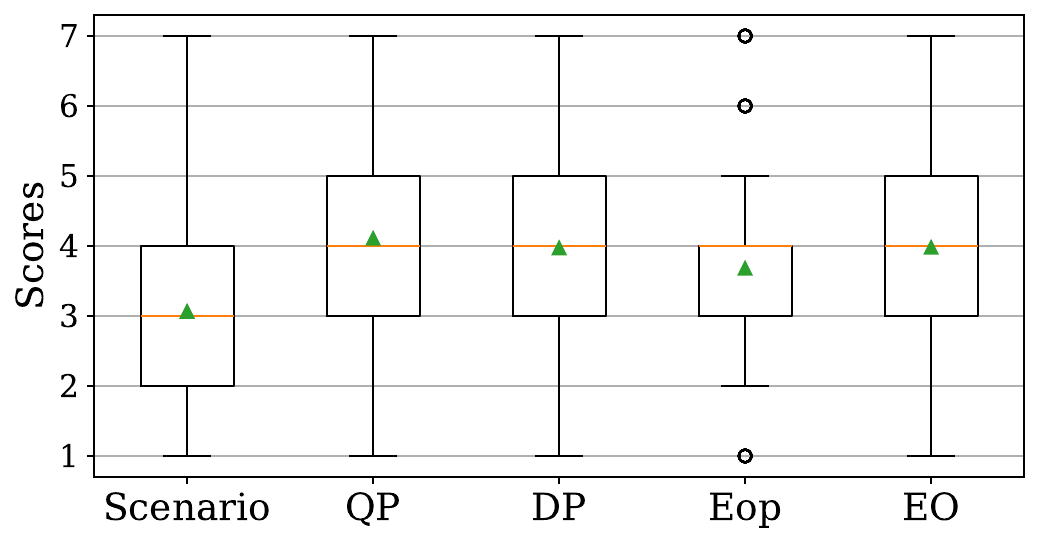}
        \subcaption{Japan}
        \label{fig:employee_boxplot_france}
    \end{minipage}
    \begin{minipage}[t]{0.24\linewidth}
    \centering
    \includegraphics[width=1.0\linewidth]{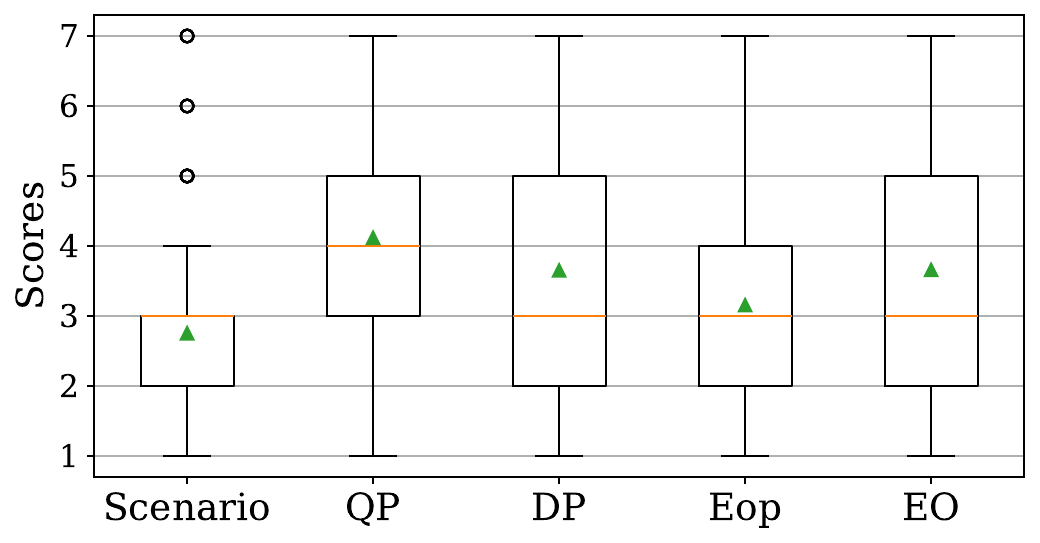}
    \subcaption{US}
    \label{fig:employee_boxplot_us}
    \end{minipage}
\caption{Scores of answers for employee scenario}
\label{fig:employee_boxplot}
\end{figure*}

%% file: image/tex/result_gender.tex
\begin{figure*}[!t]
 \centering
    \begin{minipage}[t]{0.32\linewidth}
        \centering
        \includegraphics[width=1.0\linewidth]{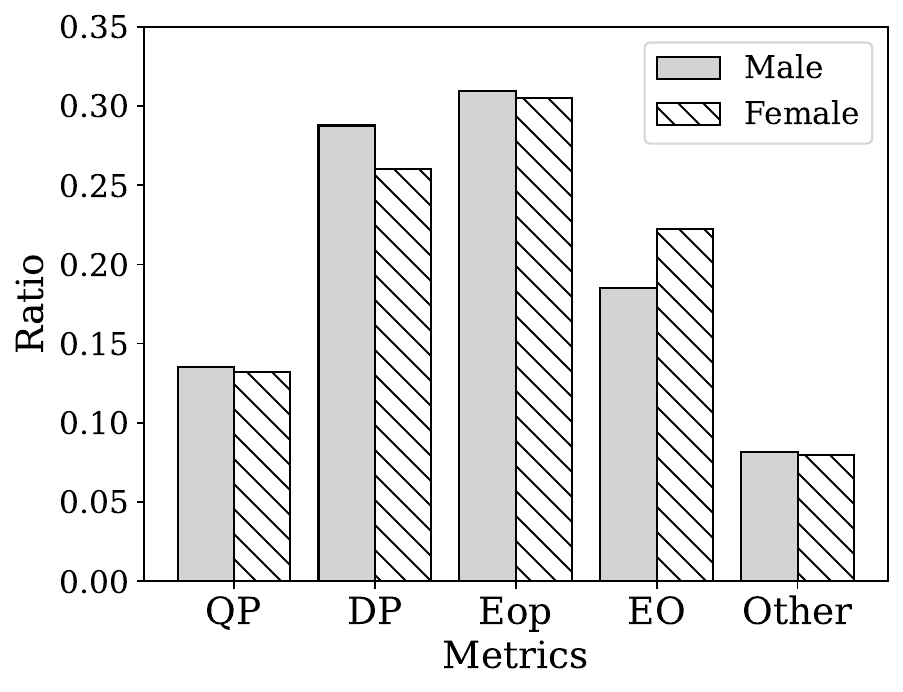}
       \subcaption{Hiring*}
       \label{fig:hiring_gender}
    \end{minipage}
    \begin{minipage}[t]{0.32\linewidth}
        \centering
        \includegraphics[width=1.0\linewidth]{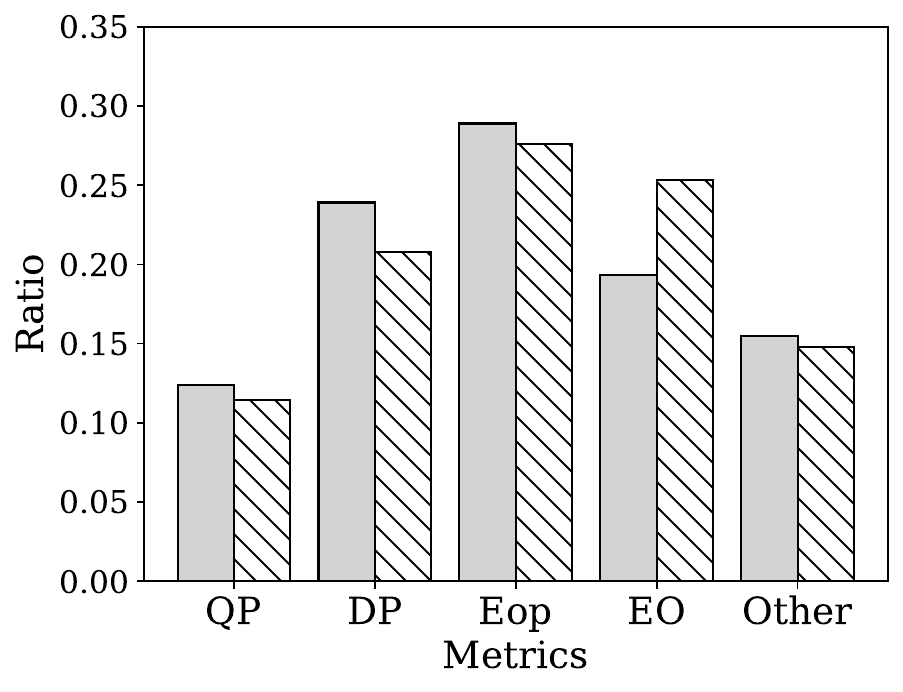}
        \subcaption{Art project**}
        \label{fig:art_gender}
    \end{minipage}
    \begin{minipage}[t]{0.32\linewidth}
        \centering
        \includegraphics[width=1.0\linewidth]{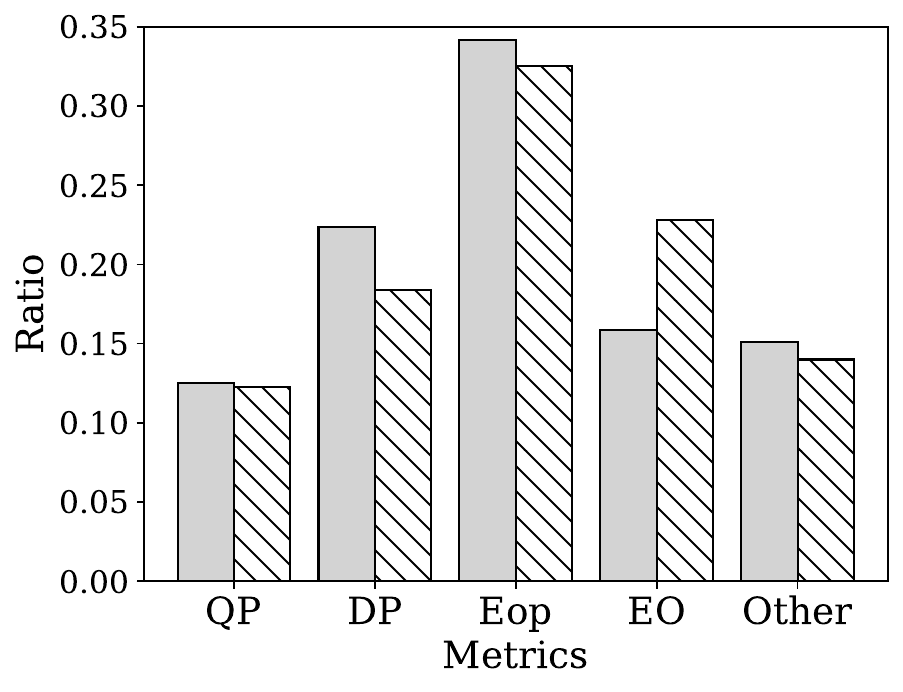}
        \subcaption{Employee award**}
        \label{fig:employee_gender}
    \end{minipage}
\caption{Difference between gender.}
\label{fig:gender}
\end{figure*}

 \begin{figure*}[!t]
 \centering
    \begin{minipage}[t]{0.24\linewidth}
        \centering
        \includegraphics[width=1.0\linewidth]{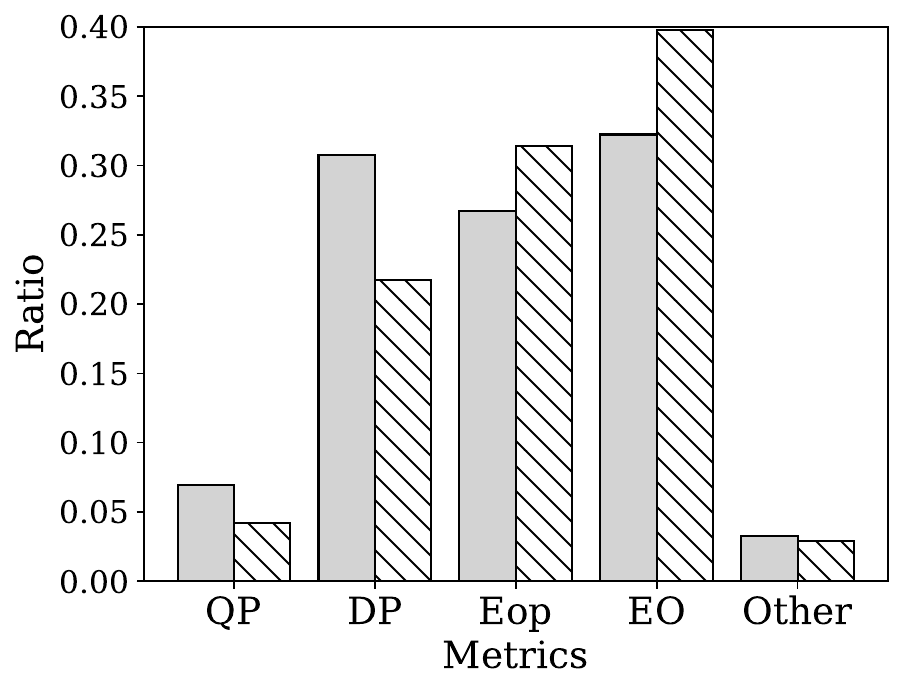}
       \subcaption{China***}
       \label{fig:hiring_boxplot_china}
    \end{minipage}
    \begin{minipage}[t]{0.24\linewidth}
        \centering
        \includegraphics[width=1.0\linewidth]{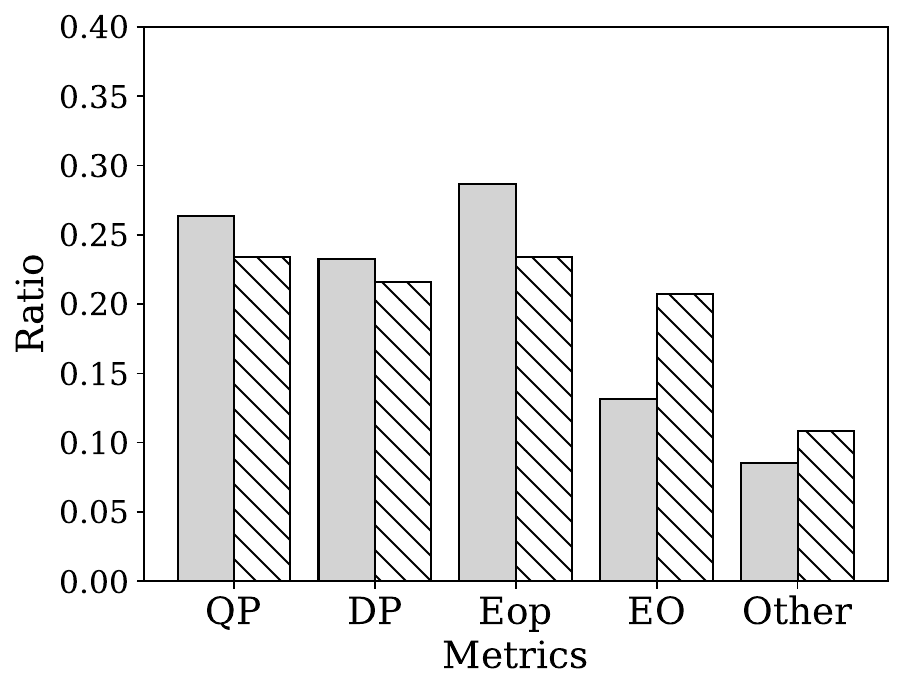}
        \subcaption{France}
        \label{fig:hiring_boxplot_japan}
    \end{minipage}
    \begin{minipage}[t]{0.24\linewidth}
        \centering
        \includegraphics[width=1.0\linewidth]{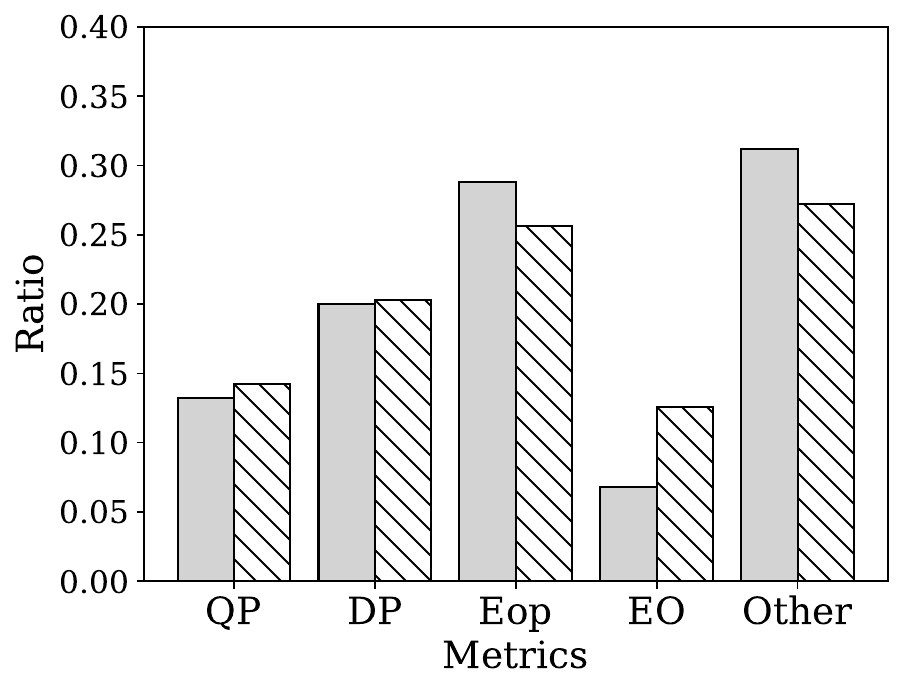}
        \subcaption{Japan*}
        \label{fig:hiring_boxplot_france}
    \end{minipage}
    \begin{minipage}[t]{0.24\linewidth}
    \centering
    \includegraphics[width=1.0\linewidth]{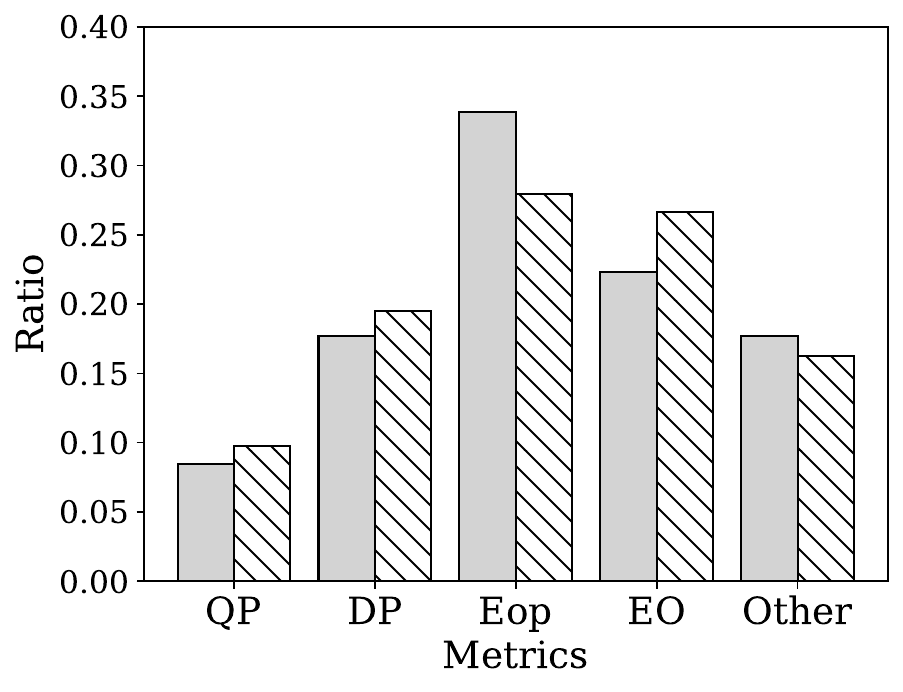}
    \subcaption{US}
    \label{fig:hiring_boxplot_us}
    \end{minipage}
\caption{Difference between genders in countries in art project scenario}
\label{fig:art_gender_county}
\end{figure*}

 \begin{figure*}[!t]
 \centering
    \begin{minipage}[t]{0.24\linewidth}
        \centering
        \includegraphics[width=1.0\linewidth]{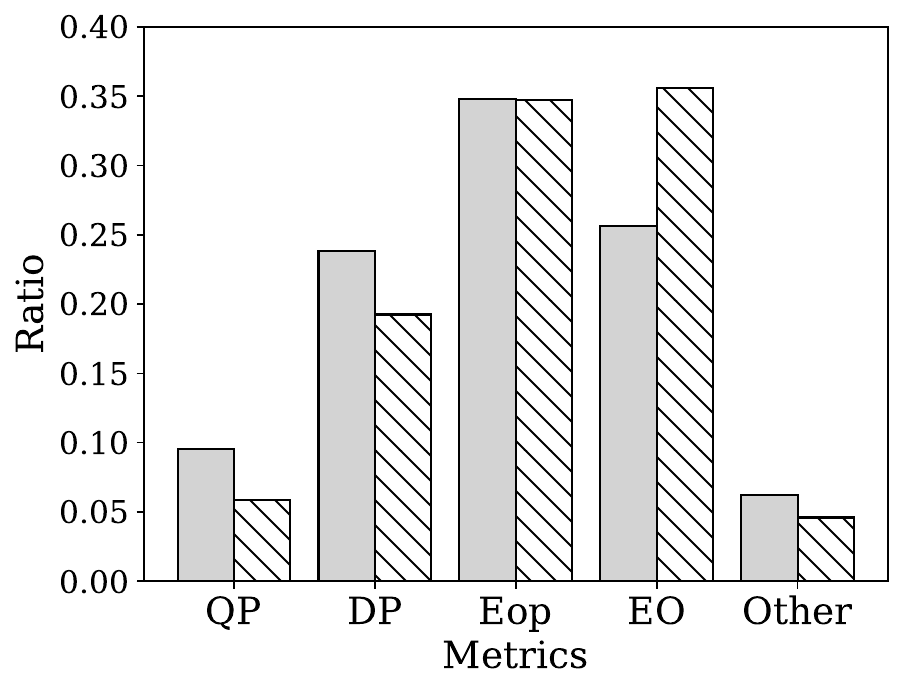}
       \subcaption{China***}
       \label{fig:hiring_boxplot_china}
    \end{minipage}
    \begin{minipage}[t]{0.24\linewidth}
        \centering
        \includegraphics[width=1.0\linewidth]{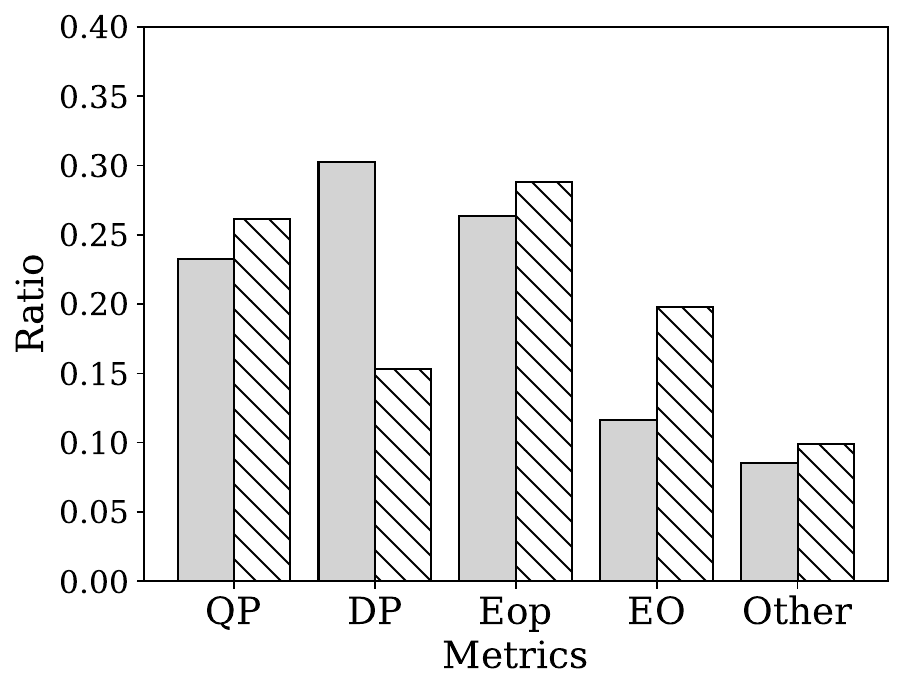}
        \subcaption{France***}
        \label{fig:hiring_boxplot_japan}
    \end{minipage}
    \begin{minipage}[t]{0.24\linewidth}
        \centering
        \includegraphics[width=1.0\linewidth]{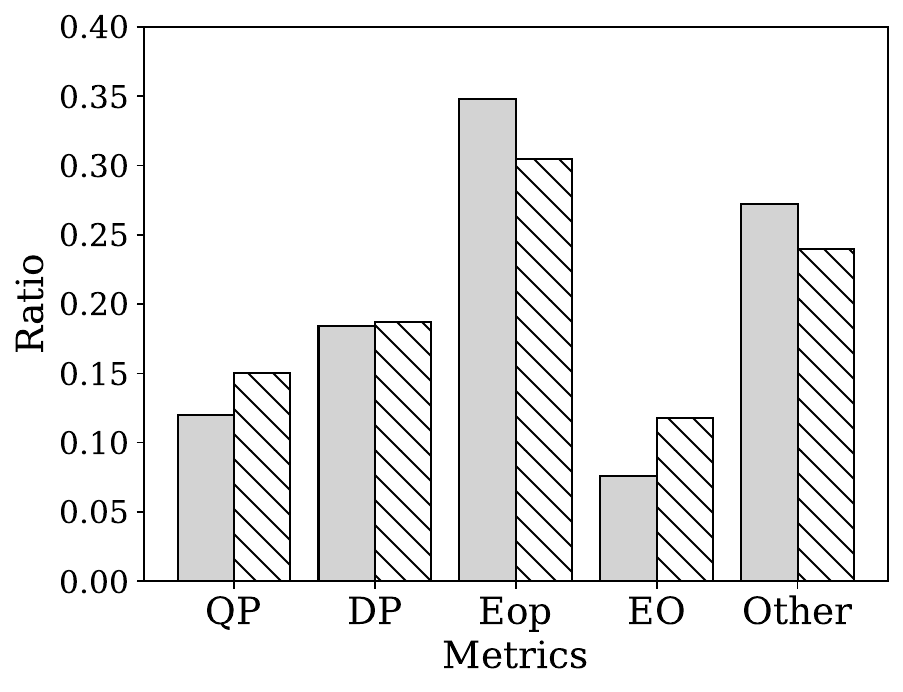}
        \subcaption{Japan*}
        \label{fig:hiring_boxplot_france}
    \end{minipage}
    \begin{minipage}[t]{0.24\linewidth}
    \centering
    \includegraphics[width=1.0\linewidth]{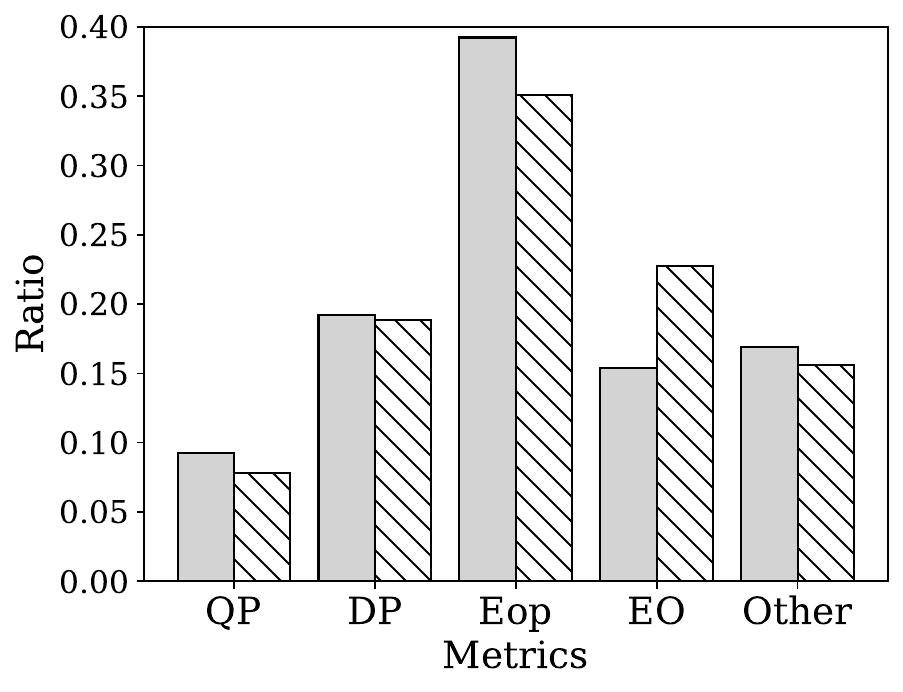}
    \subcaption{US}
    \label{fig:hiring_boxplot_us}
    \end{minipage}
\caption{Difference between genders in countries in employee award scenario}
\label{fig:employee_gender_county}
\end{figure*}

%% file: image/tex/result_ethnicity.tex
 \begin{figure*}[!t]
 \centering
    \begin{minipage}[t]{0.32\linewidth}
        \centering
        \includegraphics[width=1.0\linewidth]{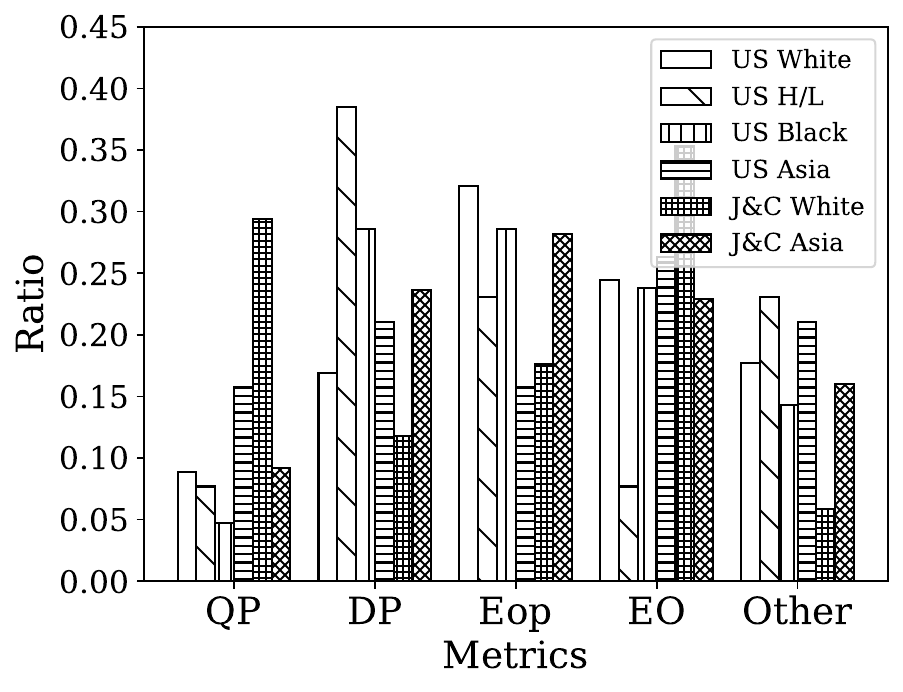}
        \subcaption{Art project: Asia**}
        \label{fig:art_age}
    \end{minipage}
    \begin{minipage}[t]{0.32\linewidth}
        \centering
        \includegraphics[width=1.0\linewidth]{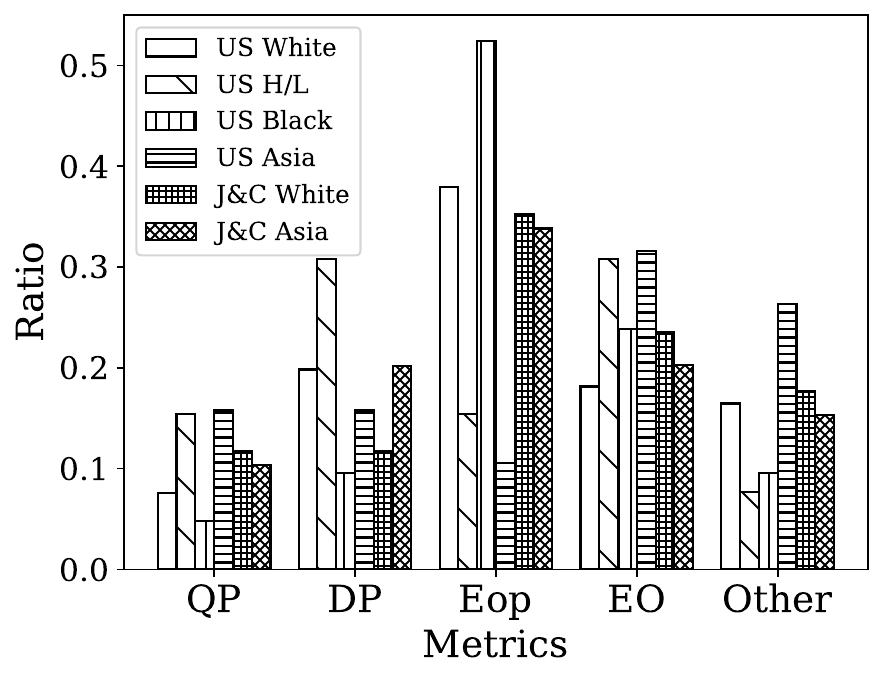}
        \subcaption{Employee award: US*}
        \label{fig:employee_age}
    \end{minipage}
\caption{Difference between ethnicity; H/L indicates Hispanic or Latinx.}
\label{fig:ethnicity}
\end{figure*}

%% file: image/tex/result_religions.tex
 \begin{figure*}[!t]
 \centering
    \begin{minipage}[t]{0.32\linewidth}
        \centering
        \includegraphics[width=1.0\linewidth]{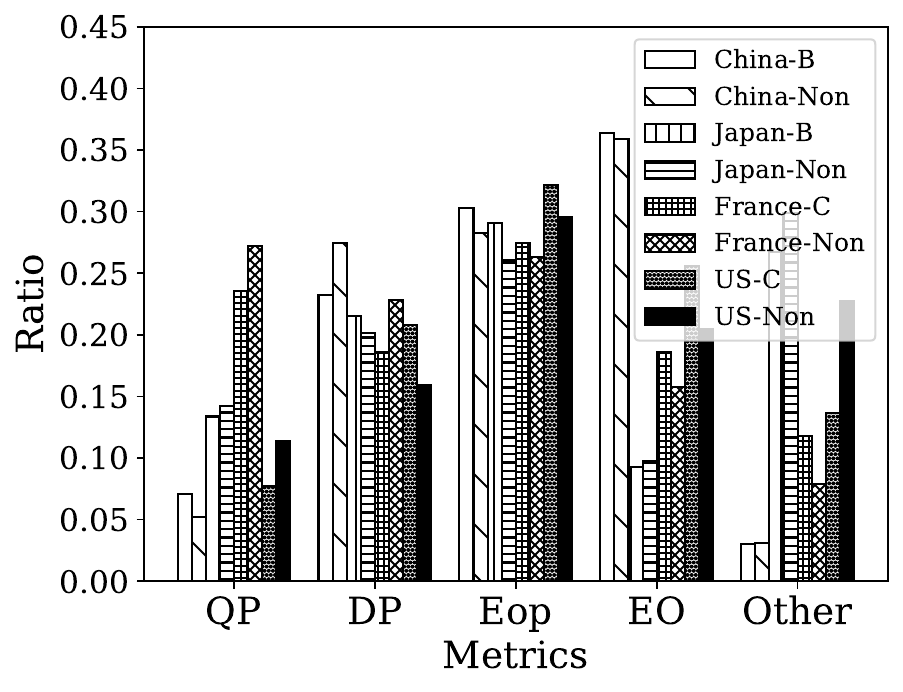}
        \subcaption{Art project: US*}
        \label{fig:art_age}
    \end{minipage}
    \begin{minipage}[t]{0.32\linewidth}
        \centering
        \includegraphics[width=1.0\linewidth]{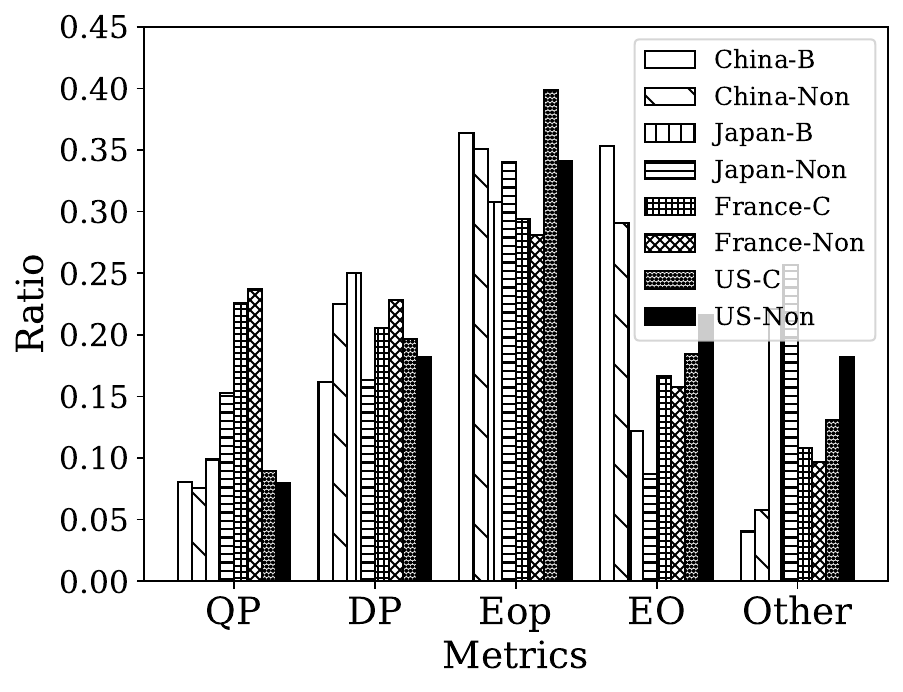}
        \subcaption{Employee award: Japan**}
        \label{fig:employee_age}
    \end{minipage}
\caption{Difference between religion; B, C, and Non indicate Buddhism, Christianity, and no religion, respectively.}
\label{fig:religion}
\end{figure*}

%% file: image/tex/result_age.tex
 \begin{figure*}[!t]
 \centering
    \begin{minipage}[t]{0.3\linewidth}
        \centering
        \includegraphics[width=1.0\linewidth]{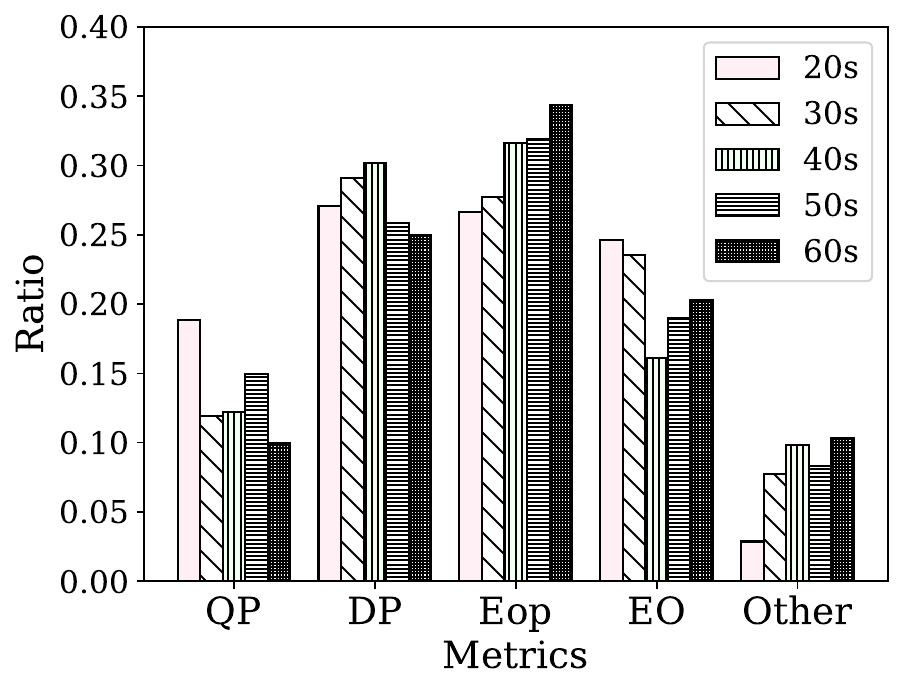}
       \subcaption{Hiring***}
       \label{fig:hiring_age}
    \end{minipage}
    \begin{minipage}[t]{0.30\linewidth}
        \centering
        \includegraphics[width=1.0\linewidth]{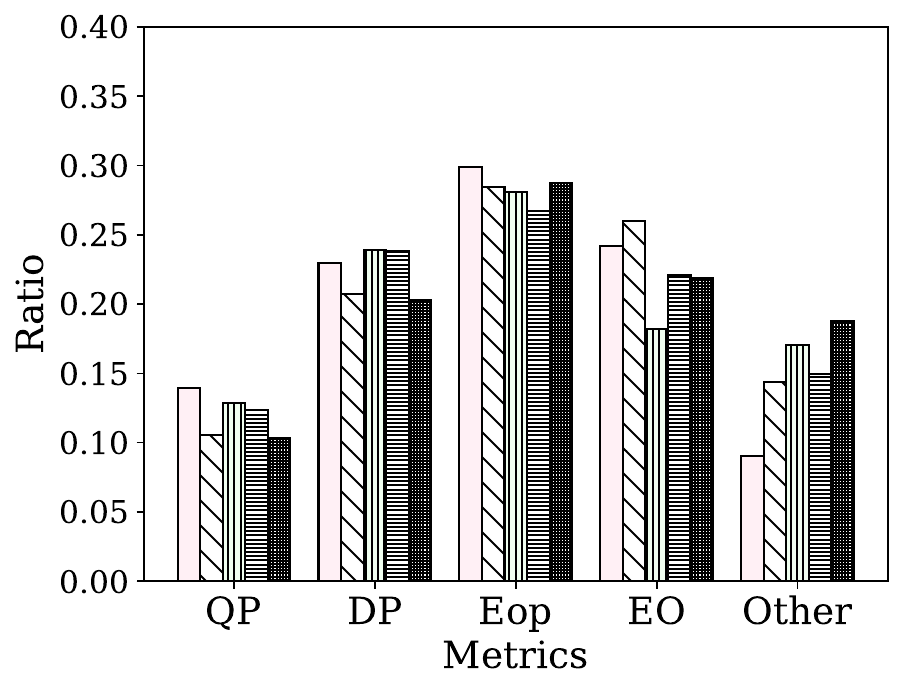}
        \subcaption{Art project*}
        \label{fig:art_age}
    \end{minipage}
    \begin{minipage}[t]{0.30\linewidth}
        \centering
        \includegraphics[width=1.0\linewidth]{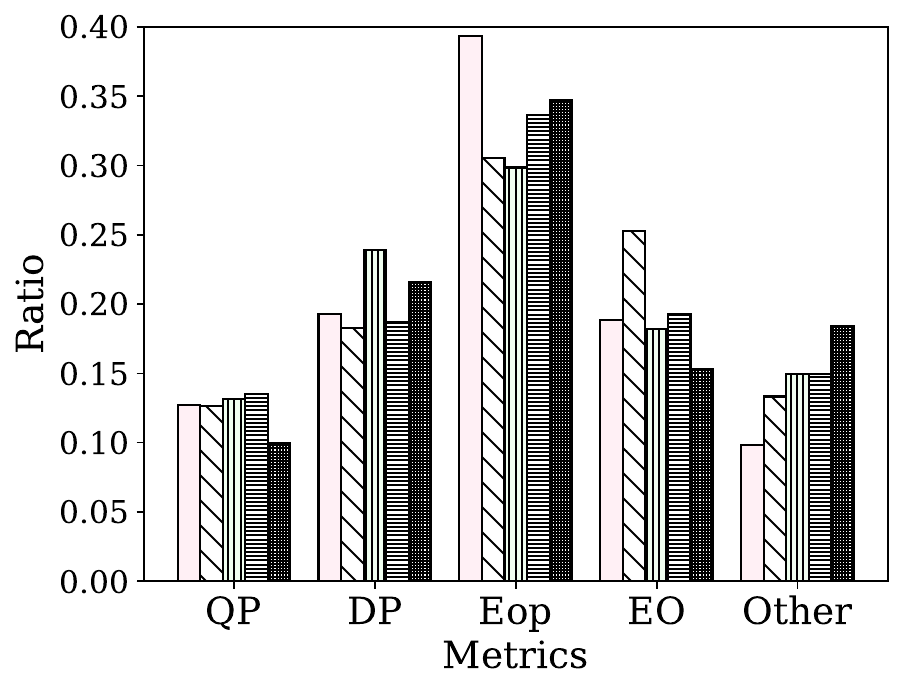}
        \subcaption{Employee award**}
        \label{fig:employee_age}
    \end{minipage}
\caption{Difference between age.}
\label{fig:age}
\end{figure*}


 \begin{figure*}[!t]
 \centering
    \begin{minipage}[t]{0.24\linewidth}
        \centering
        \includegraphics[width=1.0\linewidth]{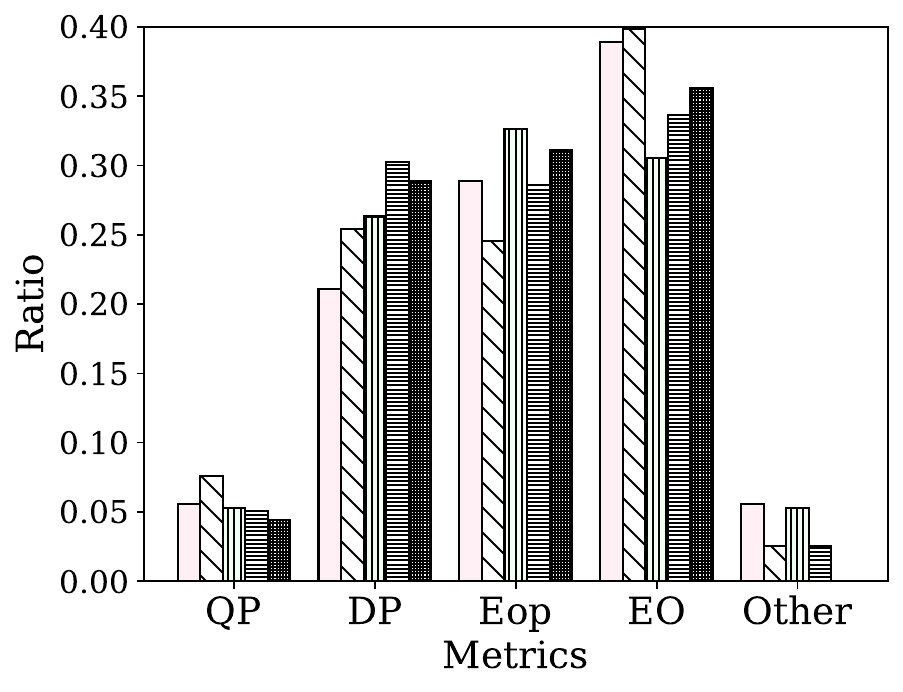}
       \subcaption{China}
       \label{fig:hiring_boxplot_china}
    \end{minipage}
    \begin{minipage}[t]{0.24\linewidth}
        \centering
        \includegraphics[width=1.0\linewidth]{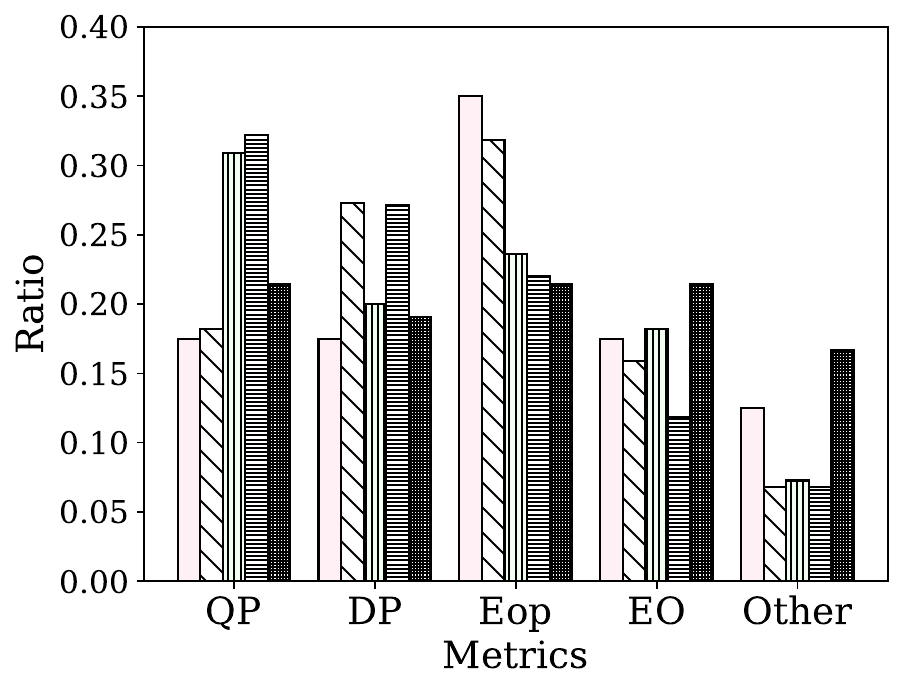}
        \subcaption{France}
        \label{fig:hiring_boxplot_japan}
    \end{minipage}
    \begin{minipage}[t]{0.24\linewidth}
        \centering
        \includegraphics[width=1.0\linewidth]{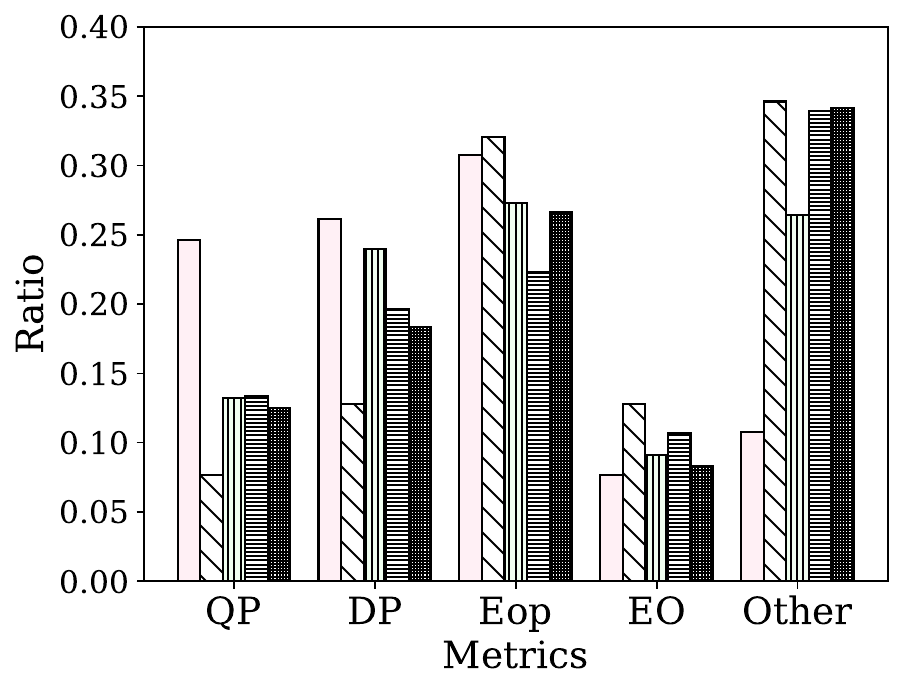}
        \subcaption{Japan**}
        \label{fig:hiring_boxplot_france}
    \end{minipage}
    \begin{minipage}[t]{0.24\linewidth}
    \centering
    \includegraphics[width=1.0\linewidth]{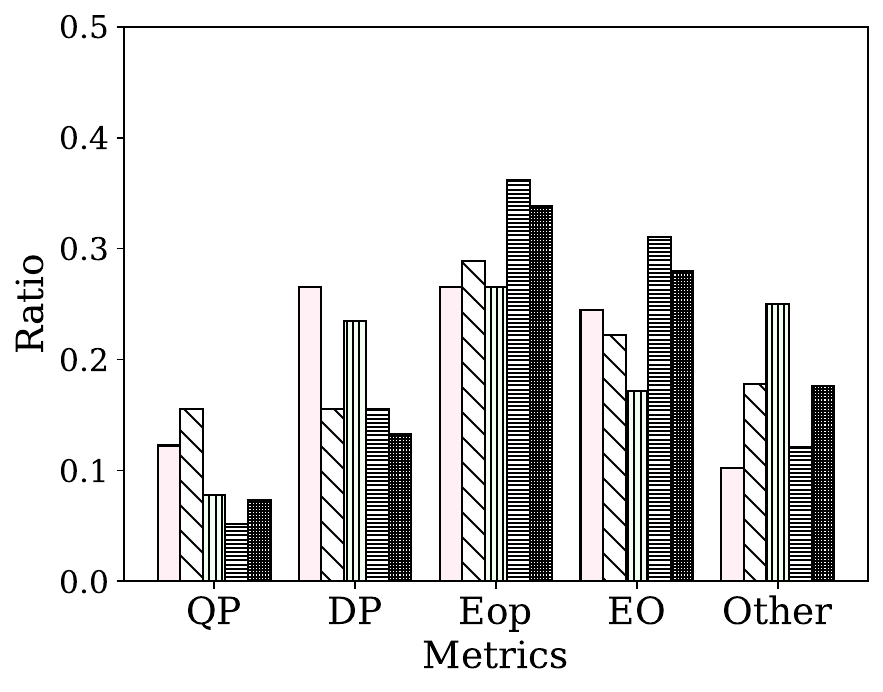}
    \subcaption{US*}
    \label{fig:hiring_boxplot_us}
    \end{minipage}
\caption{Difference between age in countries in art project scenario}
\label{fig:art_age_county}
\end{figure*}

 \begin{figure*}[!t]
 \centering
    \begin{minipage}[t]{0.24\linewidth}
        \centering
        \includegraphics[width=1.0\linewidth]{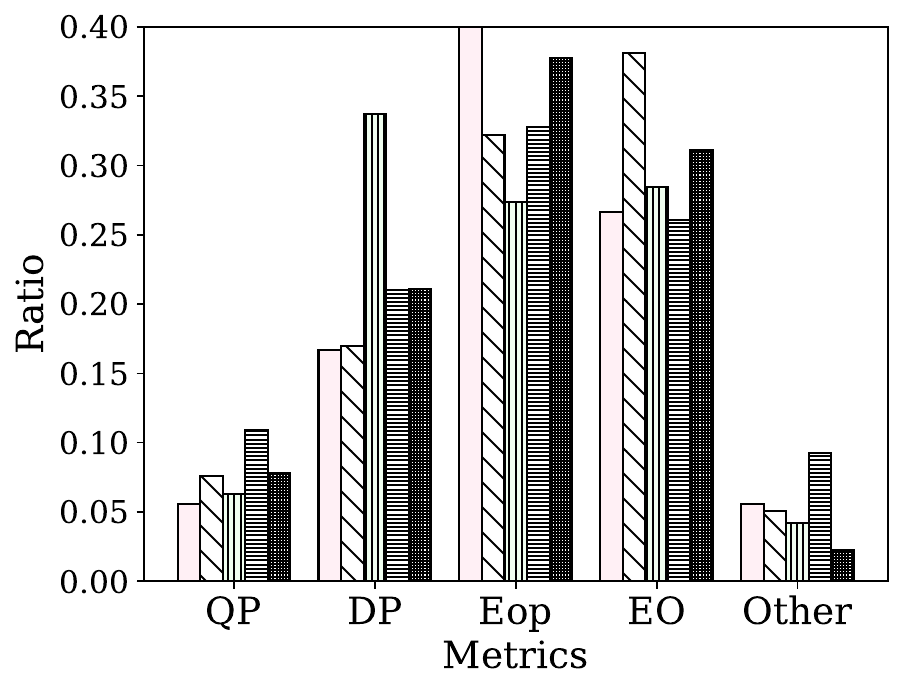}
       \subcaption{China**}
       \label{fig:hiring_boxplot_china}
    \end{minipage}
    \begin{minipage}[t]{0.24\linewidth}
        \centering
        \includegraphics[width=1.0\linewidth]{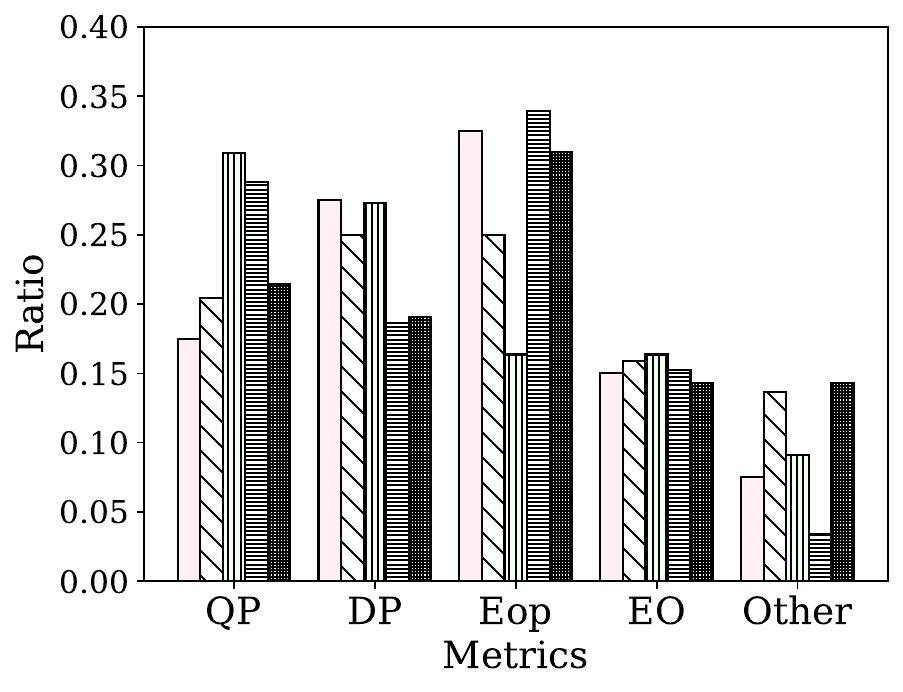}
        \subcaption{France}
        \label{fig:hiring_boxplot_japan}
    \end{minipage}
    \begin{minipage}[t]{0.24\linewidth}
        \centering
        \includegraphics[width=1.0\linewidth]{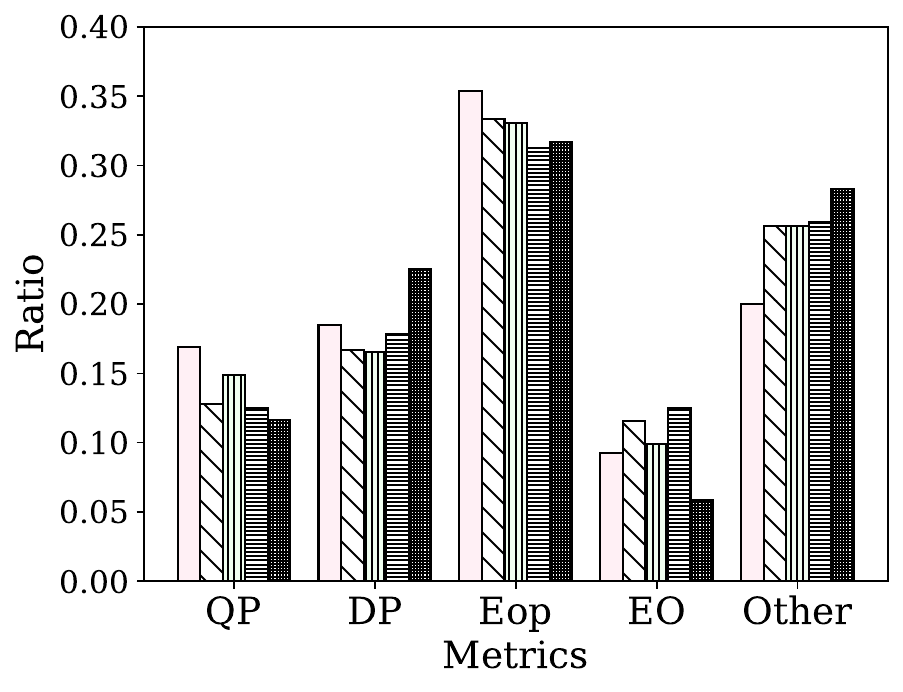}
        \subcaption{Japan}
        \label{fig:hiring_boxplot_france}
    \end{minipage}
    \begin{minipage}[t]{0.24\linewidth}
    \centering
    \includegraphics[width=1.0\linewidth]{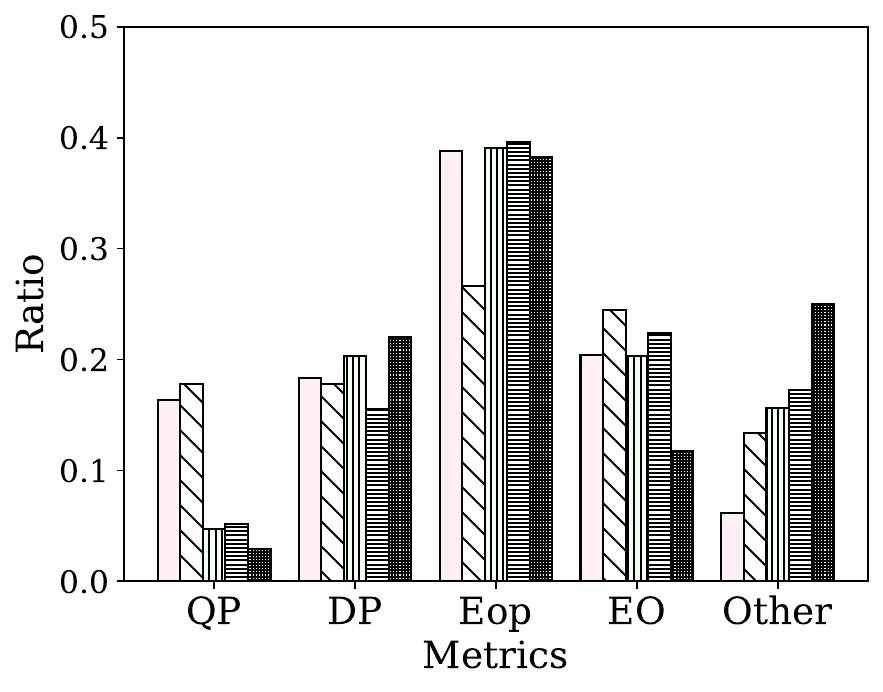}
    \subcaption{US**}
    \label{fig:hiring_boxplot_us}
    \end{minipage}
\caption{Difference between age in countries in employee award scenario}
\label{fig:employee_age_county}
\end{figure*}

%% file: image/tex/result_education.tex
 \begin{figure*}[!t]
 \centering
    \begin{minipage}[t]{0.3\linewidth}
        \centering
        \includegraphics[width=1.0\linewidth]{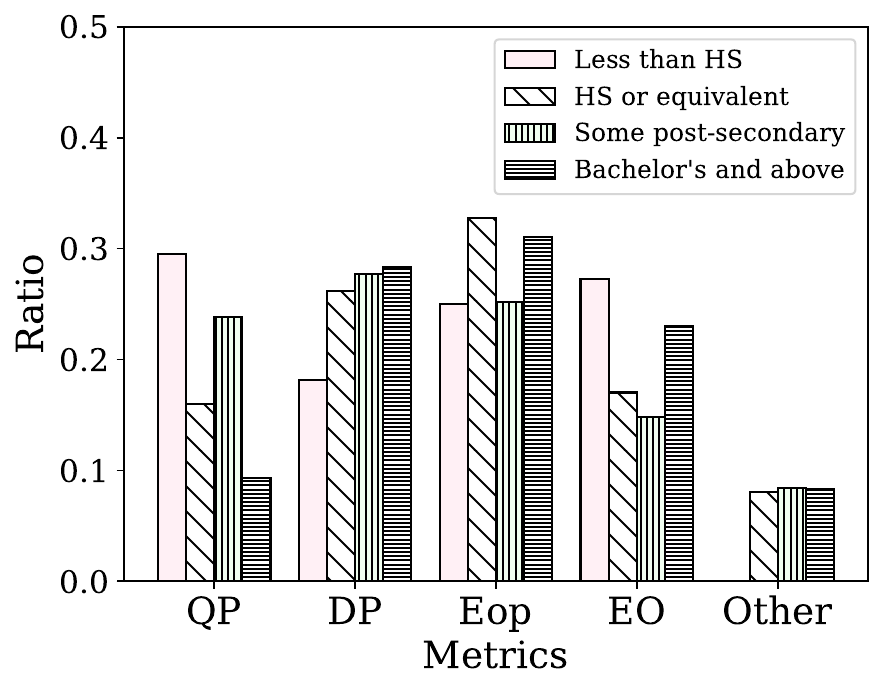}
       \subcaption{Hiring ***}
       \label{fig:hiring_age}
    \end{minipage}
    \begin{minipage}[t]{0.30\linewidth}
        \centering
        \includegraphics[width=1.0\linewidth]{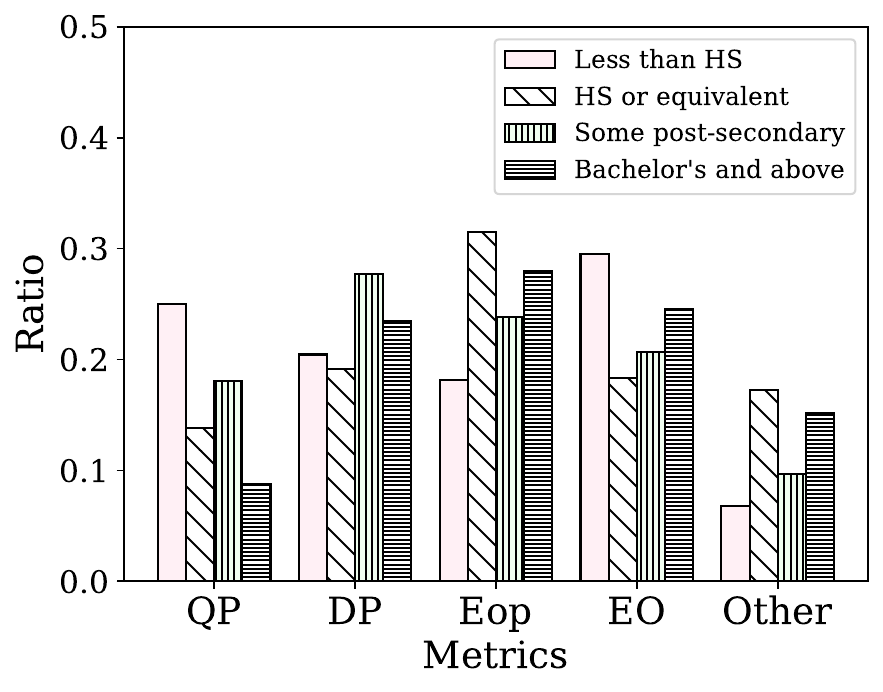}
        \subcaption{Art project ***}
        \label{fig:art_age}
    \end{minipage}
    \begin{minipage}[t]{0.30\linewidth}
        \centering
        \includegraphics[width=1.0\linewidth]{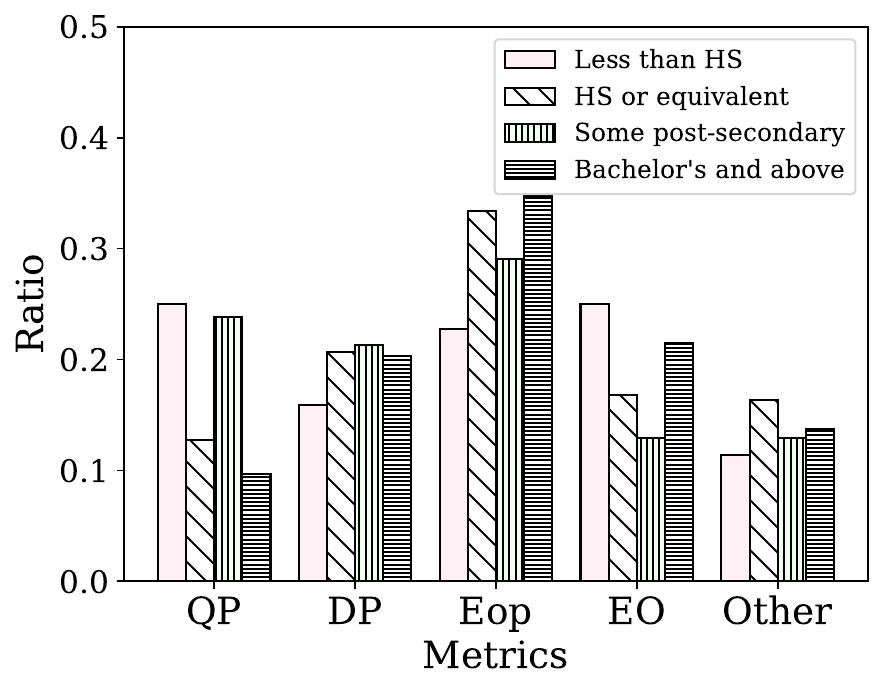}
        \subcaption{Employee award ***}
        \label{fig:employee_age}
    \end{minipage}
\caption{Difference between education.}
\label{fig:education}
\end{figure*}


 \begin{figure*}[!t]
 \centering
    \begin{minipage}[t]{0.24\linewidth}
        \centering
        \includegraphics[width=1.0\linewidth]{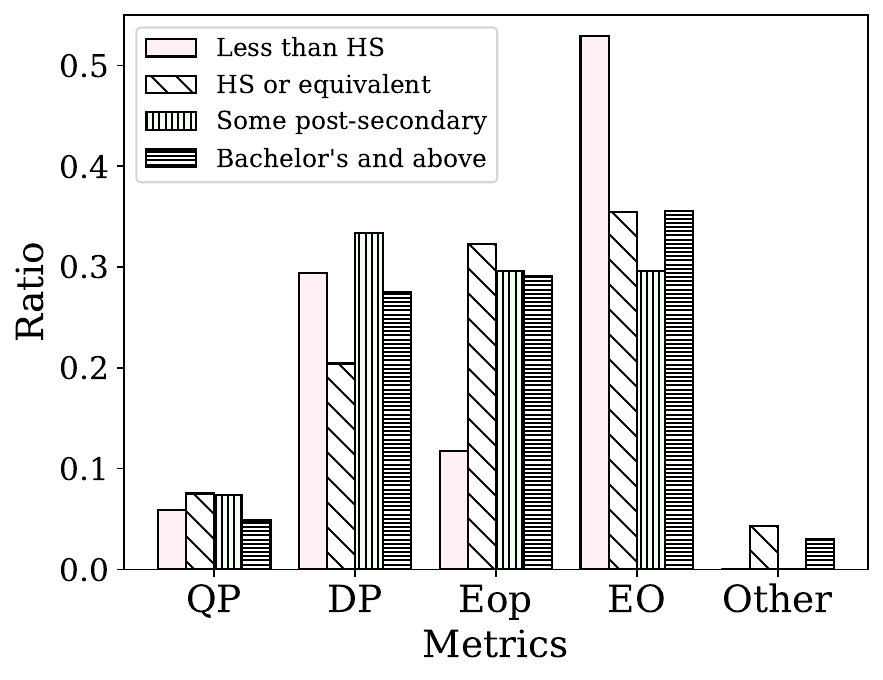}
       \subcaption{China}
       \label{fig:hiring_boxplot_china}
    \end{minipage}
    \begin{minipage}[t]{0.24\linewidth}
        \centering
        \includegraphics[width=1.0\linewidth]{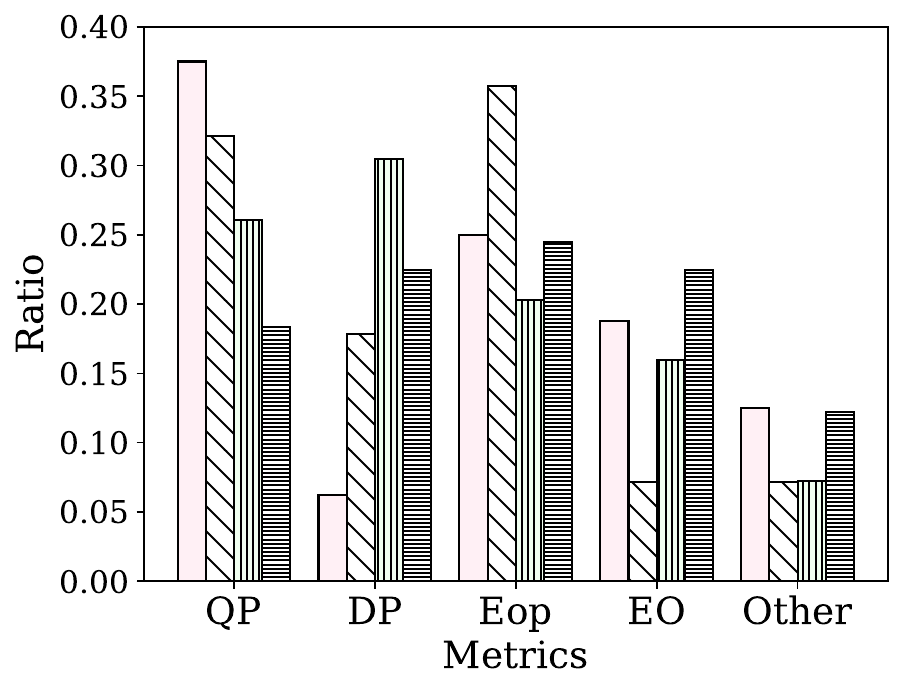}
        \subcaption{France*}
        \label{fig:hiring_boxplot_japan}
    \end{minipage}
    \begin{minipage}[t]{0.24\linewidth}
        \centering
        \includegraphics[width=1.0\linewidth]{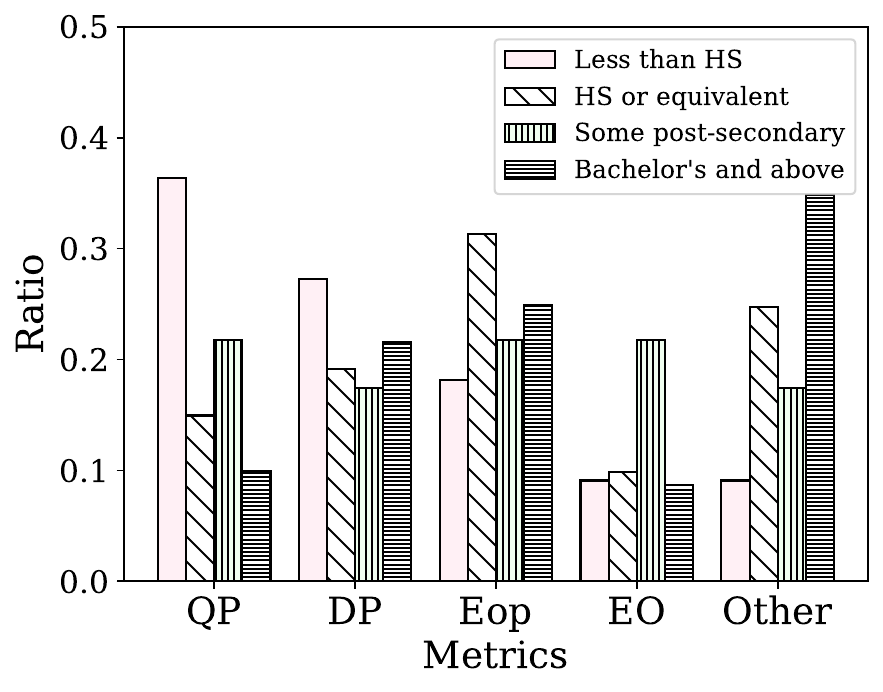}
        \subcaption{Japan**}
        \label{fig:hiring_boxplot_france}
    \end{minipage}
    \begin{minipage}[t]{0.24\linewidth}
    \centering
    \includegraphics[width=1.0\linewidth]{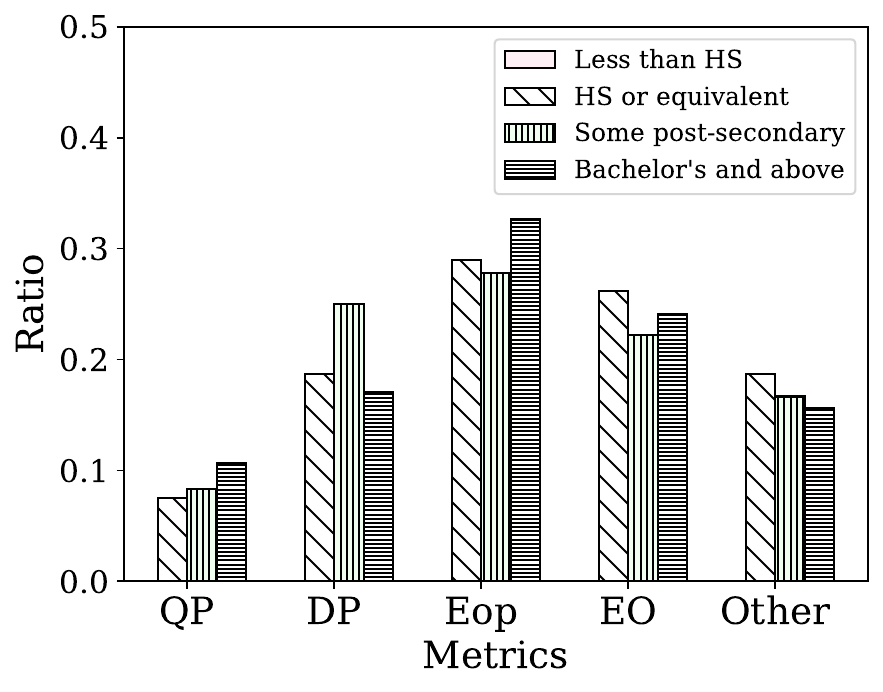}
    \subcaption{US}
    \label{fig:hiring_boxplot_us}
    \end{minipage}
\caption{Difference between education in countries in art project scenario}
\label{fig:art_education_county}
\end{figure*}

 \begin{figure*}[!t]
 \centering
    \begin{minipage}[t]{0.24\linewidth}
        \centering
        \includegraphics[width=1.0\linewidth]{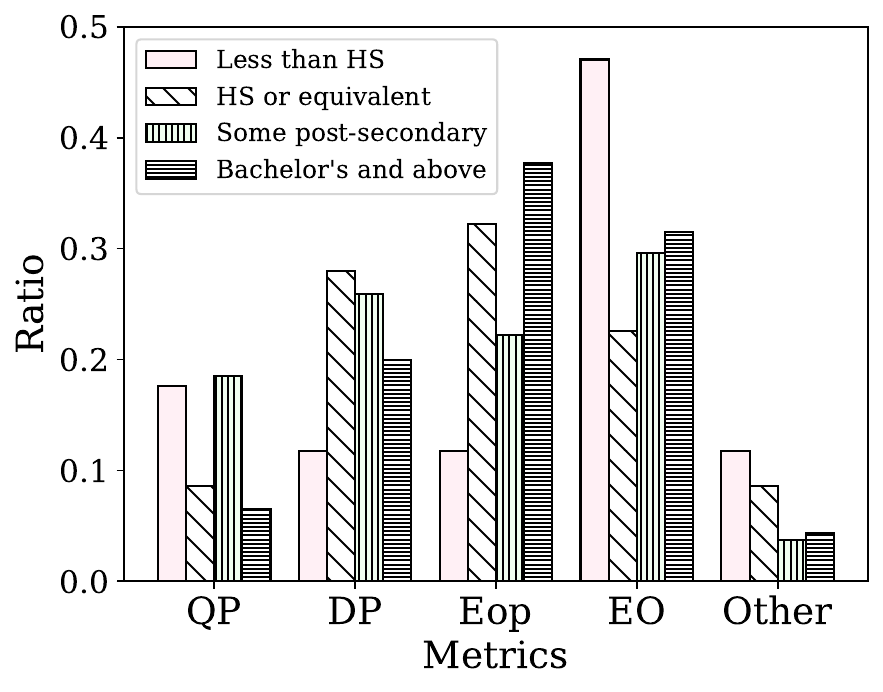}
       \subcaption{China**}
       \label{fig:hiring_boxplot_china}
    \end{minipage}
    \begin{minipage}[t]{0.24\linewidth}
        \centering
        \includegraphics[width=1.0\linewidth]{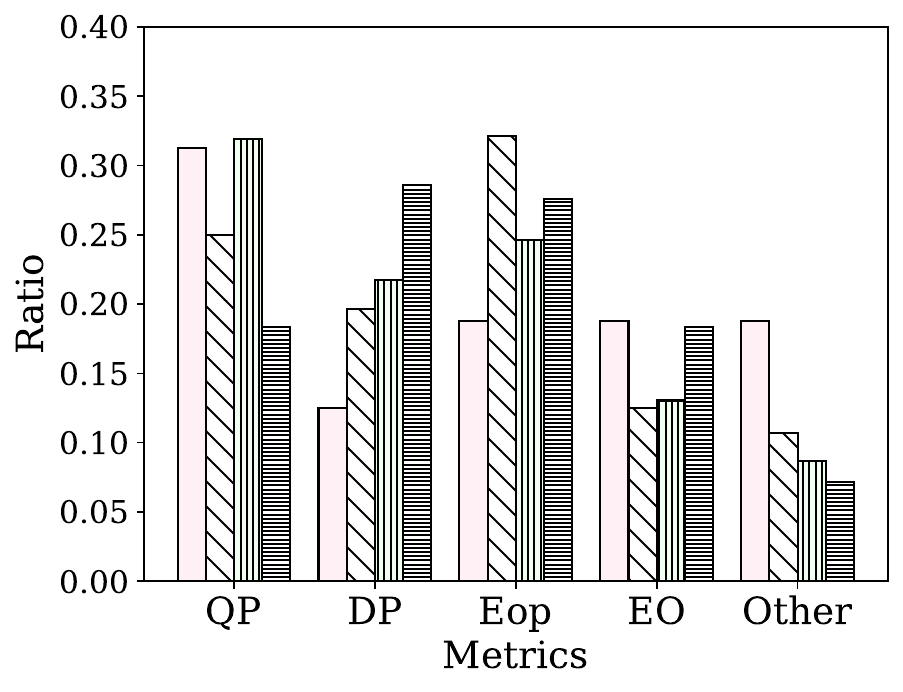}
        \subcaption{France}
        \label{fig:hiring_boxplot_japan}
    \end{minipage}
    \begin{minipage}[t]{0.24\linewidth}
        \centering
        \includegraphics[width=1.0\linewidth]{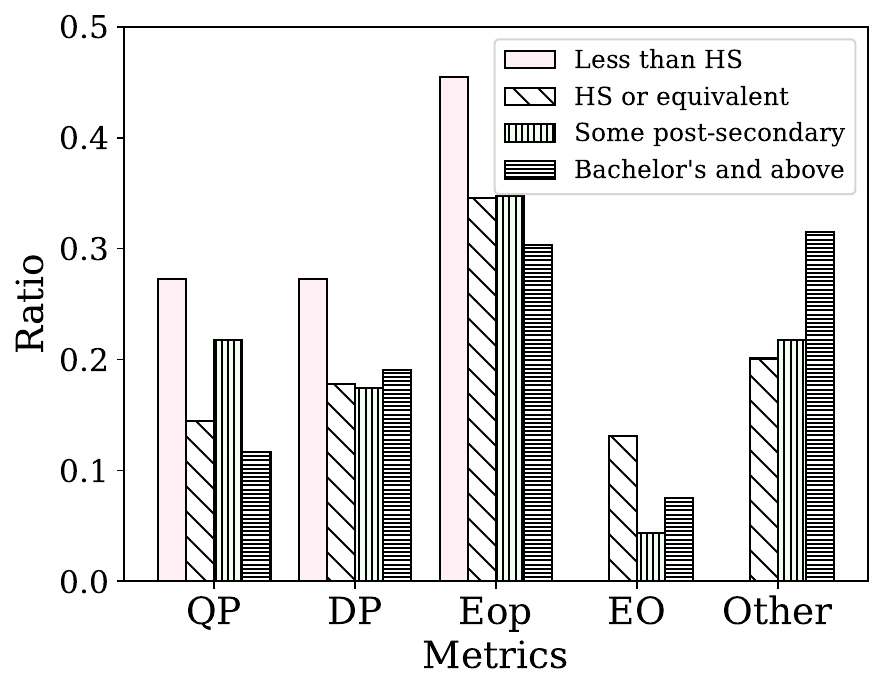}
        \subcaption{Japan}
        \label{fig:hiring_boxplot_france}
    \end{minipage}
    \begin{minipage}[t]{0.24\linewidth}
    \centering
    \includegraphics[width=1.0\linewidth]{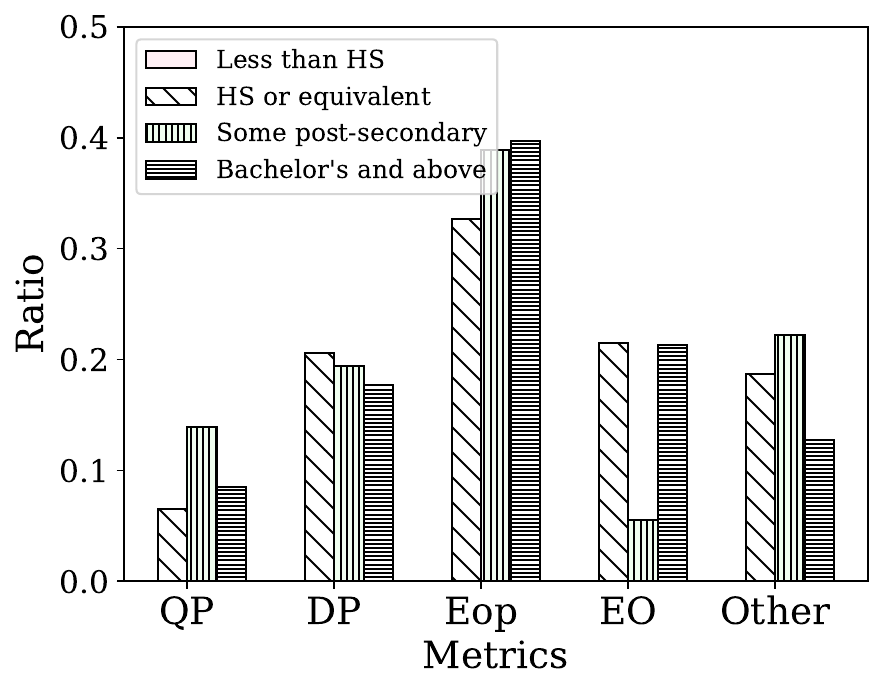}
    \subcaption{US**}
    \label{fig:hiring_boxplot_us}
    \end{minipage}
\caption{Difference between experience in countries in employee award scenario}
\label{fig:employee_education_county}
\end{figure*}

%% file: image/tex/result_experience.tex
 \begin{figure*}[!t]
 \centering
    \begin{minipage}[t]{0.3\linewidth}
        \centering
        \includegraphics[width=1.0\linewidth]{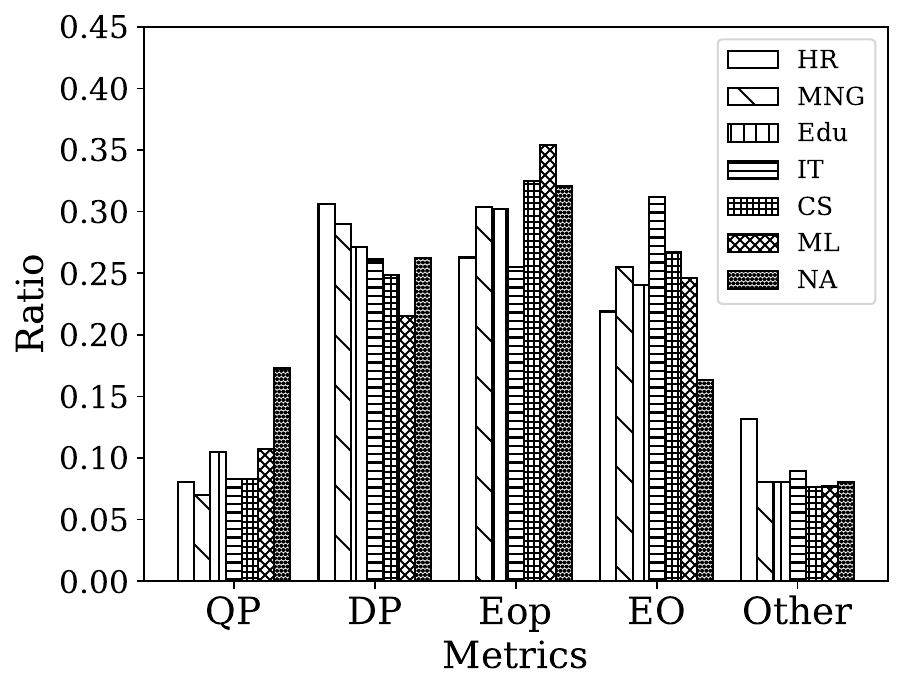}
       \subcaption{Hiring***}
       \label{fig:hiring_age}
    \end{minipage}
    \begin{minipage}[t]{0.30\linewidth}
        \centering
        \includegraphics[width=1.0\linewidth]{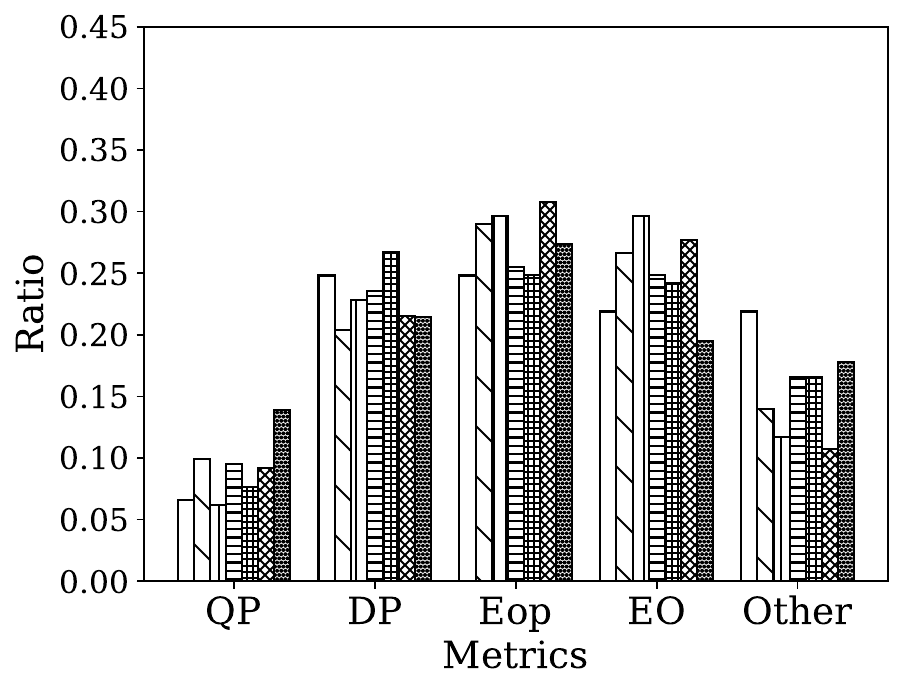}
        \subcaption{Art project**}
        \label{fig:art_age}
    \end{minipage}
    \begin{minipage}[t]{0.30\linewidth}
        \centering
        \includegraphics[width=1.0\linewidth]{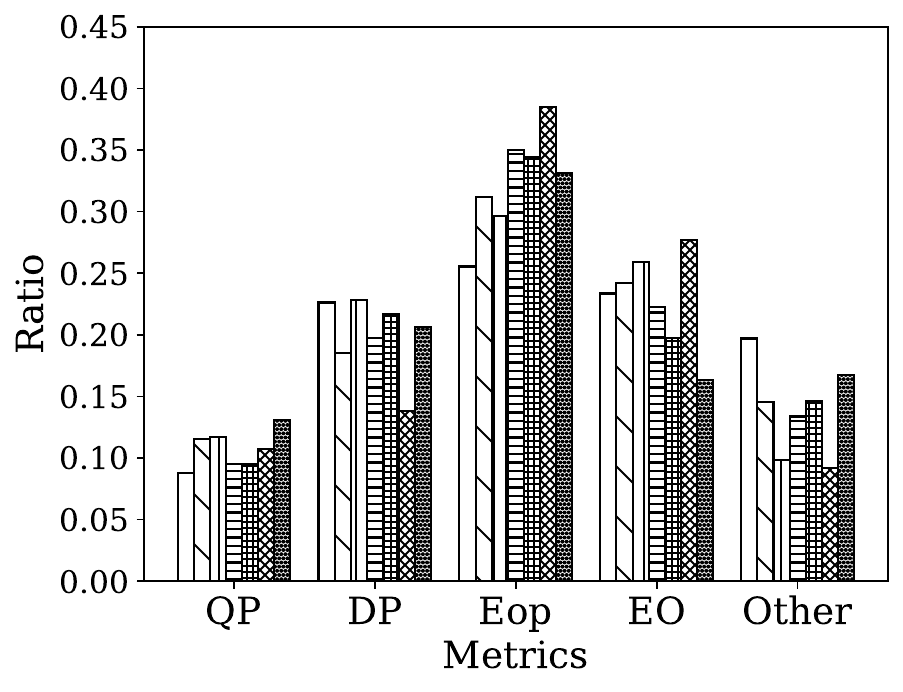}
        \subcaption{Employee award*}
        \label{fig:employee_age}
    \end{minipage}
\caption{Difference between experience.}
\label{fig:experience}
\end{figure*}


 \begin{figure*}[!t]
 \centering
    \begin{minipage}[t]{0.24\linewidth}
        \centering
        \includegraphics[width=1.0\linewidth]{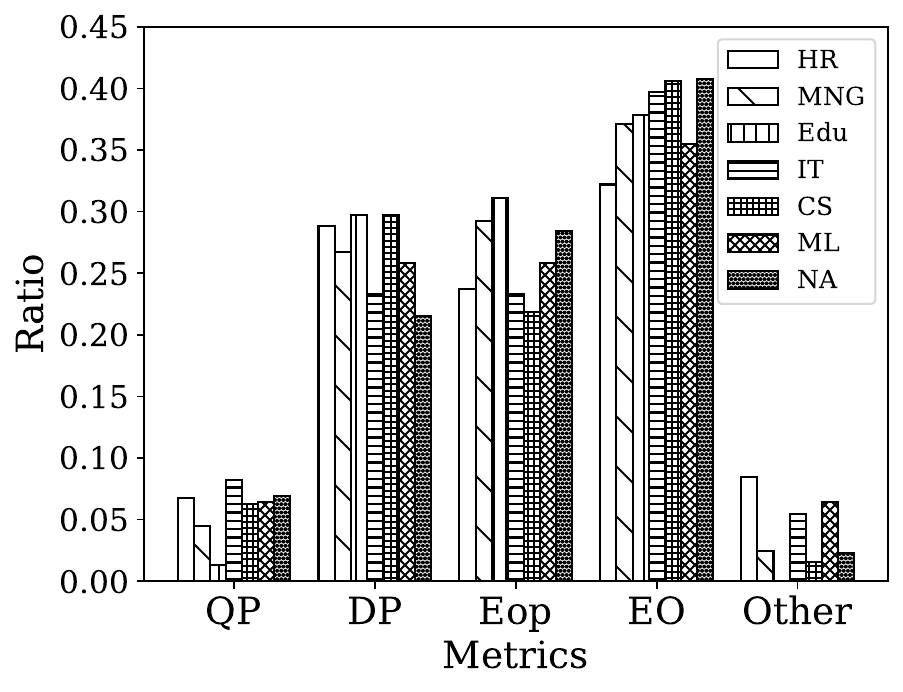}
       \subcaption{China}
       \label{fig:hiring_boxplot_china}
    \end{minipage}
    \begin{minipage}[t]{0.24\linewidth}
        \centering
        \includegraphics[width=1.0\linewidth]{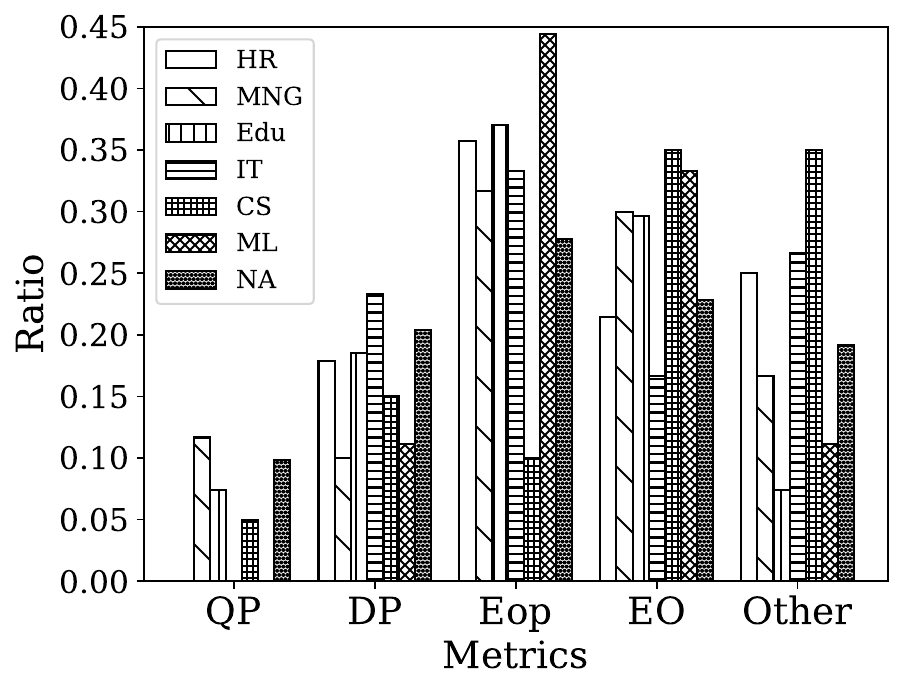}
        \subcaption{France}
        \label{fig:hiring_boxplot_japan}
    \end{minipage}
    \begin{minipage}[t]{0.24\linewidth}
        \centering
        \includegraphics[width=1.0\linewidth]{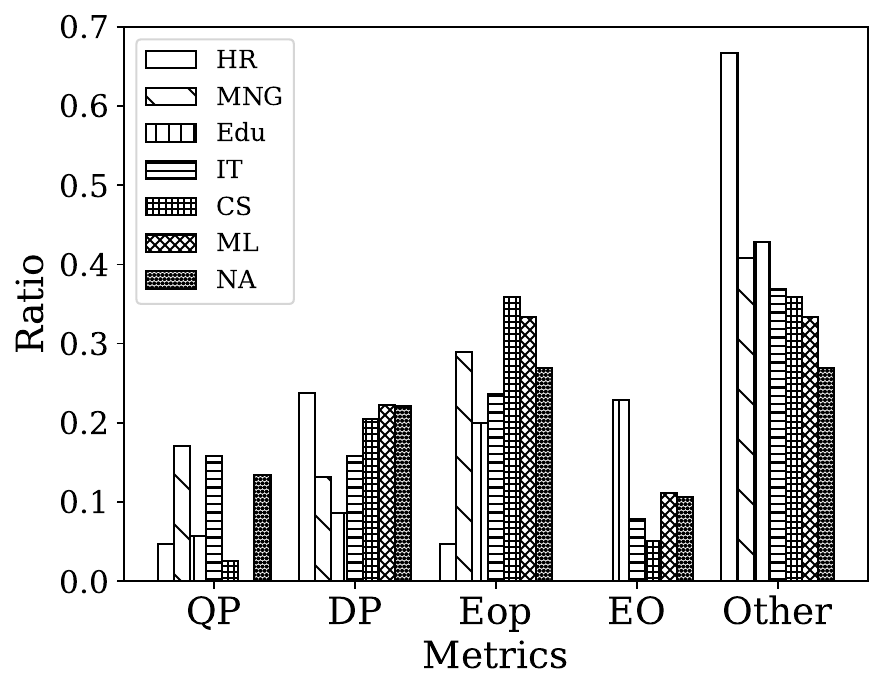}
        \subcaption{Japan***}
        \label{fig:hiring_boxplot_france}
    \end{minipage}
    \begin{minipage}[t]{0.24\linewidth}
    \centering
    \includegraphics[width=1.0\linewidth]{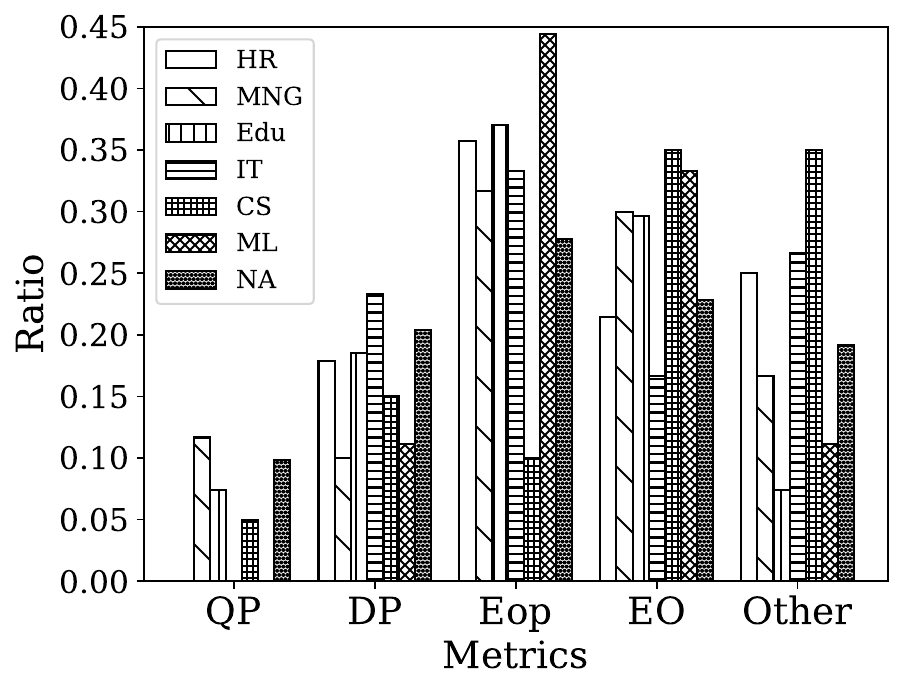}
    \subcaption{US*}
    \label{fig:hiring_boxplot_us}
    \end{minipage}
\caption{Difference between experience in countries in art project scenario}
\label{fig:art_experience_county}
\end{figure*}

 \begin{figure*}[!t]
 \centering
    \begin{minipage}[t]{0.24\linewidth}
        \centering
        \includegraphics[width=1.0\linewidth]{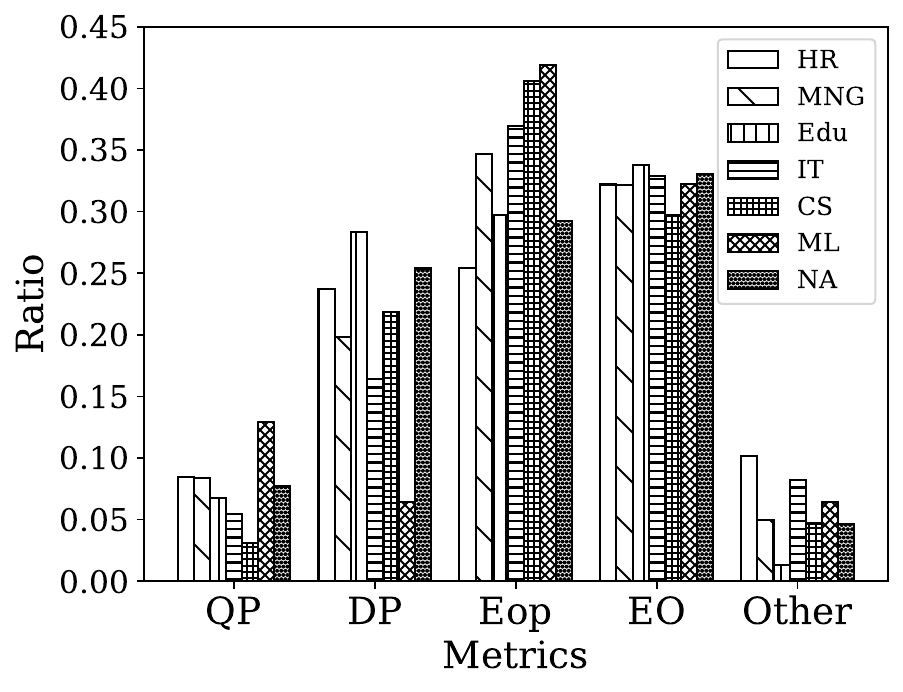}
       \subcaption{China}
       \label{fig:hiring_boxplot_china}
    \end{minipage}
    \begin{minipage}[t]{0.24\linewidth}
        \centering
        \includegraphics[width=1.0\linewidth]{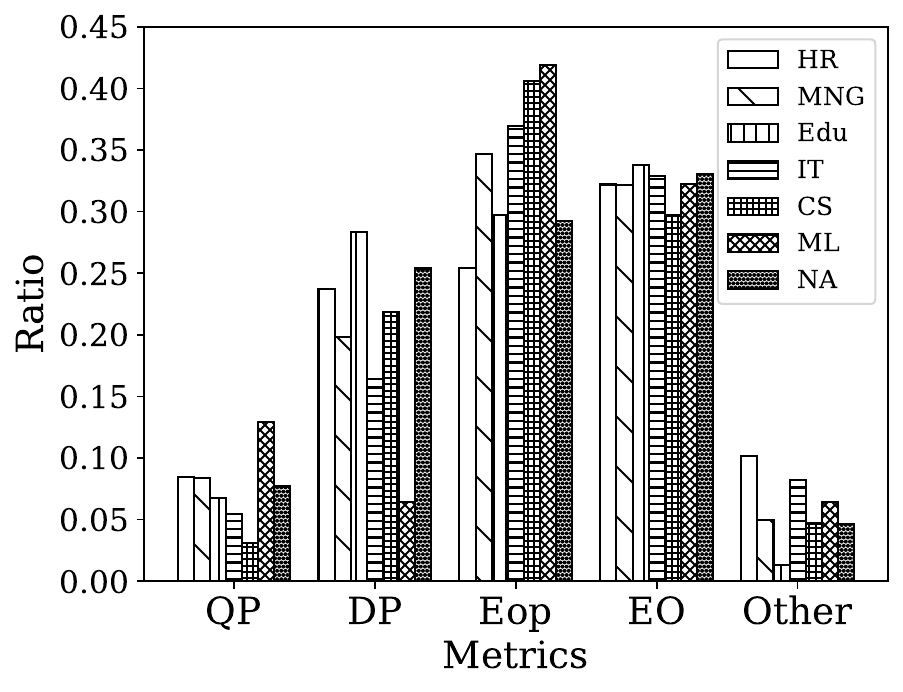}
        \subcaption{France}
        \label{fig:hiring_boxplot_japan}
    \end{minipage}
    \begin{minipage}[t]{0.24\linewidth}
        \centering
        \includegraphics[width=1.0\linewidth]{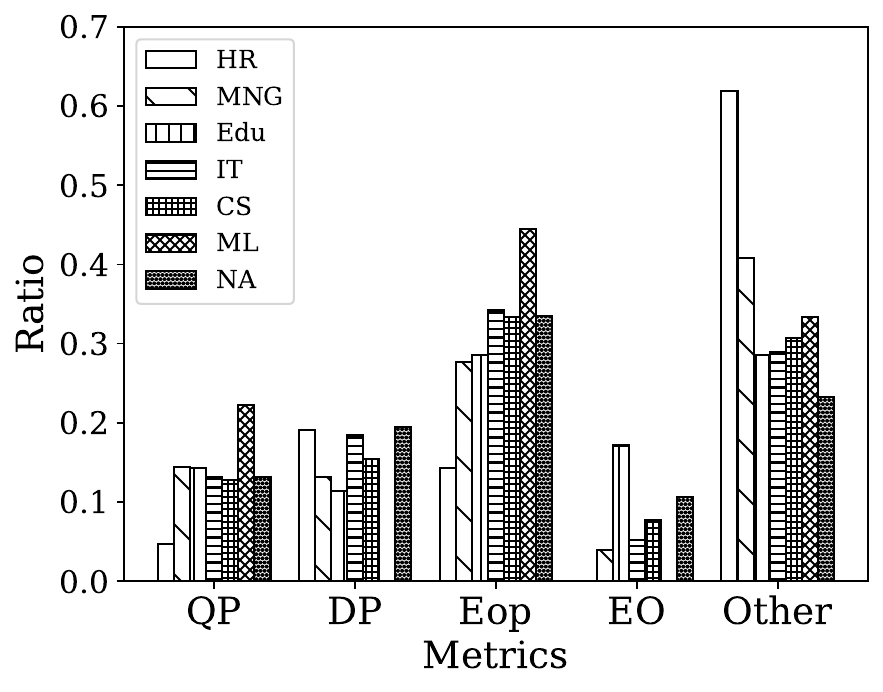}
        \subcaption{Japan**}
        \label{fig:hiring_boxplot_france}
    \end{minipage}
    \begin{minipage}[t]{0.24\linewidth}
    \centering
    \includegraphics[width=1.0\linewidth]{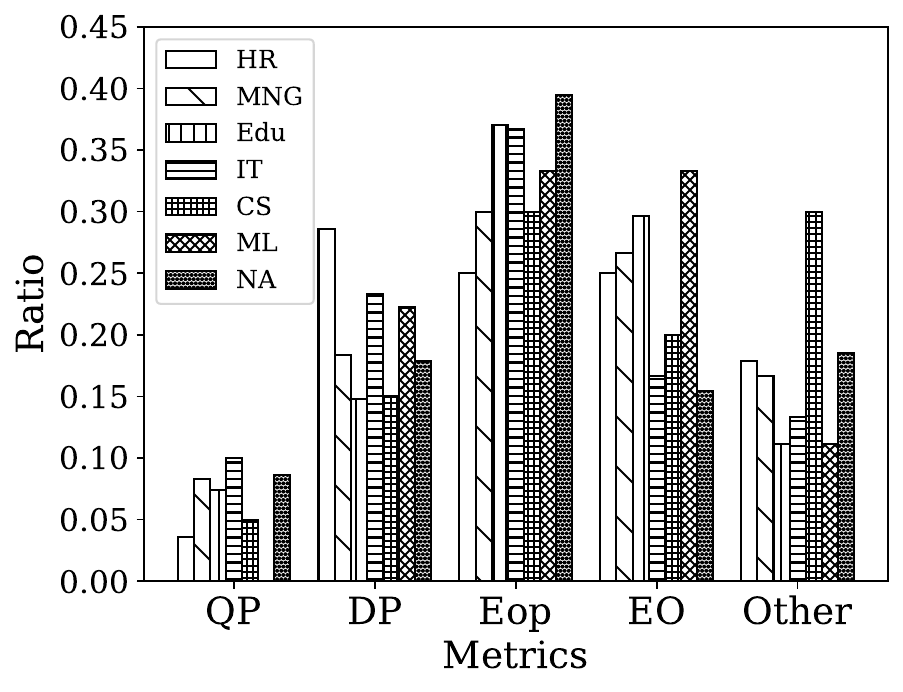}
    \subcaption{US}
    \label{fig:hiring_boxplot_us}
    \end{minipage}
\caption{Difference between experience in countries in employee award scenario}
\label{fig:employee_experience_county}
\end{figure*}

%% file: image/tex/result_correctness.tex

 \begin{figure*}[!t]
 \centering
    \begin{minipage}[t]{0.24\linewidth}
        \centering
        \includegraphics[width=1.0\linewidth]{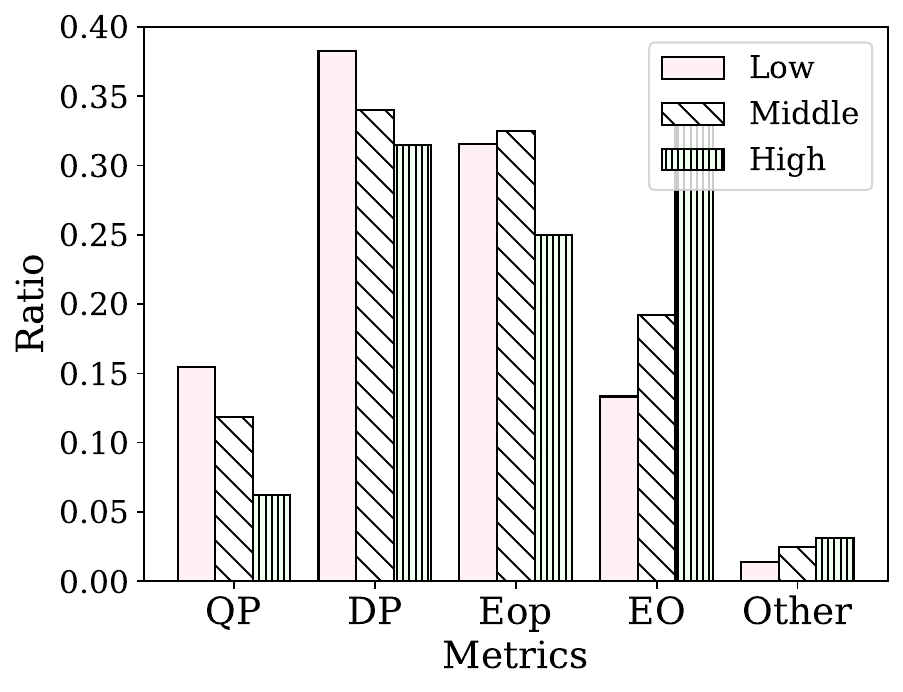}
       \subcaption{China***}
       \label{fig:hiring_boxplot_china}
    \end{minipage}
    \begin{minipage}[t]{0.24\linewidth}
        \centering
        \includegraphics[width=1.0\linewidth]{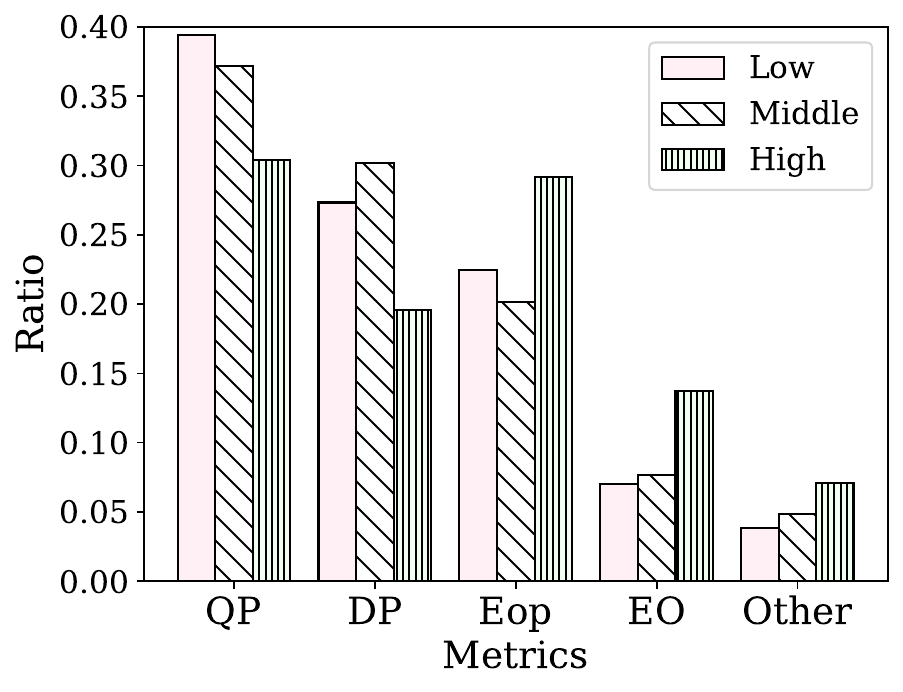}
        \subcaption{France**}
        \label{fig:hiring_boxplot_japan}
    \end{minipage}
    \begin{minipage}[t]{0.24\linewidth}
        \centering
        \includegraphics[width=1.0\linewidth]{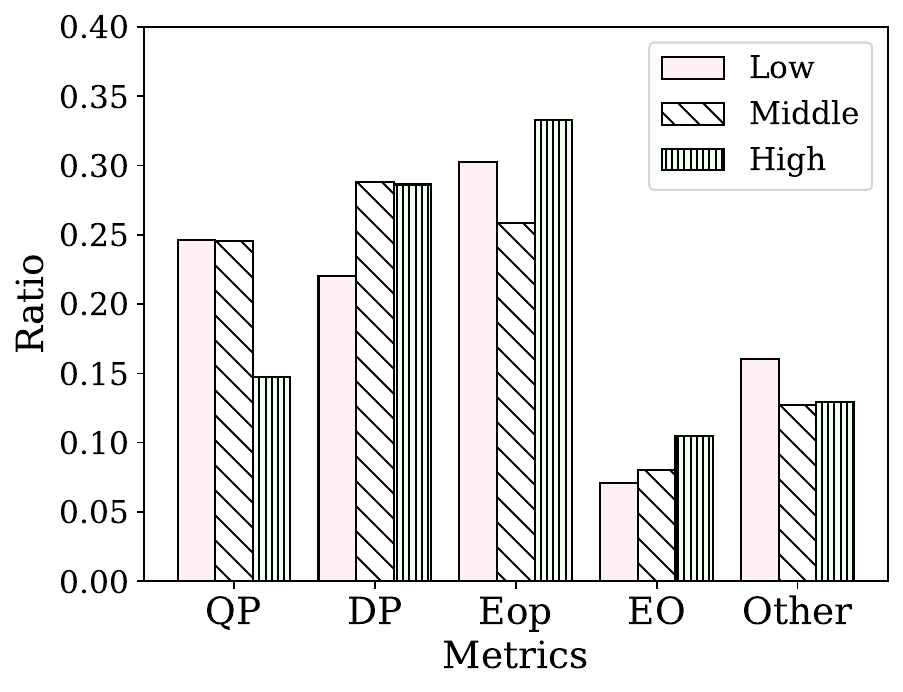}
        \subcaption{Japan***}
        \label{fig:hiring_boxplot_france}
    \end{minipage}
    \begin{minipage}[t]{0.24\linewidth}
    \centering
    \includegraphics[width=1.0\linewidth]{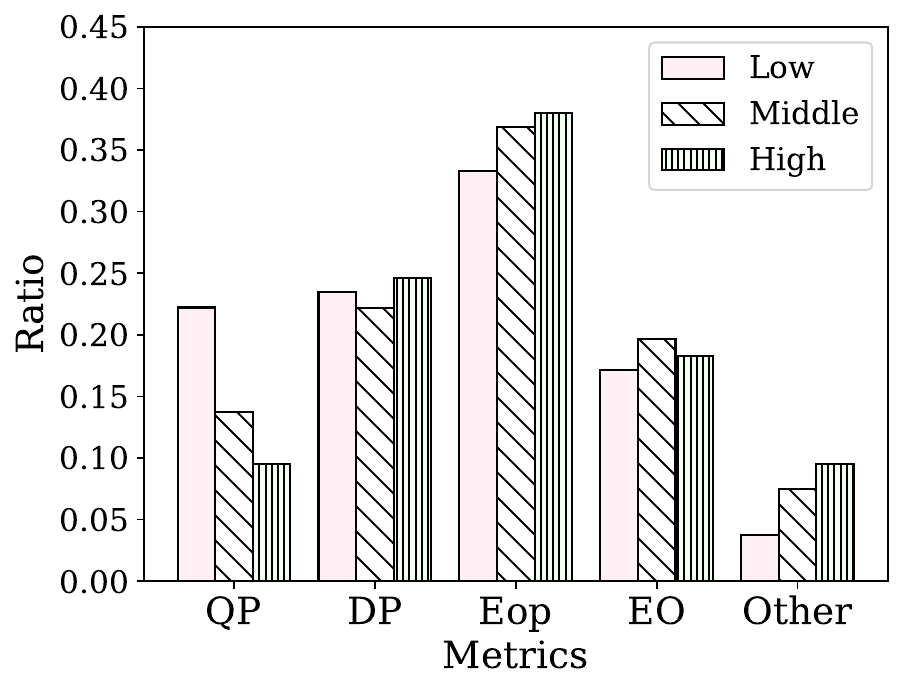}
    \subcaption{US***}
    \label{fig:hiring_boxplot_us}
    \end{minipage}
\caption{Difference between correctness in countries in hiring scenario}
\label{fig:art_correctness_county}
\end{figure*}

 \begin{figure*}[!t]
 \centering
    \begin{minipage}[t]{0.24\linewidth}
        \centering
        \includegraphics[width=1.0\linewidth]{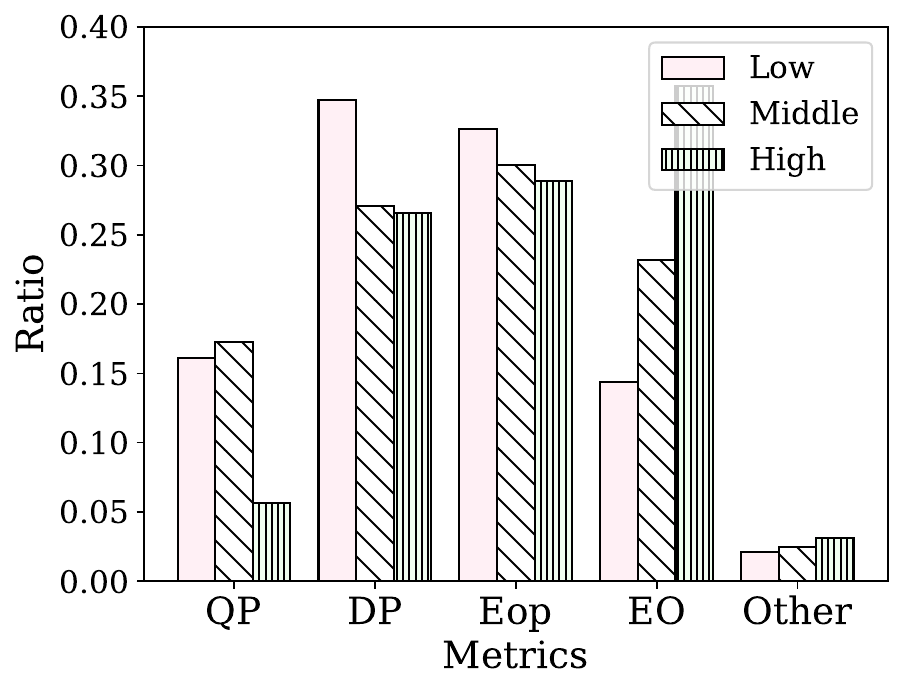}
       \subcaption{China***}
       \label{fig:hiring_boxplot_china}
    \end{minipage}
    \begin{minipage}[t]{0.24\linewidth}
        \centering
        \includegraphics[width=1.0\linewidth]{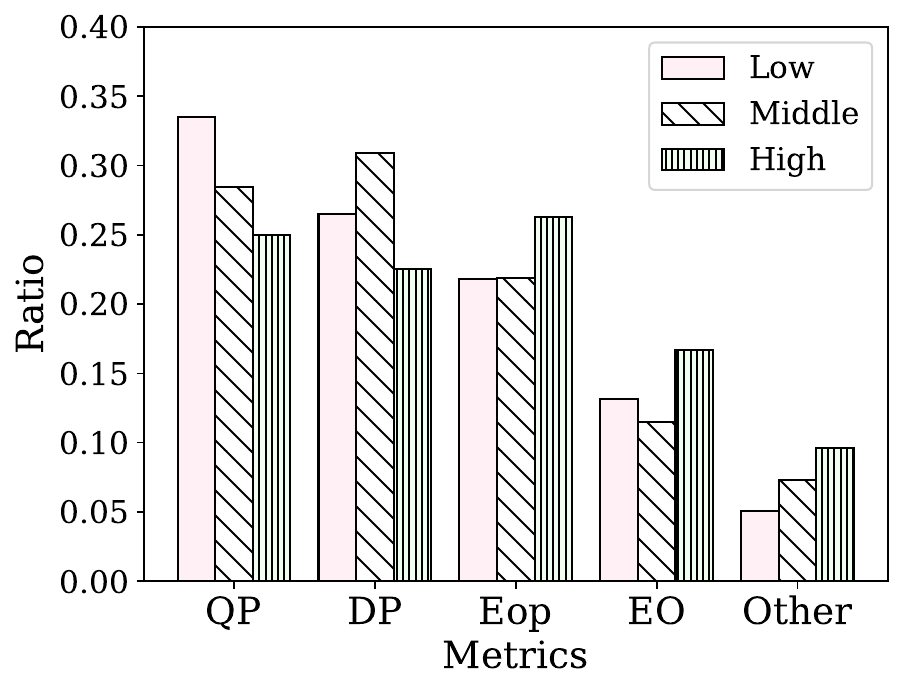}
        \subcaption{France***}
        \label{fig:hiring_boxplot_japan}
    \end{minipage}
    \begin{minipage}[t]{0.24\linewidth}
        \centering
        \includegraphics[width=1.0\linewidth]{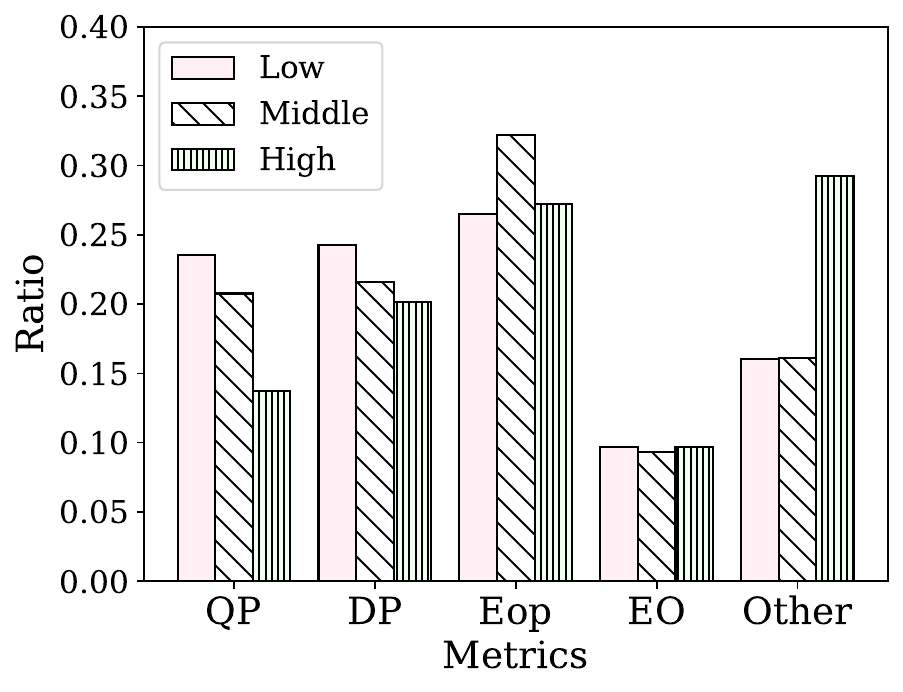}
        \subcaption{Japan***}
        \label{fig:hiring_boxplot_france}
    \end{minipage}
    \begin{minipage}[t]{0.24\linewidth}
    \centering
    \includegraphics[width=1.0\linewidth]{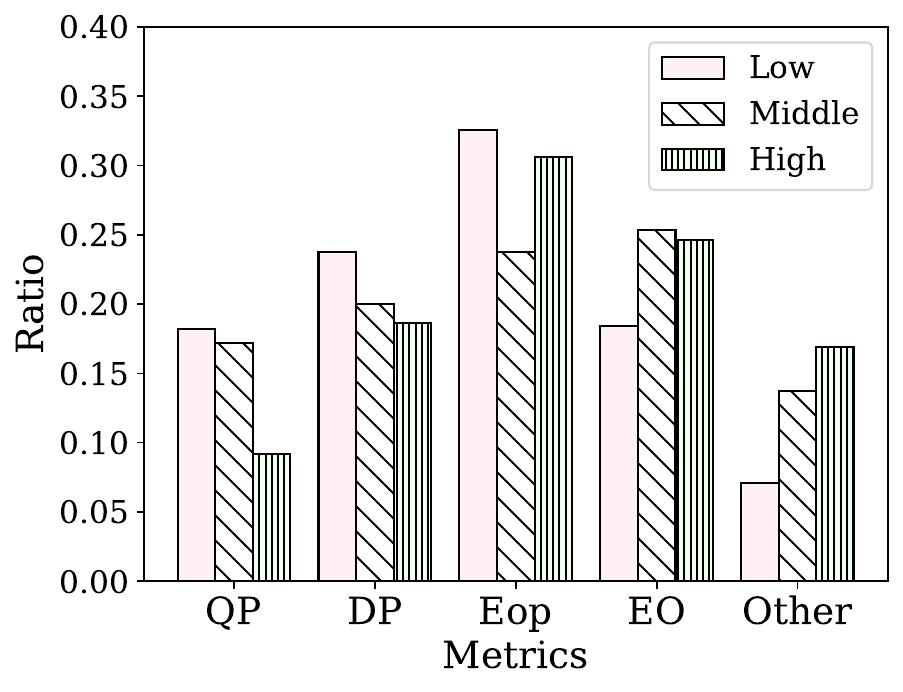}
    \subcaption{US***}
    \label{fig:hiring_boxplot_us}
    \end{minipage}
\caption{Difference between correctness in countries in art project scenario}
\label{fig:art_correctness_county}
\end{figure*}

 \begin{figure*}[!t]
 \centering
    \begin{minipage}[t]{0.24\linewidth}
        \centering
        \includegraphics[width=1.0\linewidth]{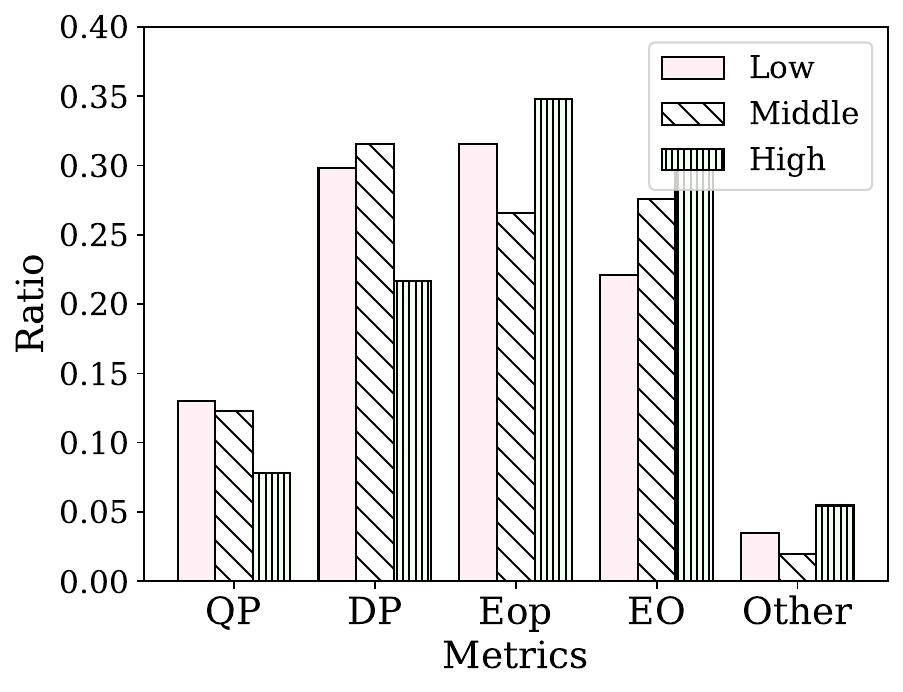}
       \subcaption{China***}
       \label{fig:hiring_boxplot_china}
    \end{minipage}
    \begin{minipage}[t]{0.24\linewidth}
        \centering
        \includegraphics[width=1.0\linewidth]{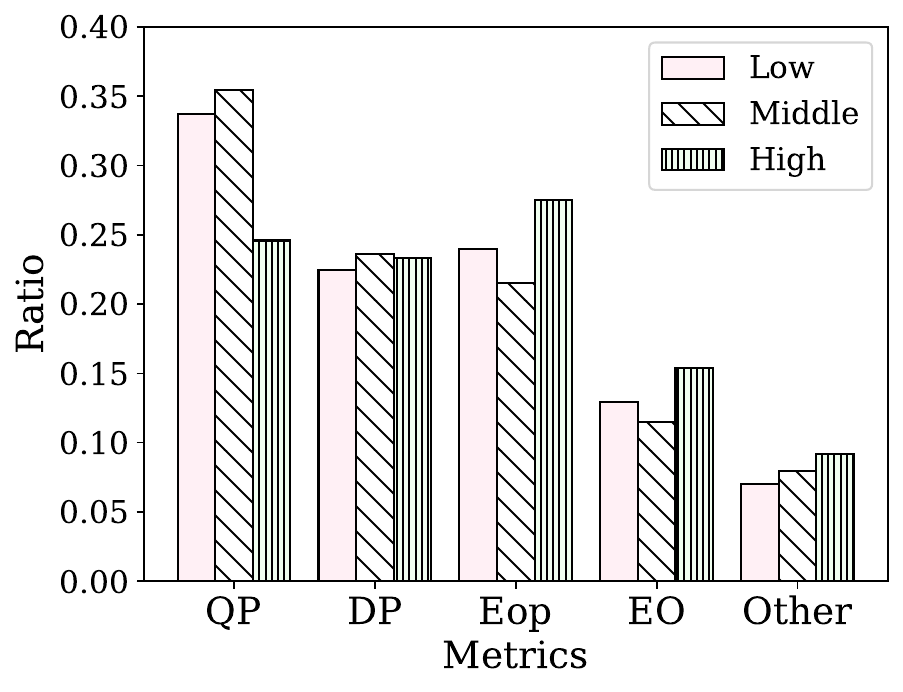}
        \subcaption{France**}
        \label{fig:hiring_boxplot_japan}
    \end{minipage}
    \begin{minipage}[t]{0.24\linewidth}
        \centering
        \includegraphics[width=1.0\linewidth]{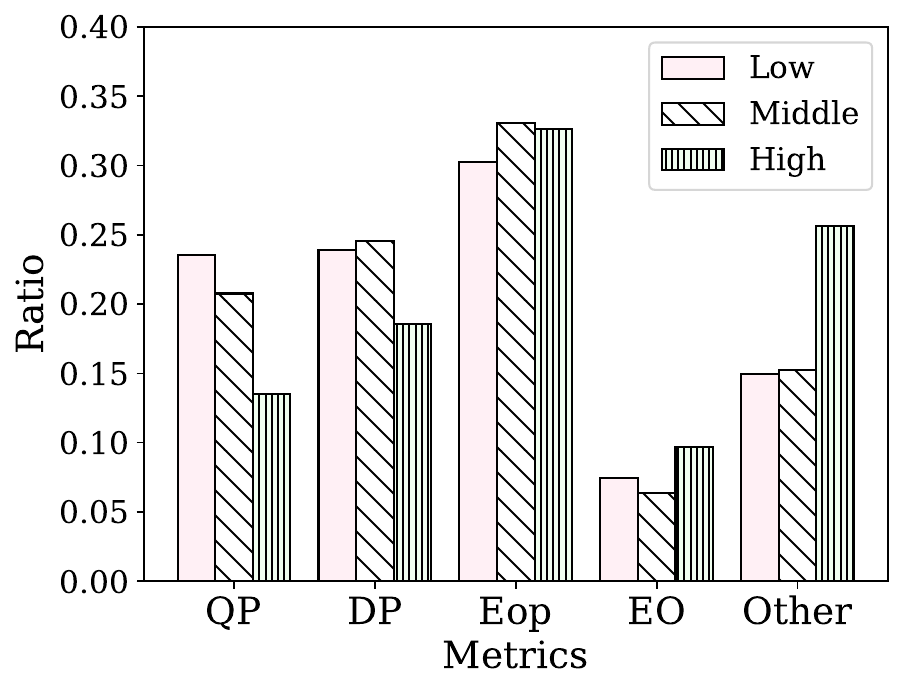}
        \subcaption{Japan***}
        \label{fig:hiring_boxplot_france}
    \end{minipage}
    \begin{minipage}[t]{0.24\linewidth}
    \centering
    \includegraphics[width=1.0\linewidth]{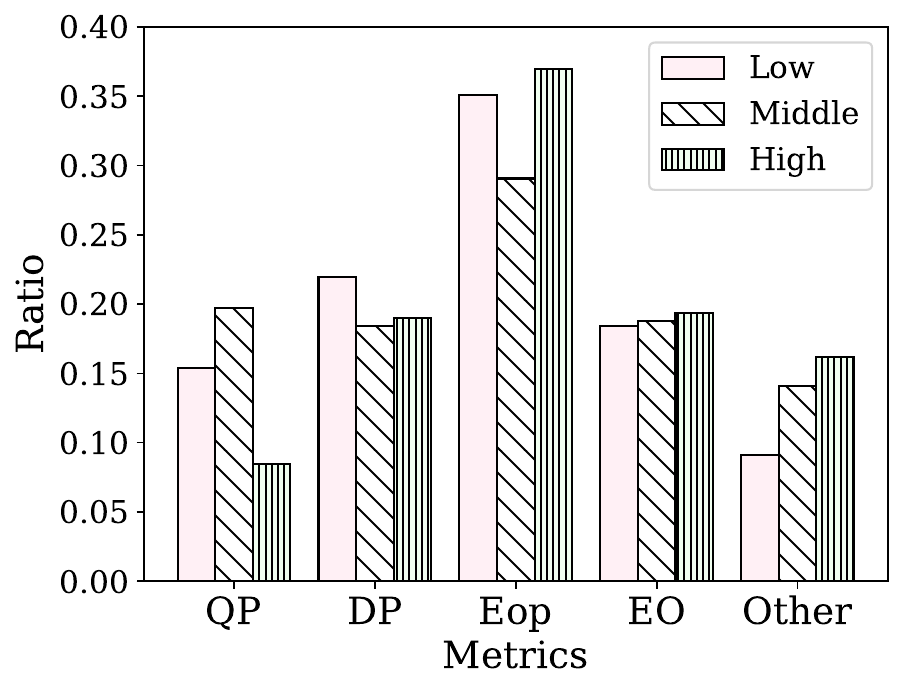}
    \subcaption{US***}
    \label{fig:hiring_boxplot_us}
    \end{minipage}
\caption{Difference between correctness in countries in employee award scenario}
\label{fig:employee_correctness_county}
\end{figure*}

%% file: 00-main.bbl

\begin{thebibliography}{44}


\ifx \showCODEN    \undefined \def \showCODEN     #1{\unskip}     \fi
\ifx \showDOI      \undefined \def \showDOI       #1{#1}\fi
\ifx \showISBNx    \undefined \def \showISBNx     #1{\unskip}     \fi
\ifx \showISBNxiii \undefined \def \showISBNxiii  #1{\unskip}     \fi
\ifx \showISSN     \undefined \def \showISSN      #1{\unskip}     \fi
\ifx \showLCCN     \undefined \def \showLCCN      #1{\unskip}     \fi
\ifx \shownote     \undefined \def \shownote      #1{#1}          \fi
\ifx \showarticletitle \undefined \def \showarticletitle #1{#1}   \fi
\ifx \showURL      \undefined \def \showURL       {\relax}        \fi
\providecommand\bibfield[2]{#2}
\providecommand\bibinfo[2]{#2}
\providecommand\natexlab[1]{#1}
\providecommand\showeprint[2][]{arXiv:#2}

\bibitem[Alesina and Angeletos(2005)]%
        {alesina2005fairness}
\bibfield{author}{\bibinfo{person}{Alberto Alesina} {and} \bibinfo{person}{George-Marios Angeletos}.} \bibinfo{year}{2005}\natexlab{}.
\newblock \showarticletitle{Fairness and redistribution}.
\newblock \bibinfo{journal}{\emph{American economic review}} \bibinfo{volume}{95}, \bibinfo{number}{4} (\bibinfo{year}{2005}), \bibinfo{pages}{960--980}.
\newblock


\bibitem[Alesina et~al\mbox{.}(2001)]%
        {alesina2001doesn}
\bibfield{author}{\bibinfo{person}{Alberto~F Alesina}, \bibinfo{person}{Edward~L Glaeser}, {and} \bibinfo{person}{Bruce Sacerdote}.} \bibinfo{year}{2001}\natexlab{}.
\newblock \bibinfo{title}{Why doesn't the US have a European-style welfare system?}
\newblock
\newblock


\bibitem[Anderson(2004)]%
        {anderson2004pursuit}
\bibfield{author}{\bibinfo{person}{Terry~H Anderson}.} \bibinfo{year}{2004}\natexlab{}.
\newblock \bibinfo{booktitle}{\emph{The Pursuit of Fairness: A History of Affirmative Action}}.
\newblock \bibinfo{publisher}{Oxford University Press}.
\newblock


\bibitem[Angwin et~al\mbox{.}(2022)]%
        {angwin2022machine}
\bibfield{author}{\bibinfo{person}{Julia Angwin}, \bibinfo{person}{Jeff Larson}, \bibinfo{person}{Surya Mattu}, {and} \bibinfo{person}{Lauren Kirchner}.} \bibinfo{year}{2022}\natexlab{}.
\newblock \showarticletitle{Machine bias}.
\newblock In \bibinfo{booktitle}{\emph{Ethics of data and analytics}}. \bibinfo{pages}{254--264}.
\newblock


\bibitem[Awad et~al\mbox{.}(2018)]%
        {awad2018moral}
\bibfield{author}{\bibinfo{person}{Edmond Awad}, \bibinfo{person}{Sohan Dsouza}, \bibinfo{person}{Richard Kim}, \bibinfo{person}{Jonathan Schulz}, \bibinfo{person}{Joseph Henrich}, \bibinfo{person}{Azim Shariff}, \bibinfo{person}{Jean-Fran{\c{c}}ois Bonnefon}, {and} \bibinfo{person}{Iyad Rahwan}.} \bibinfo{year}{2018}\natexlab{}.
\newblock \showarticletitle{The moral machine experiment}.
\newblock \bibinfo{journal}{\emph{Nature}} \bibinfo{volume}{563}, \bibinfo{number}{7729} (\bibinfo{year}{2018}), \bibinfo{pages}{59--64}.
\newblock


\bibitem[Bereni(2007)]%
        {bereni2007french}
\bibfield{author}{\bibinfo{person}{Laure Bereni}.} \bibinfo{year}{2007}\natexlab{}.
\newblock \showarticletitle{French feminists renegotiate republican universalism: The gender parity campaign}.
\newblock \bibinfo{journal}{\emph{French Politics}}  \bibinfo{volume}{5} (\bibinfo{year}{2007}), \bibinfo{pages}{191--209}.
\newblock


\bibitem[Blanca~Mena et~al\mbox{.}(2017)]%
        {blanca2017non}
\bibfield{author}{\bibinfo{person}{M~Jos{\'e} Blanca~Mena}, \bibinfo{person}{Rafael Alarc{\'o}n~Postigo}, \bibinfo{person}{Jaume Arnau~Gras}, \bibinfo{person}{Roser Bono~Cabr{\'e}}, {and} \bibinfo{person}{Rebecca Bendayan}.} \bibinfo{year}{2017}\natexlab{}.
\newblock \showarticletitle{Non-normal data: Is ANOVA still a valid option?}
\newblock \bibinfo{journal}{\emph{Psicothema, 2017, vol. 29, num. 4, p. 552-557}} (\bibinfo{year}{2017}).
\newblock


\bibitem[Buolamwini and Gebru(2018)]%
        {buolamwini2018gender}
\bibfield{author}{\bibinfo{person}{Joy Buolamwini} {and} \bibinfo{person}{Timnit Gebru}.} \bibinfo{year}{2018}\natexlab{}.
\newblock \showarticletitle{Gender shades: Intersectional accuracy disparities in commercial gender classification}. In \bibinfo{booktitle}{\emph{FAT}}. \bibinfo{pages}{77--91}.
\newblock


\bibitem[Caliskan et~al\mbox{.}(2017)]%
        {caliskan2017semantics}
\bibfield{author}{\bibinfo{person}{Aylin Caliskan}, \bibinfo{person}{Joanna~J Bryson}, {and} \bibinfo{person}{Arvind Narayanan}.} \bibinfo{year}{2017}\natexlab{}.
\newblock \showarticletitle{Semantics derived automatically from language corpora contain human-like biases}.
\newblock \bibinfo{journal}{\emph{Science}} \bibinfo{volume}{356}, \bibinfo{number}{6334} (\bibinfo{year}{2017}), \bibinfo{pages}{183--186}.
\newblock


\bibitem[Carey and Wu(2023)]%
        {carey2023statistical}
\bibfield{author}{\bibinfo{person}{Alycia~N Carey} {and} \bibinfo{person}{Xintao Wu}.} \bibinfo{year}{2023}\natexlab{}.
\newblock \showarticletitle{The statistical fairness field guide: perspectives from social and formal sciences}.
\newblock \bibinfo{journal}{\emph{AI and Ethics}} \bibinfo{volume}{3}, \bibinfo{number}{1} (\bibinfo{year}{2023}), \bibinfo{pages}{1--23}.
\newblock


\bibitem[Caton and Haas(2024)]%
        {10.1145/3616865}
\bibfield{author}{\bibinfo{person}{Simon Caton} {and} \bibinfo{person}{Christian Haas}.} \bibinfo{year}{2024}\natexlab{}.
\newblock \showarticletitle{Fairness in Machine Learning: A Survey}.
\newblock \bibinfo{journal}{\emph{ACM Comput. Surv.}} \bibinfo{volume}{56}, \bibinfo{number}{7} (\bibinfo{year}{2024}).
\newblock


\bibitem[Celis et~al\mbox{.}(2017)]%
        {celis2017ranking}
\bibfield{author}{\bibinfo{person}{L~Elisa Celis}, \bibinfo{person}{Damian Straszak}, {and} \bibinfo{person}{Nisheeth~K Vishnoi}.} \bibinfo{year}{2017}\natexlab{}.
\newblock \showarticletitle{Ranking with fairness constraints}.
\newblock \bibinfo{journal}{\emph{arXiv preprint arXiv:1704.06840}} (\bibinfo{year}{2017}).
\newblock


\bibitem[Cheng et~al\mbox{.}(2021)]%
        {cheng2021soliciting}
\bibfield{author}{\bibinfo{person}{Hao-Fei Cheng}, \bibinfo{person}{Logan Stapleton}, \bibinfo{person}{Ruiqi Wang}, \bibinfo{person}{Paige Bullock}, \bibinfo{person}{Alexandra Chouldechova}, \bibinfo{person}{Zhiwei Steven~Steven Wu}, {and} \bibinfo{person}{Haiyi Zhu}.} \bibinfo{year}{2021}\natexlab{}.
\newblock \showarticletitle{Soliciting stakeholders’ fairness notions in child maltreatment predictive systems}. In \bibinfo{booktitle}{\emph{CHI}}. \bibinfo{pages}{1--17}.
\newblock


\bibitem[Chouldechova(2017)]%
        {chouldechova2017fair}
\bibfield{author}{\bibinfo{person}{Alexandra Chouldechova}.} \bibinfo{year}{2017}\natexlab{}.
\newblock \showarticletitle{Fair prediction with disparate impact: A study of bias in recidivism prediction instruments}.
\newblock \bibinfo{journal}{\emph{Big data}} \bibinfo{volume}{5}, \bibinfo{number}{2} (\bibinfo{year}{2017}), \bibinfo{pages}{153--163}.
\newblock


\bibitem[Datta et~al\mbox{.}(2015)]%
        {datta2014automated}
\bibfield{author}{\bibinfo{person}{Amit Datta}, \bibinfo{person}{Michael~Carl Tschantz}, {and} \bibinfo{person}{Anupam Datta}.} \bibinfo{year}{2015}\natexlab{}.
\newblock \showarticletitle{Automated Experiments on Ad Privacy Settings}.
\newblock \bibinfo{journal}{\emph{Proc. Priv. Enhancing Technol.}} \bibinfo{volume}{2015}, \bibinfo{number}{1} (\bibinfo{year}{2015}), \bibinfo{pages}{92--112}.
\newblock


\bibitem[Fleisher(2021)]%
        {fleisher2021s}
\bibfield{author}{\bibinfo{person}{Will Fleisher}.} \bibinfo{year}{2021}\natexlab{}.
\newblock \showarticletitle{What's fair about individual fairness?}. In \bibinfo{booktitle}{\emph{AIES}}. \bibinfo{pages}{480--490}.
\newblock


\bibitem[Grgic-Hlaca et~al\mbox{.}(2018)]%
        {grgic2018human}
\bibfield{author}{\bibinfo{person}{Nina Grgic-Hlaca}, \bibinfo{person}{Elissa~M Redmiles}, \bibinfo{person}{Krishna~P Gummadi}, {and} \bibinfo{person}{Adrian Weller}.} \bibinfo{year}{2018}\natexlab{}.
\newblock \showarticletitle{Human perceptions of fairness in algorithmic decision making: A case study of criminal risk prediction}. In \bibinfo{booktitle}{\emph{Proceedings of the 2018 world wide web conference}}. \bibinfo{pages}{903--912}.
\newblock


\bibitem[Hardt et~al\mbox{.}(2016)]%
        {hardt2016equality}
\bibfield{author}{\bibinfo{person}{Moritz Hardt}, \bibinfo{person}{Eric Price}, {and} \bibinfo{person}{Nati Srebro}.} \bibinfo{year}{2016}\natexlab{}.
\newblock \showarticletitle{Equality of opportunity in supervised learning}.
\newblock \bibinfo{journal}{\emph{Advances in neural information processing systems}}  \bibinfo{volume}{29} (\bibinfo{year}{2016}).
\newblock


\bibitem[Harrison et~al\mbox{.}(2020)]%
        {harrison2020empirical}
\bibfield{author}{\bibinfo{person}{Galen Harrison}, \bibinfo{person}{Julia Hanson}, \bibinfo{person}{Christine Jacinto}, \bibinfo{person}{Julio Ramirez}, {and} \bibinfo{person}{Blase Ur}.} \bibinfo{year}{2020}\natexlab{}.
\newblock \showarticletitle{An empirical study on the perceived fairness of realistic, imperfect machine learning models}. In \bibinfo{booktitle}{\emph{FAccT}}. \bibinfo{pages}{392--402}.
\newblock


\bibitem[Kang et~al\mbox{.}(2020)]%
        {kang2020inform}
\bibfield{author}{\bibinfo{person}{Jian Kang}, \bibinfo{person}{Jingrui He}, \bibinfo{person}{Ross Maciejewski}, {and} \bibinfo{person}{Hanghang Tong}.} \bibinfo{year}{2020}\natexlab{}.
\newblock \showarticletitle{Inform: Individual fairness on graph mining}. In \bibinfo{booktitle}{\emph{KDD}}. \bibinfo{pages}{379--389}.
\newblock


\bibitem[Kelley et~al\mbox{.}(2021)]%
        {kelley2021exciting}
\bibfield{author}{\bibinfo{person}{Patrick~Gage Kelley}, \bibinfo{person}{Yongwei Yang}, \bibinfo{person}{Courtney Heldreth}, \bibinfo{person}{Christopher Moessner}, \bibinfo{person}{Aaron Sedley}, \bibinfo{person}{Andreas Kramm}, \bibinfo{person}{David~T Newman}, {and} \bibinfo{person}{Allison Woodruff}.} \bibinfo{year}{2021}\natexlab{}.
\newblock \showarticletitle{Exciting, useful, worrying, futuristic: Public perception of artificial intelligence in 8 countries}. In \bibinfo{booktitle}{\emph{AIES}}. \bibinfo{pages}{627--637}.
\newblock


\bibitem[Kim and Leung(2007)]%
        {KIM200783}
\bibfield{author}{\bibinfo{person}{Tae-Yeol Kim} {and} \bibinfo{person}{Kwok Leung}.} \bibinfo{year}{2007}\natexlab{}.
\newblock \showarticletitle{Forming and reacting to overall fairness: A cross-cultural comparison}.
\newblock \bibinfo{journal}{\emph{Organizational Behavior and Human Decision Processes}} \bibinfo{volume}{104}, \bibinfo{number}{1} (\bibinfo{year}{2007}), \bibinfo{pages}{83--95}.
\newblock
\showISSN{0749-5978}


\bibitem[Konow(2003)]%
        {konow2003fairest}
\bibfield{author}{\bibinfo{person}{James Konow}.} \bibinfo{year}{2003}\natexlab{}.
\newblock \showarticletitle{Which is the fairest one of all? A positive analysis of justice theories}.
\newblock \bibinfo{journal}{\emph{Journal of economic literature}} \bibinfo{volume}{41}, \bibinfo{number}{4} (\bibinfo{year}{2003}), \bibinfo{pages}{1188--1239}.
\newblock


\bibitem[Kusner et~al\mbox{.}(2017)]%
        {kusner2017counterfactual}
\bibfield{author}{\bibinfo{person}{Matt~J Kusner}, \bibinfo{person}{Joshua Loftus}, \bibinfo{person}{Chris Russell}, {and} \bibinfo{person}{Ricardo Silva}.} \bibinfo{year}{2017}\natexlab{}.
\newblock \showarticletitle{Counterfactual fairness}.
\newblock \bibinfo{journal}{\emph{Advances in neural information processing systems}} (\bibinfo{year}{2017}).
\newblock


\bibitem[Mehrabi et~al\mbox{.}(2021)]%
        {mehrabi2021survey}
\bibfield{author}{\bibinfo{person}{Ninareh Mehrabi}, \bibinfo{person}{Fred Morstatter}, \bibinfo{person}{Nripsuta Saxena}, \bibinfo{person}{Kristina Lerman}, {and} \bibinfo{person}{Aram Galstyan}.} \bibinfo{year}{2021}\natexlab{}.
\newblock \showarticletitle{A survey on bias and fairness in machine learning}.
\newblock \bibinfo{journal}{\emph{ACM computing surveys}} \bibinfo{volume}{54}, \bibinfo{number}{6} (\bibinfo{year}{2021}), \bibinfo{pages}{1--35}.
\newblock


\bibitem[Peterson et~al\mbox{.}(1994)]%
        {peterson1994event}
\bibfield{author}{\bibinfo{person}{MF Peterson}, \bibinfo{person}{M Radford}, \bibinfo{person}{G Savage}, \bibinfo{person}{Y Hama}, \bibinfo{person}{AM Bouvy}, \bibinfo{person}{F van~de Vijver}, \bibinfo{person}{P Boski}, {and} \bibinfo{person}{P Schmitz}.} \bibinfo{year}{1994}\natexlab{}.
\newblock \showarticletitle{Event management and evaluated department performance in US and Japanese local governments}.
\newblock \bibinfo{journal}{\emph{Journeys in Cross-Cultural Psychology}}  \bibinfo{volume}{1} (\bibinfo{year}{1994}), \bibinfo{pages}{374--385}.
\newblock


\bibitem[Plane et~al\mbox{.}(2017)]%
        {plane2017exploring}
\bibfield{author}{\bibinfo{person}{Angelisa~C Plane}, \bibinfo{person}{Elissa~M Redmiles}, \bibinfo{person}{Michelle~L Mazurek}, {and} \bibinfo{person}{Michael~Carl Tschantz}.} \bibinfo{year}{2017}\natexlab{}.
\newblock \showarticletitle{Exploring user perceptions of discrimination in online targeted advertising}. In \bibinfo{booktitle}{\emph{USENIX Security}}. \bibinfo{pages}{935--951}.
\newblock


\bibitem[Pleiss et~al\mbox{.}(2017)]%
        {pleiss2017fairness}
\bibfield{author}{\bibinfo{person}{Geoff Pleiss}, \bibinfo{person}{Manish Raghavan}, \bibinfo{person}{Felix Wu}, \bibinfo{person}{Jon Kleinberg}, {and} \bibinfo{person}{Kilian~Q Weinberger}.} \bibinfo{year}{2017}\natexlab{}.
\newblock \showarticletitle{On fairness and calibration}.
\newblock \bibinfo{journal}{\emph{NeurIPS}} (\bibinfo{year}{2017}).
\newblock


\bibitem[Reeskens and Van~Oorschot(2012)]%
        {reeskens2012disentangling}
\bibfield{author}{\bibinfo{person}{Tim Reeskens} {and} \bibinfo{person}{Wim Van~Oorschot}.} \bibinfo{year}{2012}\natexlab{}.
\newblock \showarticletitle{Disentangling the ‘New Liberal Dilemma’: On the relation between general welfare redistribution preferences and welfare chauvinism}.
\newblock \bibinfo{journal}{\emph{International Journal of Comparative Sociology}} \bibinfo{volume}{53}, \bibinfo{number}{2} (\bibinfo{year}{2012}), \bibinfo{pages}{120--139}.
\newblock


\bibitem[Roth et~al\mbox{.}(1991)]%
        {roth1991bargaining}
\bibfield{author}{\bibinfo{person}{Alvin~E Roth}, \bibinfo{person}{Vesna Prasnikar}, \bibinfo{person}{Masahiro Okuno-Fujiwara}, {and} \bibinfo{person}{Shmuel Zamir}.} \bibinfo{year}{1991}\natexlab{}.
\newblock \showarticletitle{Bargaining and market behavior in Jerusalem, Ljubljana, Pittsburgh, and Tokyo: An experimental study}.
\newblock \bibinfo{journal}{\emph{The American economic review}} (\bibinfo{year}{1991}), \bibinfo{pages}{1068--1095}.
\newblock


\bibitem[Saha et~al\mbox{.}(2020)]%
        {saha2020measuring}
\bibfield{author}{\bibinfo{person}{Debjani Saha}, \bibinfo{person}{Candice Schumann}, \bibinfo{person}{Duncan Mcelfresh}, \bibinfo{person}{John Dickerson}, \bibinfo{person}{Michelle Mazurek}, {and} \bibinfo{person}{Michael Tschantz}.} \bibinfo{year}{2020}\natexlab{}.
\newblock \showarticletitle{Measuring non-expert comprehension of machine learning fairness metrics}. In \bibinfo{booktitle}{\emph{ICML}}.
\newblock


\bibitem[Sengewald et~al\mbox{.}(2023)]%
        {sengewald2023}
\bibfield{author}{\bibinfo{person}{Julian Sengewald}, \bibinfo{person}{Anissa Schlichter}, \bibinfo{person}{Markus Siepermann}, {and} \bibinfo{person}{Richard Lackes}.} \bibinfo{year}{2023}\natexlab{}.
\newblock \showarticletitle{Human perceptions of fairness: a survey experiment}. In \bibinfo{booktitle}{\emph{Wirtschaftsinformatik}}.
\newblock


\bibitem[Singh and Joachims(2018)]%
        {singh2018fairness}
\bibfield{author}{\bibinfo{person}{Ashudeep Singh} {and} \bibinfo{person}{Thorsten Joachims}.} \bibinfo{year}{2018}\natexlab{}.
\newblock \showarticletitle{Fairness of exposure in rankings}. In \bibinfo{booktitle}{\emph{KDD}}. \bibinfo{pages}{2219--2228}.
\newblock


\bibitem[Smith and Bond(1999)]%
        {smith1999social}
\bibfield{author}{\bibinfo{person}{Peter~B Smith} {and} \bibinfo{person}{Michael~Harris Bond}.} \bibinfo{year}{1999}\natexlab{}.
\newblock \bibinfo{booktitle}{\emph{Social psychology: Across cultures}}.
\newblock \bibinfo{publisher}{Allyn \& Bacon}.
\newblock


\bibitem[Smith et~al\mbox{.}(1994)]%
        {smith1994event}
\bibfield{author}{\bibinfo{person}{Peter~B Smith}, \bibinfo{person}{Mark~F Peterson}, {and} \bibinfo{person}{Jyuji Misumi}.} \bibinfo{year}{1994}\natexlab{}.
\newblock \showarticletitle{Event management and work team effectiveness in Japan, Britain and USA}.
\newblock \bibinfo{journal}{\emph{Journal of Occupational and Organizational Psychology}} \bibinfo{volume}{67}, \bibinfo{number}{1} (\bibinfo{year}{1994}), \bibinfo{pages}{33--43}.
\newblock


\bibitem[Smith et~al\mbox{.}(2002)]%
        {smith2002cultural}
\bibfield{author}{\bibinfo{person}{Peter~B Smith}, \bibinfo{person}{Mark~F Peterson}, {and} \bibinfo{person}{Shalom~H Schwartz}.} \bibinfo{year}{2002}\natexlab{}.
\newblock \showarticletitle{Cultural values, sources of guidance, and their relevance to managerial behavior: A 47-nation study}.
\newblock \bibinfo{journal}{\emph{Journal of cross-cultural Psychology}} \bibinfo{volume}{33}, \bibinfo{number}{2} (\bibinfo{year}{2002}), \bibinfo{pages}{188--208}.
\newblock


\bibitem[Srivastava et~al\mbox{.}(2019)]%
        {srivastava2019mathematical}
\bibfield{author}{\bibinfo{person}{Megha Srivastava}, \bibinfo{person}{Hoda Heidari}, {and} \bibinfo{person}{Andreas Krause}.} \bibinfo{year}{2019}\natexlab{}.
\newblock \showarticletitle{Mathematical notions vs. human perception of fairness: A descriptive approach to fairness for machine learning}. In \bibinfo{booktitle}{\emph{KDD}}. \bibinfo{pages}{2459--2468}.
\newblock


\bibitem[Starke et~al\mbox{.}(2022)]%
        {starke2022fairness}
\bibfield{author}{\bibinfo{person}{Christopher Starke}, \bibinfo{person}{Janine Baleis}, \bibinfo{person}{Birte Keller}, {and} \bibinfo{person}{Frank Marcinkowski}.} \bibinfo{year}{2022}\natexlab{}.
\newblock \showarticletitle{Fairness perceptions of algorithmic decision-making: A systematic review of the empirical literature}.
\newblock \bibinfo{journal}{\emph{Big Data \& Society}} \bibinfo{volume}{9}, \bibinfo{number}{2} (\bibinfo{year}{2022}).
\newblock


\bibitem[Ullstein et~al\mbox{.}(2024)]%
        {ullstein2024attitudes}
\bibfield{author}{\bibinfo{person}{Chiara Ullstein}, \bibinfo{person}{Severin Engelmann}, \bibinfo{person}{Orestis Papakyriakopoulos}, \bibinfo{person}{Yuko Ikkatai}, \bibinfo{person}{Naira~Paola Arnez-Jordan}, \bibinfo{person}{Rose Caleno}, \bibinfo{person}{Brian Mboya}, \bibinfo{person}{Shuichiro Higuma}, \bibinfo{person}{Tilman Hartwig}, \bibinfo{person}{Hiromi Yokoyama}, {et~al\mbox{.}}} \bibinfo{year}{2024}\natexlab{}.
\newblock \showarticletitle{Attitudes toward facial analysis AI: A cross-national study comparing Argentina, Kenya, Japan, and the USA}. In \bibinfo{booktitle}{\emph{FAccT}}. \bibinfo{pages}{2273--2301}.
\newblock


\bibitem[Van~Berkel et~al\mbox{.}(2021)]%
        {van2021effect}
\bibfield{author}{\bibinfo{person}{Niels Van~Berkel}, \bibinfo{person}{Jorge Goncalves}, \bibinfo{person}{Daniel Russo}, \bibinfo{person}{Simo Hosio}, {and} \bibinfo{person}{Mikael~B Skov}.} \bibinfo{year}{2021}\natexlab{}.
\newblock \showarticletitle{Effect of information presentation on fairness perceptions of machine learning predictors}. In \bibinfo{booktitle}{\emph{CHI}}. \bibinfo{pages}{1--13}.
\newblock


\bibitem[Woodruff et~al\mbox{.}(2018)]%
        {woodruff2018qualitative}
\bibfield{author}{\bibinfo{person}{Allison Woodruff}, \bibinfo{person}{Sarah~E Fox}, \bibinfo{person}{Steven Rousso-Schindler}, {and} \bibinfo{person}{Jeffrey Warshaw}.} \bibinfo{year}{2018}\natexlab{}.
\newblock \showarticletitle{A qualitative exploration of perceptions of algorithmic fairness}. In \bibinfo{booktitle}{\emph{CHI}}. \bibinfo{pages}{1--14}.
\newblock


\bibitem[Yang and Stoyanovich(2017)]%
        {yang2017measuring}
\bibfield{author}{\bibinfo{person}{Ke Yang} {and} \bibinfo{person}{Julia Stoyanovich}.} \bibinfo{year}{2017}\natexlab{}.
\newblock \showarticletitle{Measuring fairness in ranked outputs}. In \bibinfo{booktitle}{\emph{Proceedings of the 29th international conference on scientific and statistical database management}}. \bibinfo{pages}{1--6}.
\newblock


\bibitem[Yokoyama et~al\mbox{.}(2024)]%
        {yokoyama2024can}
\bibfield{author}{\bibinfo{person}{Hiromi~M Yokoyama}, \bibinfo{person}{Yuko Ikkatai}, \bibinfo{person}{Euan McKay}, \bibinfo{person}{Atsushi Inoue}, \bibinfo{person}{Azusa Minamizaki}, {and} \bibinfo{person}{Kei Kano}.} \bibinfo{year}{2024}\natexlab{}.
\newblock \showarticletitle{Can affirmative action overcome STEM gender inequality in Japan? Expectations and concerns}.
\newblock \bibinfo{journal}{\emph{Asia Pacific Business Review}} \bibinfo{volume}{30}, \bibinfo{number}{3} (\bibinfo{year}{2024}), \bibinfo{pages}{543--559}.
\newblock


\bibitem[Yurrita et~al\mbox{.}(2023)]%
        {yurrita2023disentangling}
\bibfield{author}{\bibinfo{person}{Mireia Yurrita}, \bibinfo{person}{Tim Draws}, \bibinfo{person}{Agathe Balayn}, \bibinfo{person}{Dave Murray-Rust}, \bibinfo{person}{Nava Tintarev}, {and} \bibinfo{person}{Alessandro Bozzon}.} \bibinfo{year}{2023}\natexlab{}.
\newblock \showarticletitle{Disentangling fairness perceptions in algorithmic decision-making: the effects of explanations, human oversight, and contestability}. In \bibinfo{booktitle}{\emph{CHI}}. \bibinfo{pages}{1--21}.
\newblock


\end{thebibliography}
